\documentclass[10pt,twocolumn,letterpaper]{article}

\usepackage[pagenumbers]{cvpr} % To force page numbers, e.g. for an arXiv version

\usepackage{pifont}
\usepackage{caption}
\usepackage{booktabs}
\usepackage{arydshln}
\usepackage[dvipsnames]{xcolor}
\usepackage{graphicx}
\usepackage{multirow}
\usepackage{subcaption}
\usepackage{adjustbox}
\usepackage{overpic}
\usepackage{relsize}

\newcommand{\cmark}{\textcolor{ForestGreen}{\ding{51}}} % Green checkmark
\newcommand{\xmark}{\textcolor{red}{\ding{55}}}   % Red crossmark
\makeatletter
\DeclareRobustCommand\onedot{\futurelet\@let@token\@onedot}
\def\@onedot{\ifx\@let@token.\else.\null\fi\xspace}

\def\eg{\textit{e.g}\onedot} 
\def\ie{\textit{i.e}\onedot}

\makeatother

\newcommand{\diff}[1]{%
  \if\relax\detokenize{#1}\relax % Check if the input is empty
    \textcolor{DarkRed}{#1}%
  \else
    \IfBeginWith{#1}{+}{%
      (\textcolor{OliveGreen}{#1})%
    }{%
      (\textcolor{DarkRed}{#1})%
    }%
  \fi
}
\usepackage{booktabs}       % professional-quality tables
\usepackage{amsfonts}       % blackboard math symbols
\usepackage{nicefrac}       % compact symbols for 1/2, etc.
\usepackage{microtype}      % microtypography
\usepackage{xcolor}         % colors
\usepackage{graphicx}
\usepackage{amsmath,amsfonts}

\definecolor{cvprblue}{rgb}{0.21,0.49,0.74}
\usepackage[pagebackref,breaklinks,colorlinks,allcolors=cvprblue]{hyperref}

\newcommand{\ourmethod}{DreamCache} 
\newcommand{\supplmat}{Supp.\ Mat} 

\def\paperID{5613} % *** Enter the Paper ID here
\def\confName{CVPR}
\def\confYear{2025}

\title{%Cache your Reference\\
\ourmethod
: Finetuning-Free Lightweight\\ Personalized Image Generation via Feature Caching}

\author{Emanuele Aiello$^\diamond$ ~~~~ Umberto Michieli$^\dagger$ ~~~~ Diego Valsesia$^\diamond$ ~~~~ Mete Ozay$^\dagger$ ~~~~ Enrico Magli$^\diamond$ \\
$^\diamond$ Politecnico di Torino
~~~~
$^\dagger$ Samsung R\&D Institute UK
}

\begin{document}
\maketitle
\begin{abstract}

Personalized image generation requires text-to-image generative models that capture the core features of a reference subject to allow for controlled generation across different contexts. Existing methods face challenges due to complex training requirements, high inference costs, limited flexibility, or a combination of these issues. In this paper, we introduce \ourmethod, a scalable approach for efficient and high-quality personalized image generation. By caching a small number of reference image features from a subset of layers and a single timestep of the pretrained diffusion denoiser, \ourmethod\ enables dynamic modulation of the generated image features through lightweight, trained conditioning adapters.
\ourmethod\ achieves state-of-the-art image and text alignment, utilizing an order of magnitude fewer extra parameters, and is both more computationally effective and versatile than existing models. \href{https://emanuele97x.github.io/DreamCache/}{Project Page.}

\end{abstract}

\vspace{-8pt}
\section{Introduction}
\label{sec:intro}
\vspace{-4pt}
Recent advancements in text-to-image generation, fueled by the development of diffusion models \cite{sohl2015deep, ho2020denoising}, have enabled 
%outstanding capabilities in generating 
high-quality and diverse image generation from textual descriptions. 
Diffusion models \cite{rombach2022high, saharia2022photorealistic} gradually transform random noise into images through a sequence of denoising steps, conditioned on the input text prompt. 

An active area of research is personalizing these models, enabling the generation of novel images of a reference subject in various contexts, while maintaining flexibility for text-based editing.
%An active area of research is to personalize these models so that reference subjects can be provided to generate personalized images of such subjects in various contexts while retaining flexibility in editing.
Early personalization techniques \cite{ruiz2023dreambooth, gal2022image, yang2023controllable, voynov2023p+, arar2023domain, hao2023vico, yang2023controllable, tewel2023key}, such as the seminal DreamBooth \cite{ruiz2023dreambooth} relied on fine-tuning (FT) the generative model for each reference subject. 
However, these approaches are often impractical for many use cases due to costly test-time FT, which can take several minutes per subject.
%Therefore, they tend to be impractical for many applications due to costly test-time FT, which can take several minutes per subject. 
To address this, FT-free (\ie, zero-shot) personalized image generation methods have emerged to eliminate test-time optimization. These FT-free approaches can be broadly categorized into two families: encoder-based methods and reference-based methods, each with distinct drawbacks.

Encoder-based methods \cite{Wei2023ELITEEV, gal2023encoder, ma2023subject, ye2023ip, patel2024lambda} utilize dedicated image encoders, such as CLIP \cite{radford2021learning} or DINO \cite{caron2021emerging}, to extract relevant features from reference images. 
While these encoders can produce high-quality results, they are often large, require extensive training to align text and image features, and reduce the model's flexibility \cite{ye2023ip, ma2023subject, li2023blip, pan2023kosmos}.

\begin{figure}[t]
    \centering
    \includegraphics[width=0.65\linewidth]{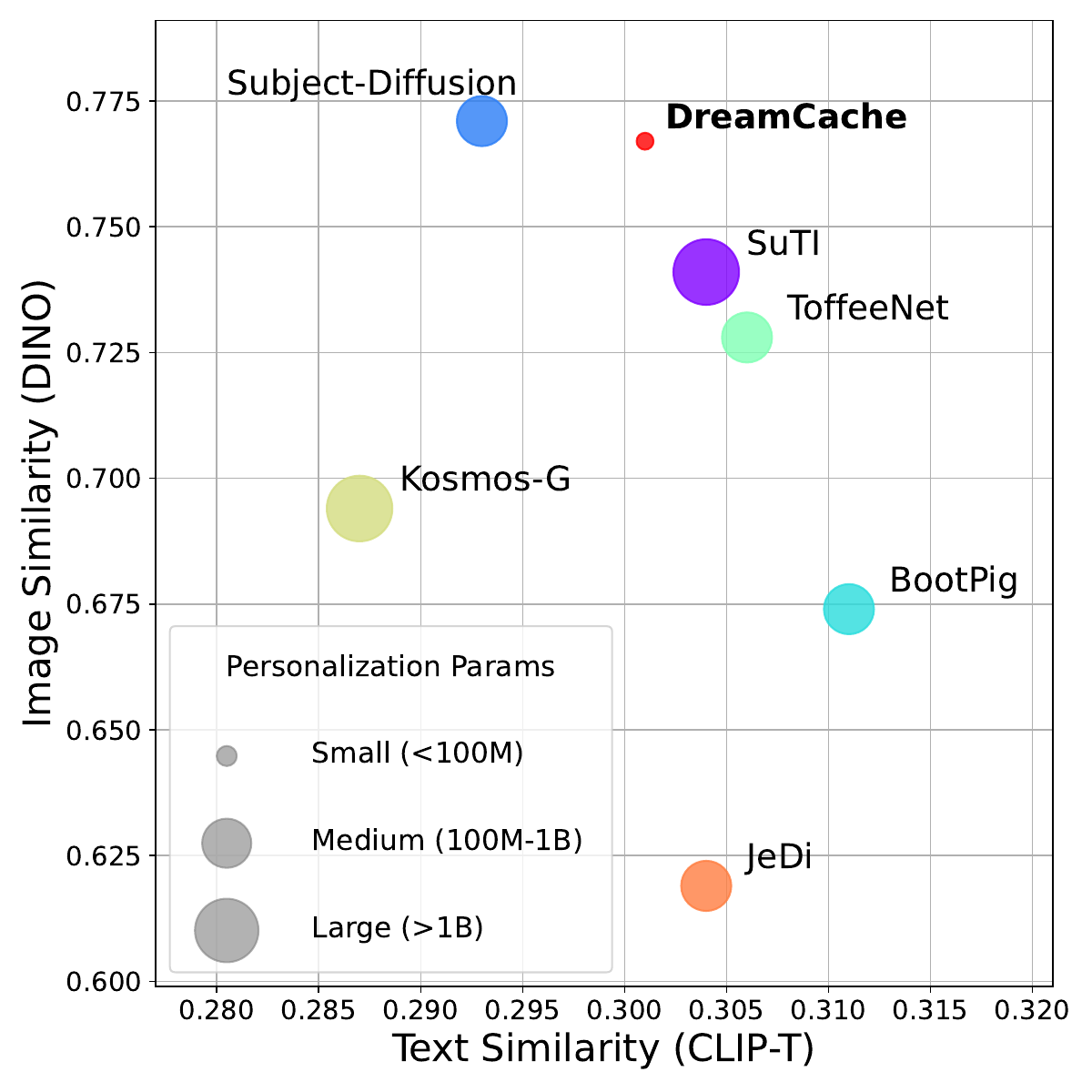}
    \caption{\textbf{DreamCache} is a finetuning-free personalized image generation method that achieves an optimal balance between subject fidelity, memory efficiency, and adherence to text prompts.}
    \label{fig:graphical_abstract}
\end{figure}

\begin{table*}[ht]
    \centering
    \setlength{\tabcolsep}{1.8pt}
    \caption{\textbf{Methods overview.} Our \ourmethod\ achieves state-of-the-art generation quality at reduced computational costs. *: value refers to the personalization stage for each personal subject.}
        \begin{tabular}{lcccccccc}
            \toprule
            
        Method  & FT-free & Enc-free & Plug\&Play & Ref-UNet-free & Extra Params  & Train Params &  \# Dataset & Train Time \\
          
            \midrule

            Textual Inversion~\cite{gal2022image}  & \xmark & \cmark & \cmark & \cmark & 768* & 768* & 3-5* & 50 min* \\

            DreamBooth~\cite{ruiz2023dreambooth}  & \xmark & \cmark & \xmark & \cmark & -  & 0.9B* & 3-5* & 
            10 min* \\

            Custom Diffusion~\cite{kumari2023multi}  & \xmark & \cmark & \xmark & \cmark & - & 57M* &  3-5* & 10 min* \\ 
            \midrule

            ELITE~\cite{Wei2023ELITEEV} & \cmark & \xmark & \xmark & \cmark & 457M & 77M & 125K & 14 days \\

            BLIP-Diffusion~\cite{li2023blip} & \cmark & \xmark & \cmark & \cmark & 380M & 1.5B & 129M & 96 days \\

            IP-Adapter~\cite{ye2023ip} & \cmark & \xmark & \cmark & \cmark & 402M & 22M & 10M & 28 days \\
            
            Kosmos-G~\cite{pan2023kosmos} & \cmark & \xmark & \cmark & \cmark & 1.6B & 1.6B &  9M &  -  \\

             JeDi \cite{zeng2024jedi} & \cmark & \cmark & \xmark & \xmark & - & 0.9B & 3M & 48 days \\

            SuTI \cite{chen2023subject} & \cmark & \xmark & \xmark & \cmark & 400M & 2.5B & 500K & -                    \\
             Subject-Diffusion~\cite{ma2023subject} & \cmark & \xmark & \cmark & \xmark & 700M & 700M & 76M & - \\
           
            BootPig    \cite{purushwalkam2024bootpig} & \cmark & \cmark & \xmark & \xmark & 0.95B & 0.95B & 200K & 18 hours                    \\

            ToffeeNet \cite{zhou2024toffee} & \cmark & \xmark & \xmark & \cmark & 632M & 0.9B & 5M & -                     \\
            CAFE     \cite{zhou2024customization} & \cmark & \xmark & \xmark & \cmark & 14B & 1B & 355K & -                   \\
            \textbf{\ourmethod\ (ours)} & \cmark & \cmark  & \cmark & \cmark & 25M & 25M &  400K &  40 hours \\
             % & & & & (no extra params) \\
            
            \bottomrule
        \end{tabular}
    
    \label{tab:previous_methods_overview}
\end{table*}

In contrast, reference-based methods \cite{purushwalkam2024bootpig, zeng2024jedi} condition the diffusion model directly on reference features drawn from the U-Net denoiser, integrating these features at each denoising step. While effective, these methods require feature extraction at every generation step, leading to higher computational costs and memory demands. Additionally, they often require an input textual caption for conditioning, 
%throughout the image encoding process, 
which introduces variability and can decrease output precision.

Some recent works have proposed to finetune the U-Net backbone itself \cite{purushwalkam2024bootpig, zeng2024jedi, zhou2024toffee}. However, this hinders the model's ability to switch between personalized and non-personalized tasks and risks inducing the ``language drift'' phenomenon, where the personalization training degrades the model's linguistic  comprehension \cite{ruiz2023dreambooth, kumari2023multi}.

In this work, we propose \ourmethod, a novel finetuning-free approach to personalized image generation, that overcomes the limitations of existing methods (see Fig.~\ref{fig:graphical_abstract}) by using a feature caching mechanism that enables text-free encoding and efficient conditioning during personalization.

Specifically, we first create a synthetic dataset \cite{purushwalkam2024bootpig} containing triplets of captions, target images, and reference subjects to capture subjects in various contexts. 
Next, we pretrain lightweight attention-based conditioning adapters to inject subject-specific features into the image generation process. 
During personalization, the reference image is processed through the pretrained denoiser of the base diffusion model without text conditioning, thus eliminating the need for user-generated captions, while caching features from a small subset of layers at a single timestep. 
For personalized sampling, these cached features are injected into the denoiser through the pretrained conditioning adapters.

Table~\ref{tab:previous_methods_overview} summarizes the key properties of existing methods and illustrates how \ourmethod\ fits within the current landscape; further details are explored in Sec.~\ref{sec:bkg}. 
As an encoder-free approach, \ourmethod\ introduces only a small number of additional parameters, making it significantly lightweight and suitable for deployment on resource-constrained devices.
%such as mobile. 
For example, methods like \cite{Wei2023ELITEEV} and \cite{li2023blip} introduce 380M parameters due to their reliance on CLIP encoders, whereas \ourmethod\ requires only 25M additional parameters. Moreover, caching features from a few selected U-Net layers at a single preprocessing timestep bypasses the need for full U-Net reference processing during generation, leading to substantial computational and memory savings that enable real-time, high-quality personalized generation. Another key advantage of \ourmethod\ is its \textit{plug-and-play} nature, allowing concurrent generation of personalized and non-personalized content without altering original U-Net weights, thus preserving the integrity of the original model and enabling a wider range of deployment scenarios, especially on mobile platforms.

In summary, \ourmethod\ represents a significant step toward practical, and scalable personalized image generation, with the following contributions:
% Overall, we believe that \ourmethod\ represents a significant step forward in making personalized image generation practical for a wide range of applications. Our contributions can be summarized as follows:

\begin{itemize}
    \item We propose a feature caching approach that creates multi-resolution representations of the reference image in a caption-free and efficient manner.

    \item We design an attention-based conditioning mechanism that leverages the cached features for personalized image generation, achieving computational- and memory- efficient personalized sampling.
    %that is lightweight, efficient and reduces the computational burden during personalization stage.
    
    \item Our approach achieves state-of-the-art quality in personalized image generation at substantially lower computational and data costs compared to existing methods.
    
    %\item We commit to releasing our code, model, and synthetic dataset to promote further research upon acceptance.

\end{itemize}

\section{Background and Related Work}\label{sec:bkg}

Personalized image generation aims at generating images containing a specific subject. This task has been widely studied, resulting in two main approaches: fine-tuning methods, which require test-time finetuning on multiple subject reference images, and finetuning-free (zero-shot) methods, which learn a generalizable conditioning mechanism to generate reference subjects without the need for further optimization.

\begin{figure*}[htbp]
    \centering
    \setlength{\tabcolsep}{1.5pt} % Adjust padding between columns if needed
    \begin{tabular}{c c c c c}
        % Column prompts (First row)
        \textit{``A dragon...''} & \textit{``as street graffiti''} & \textit{``playing with fire''} & \textit{``as a plushie''} & \textit{``working as a barista''} \\
        
        % First sample images
        \includegraphics[width=0.19\linewidth]{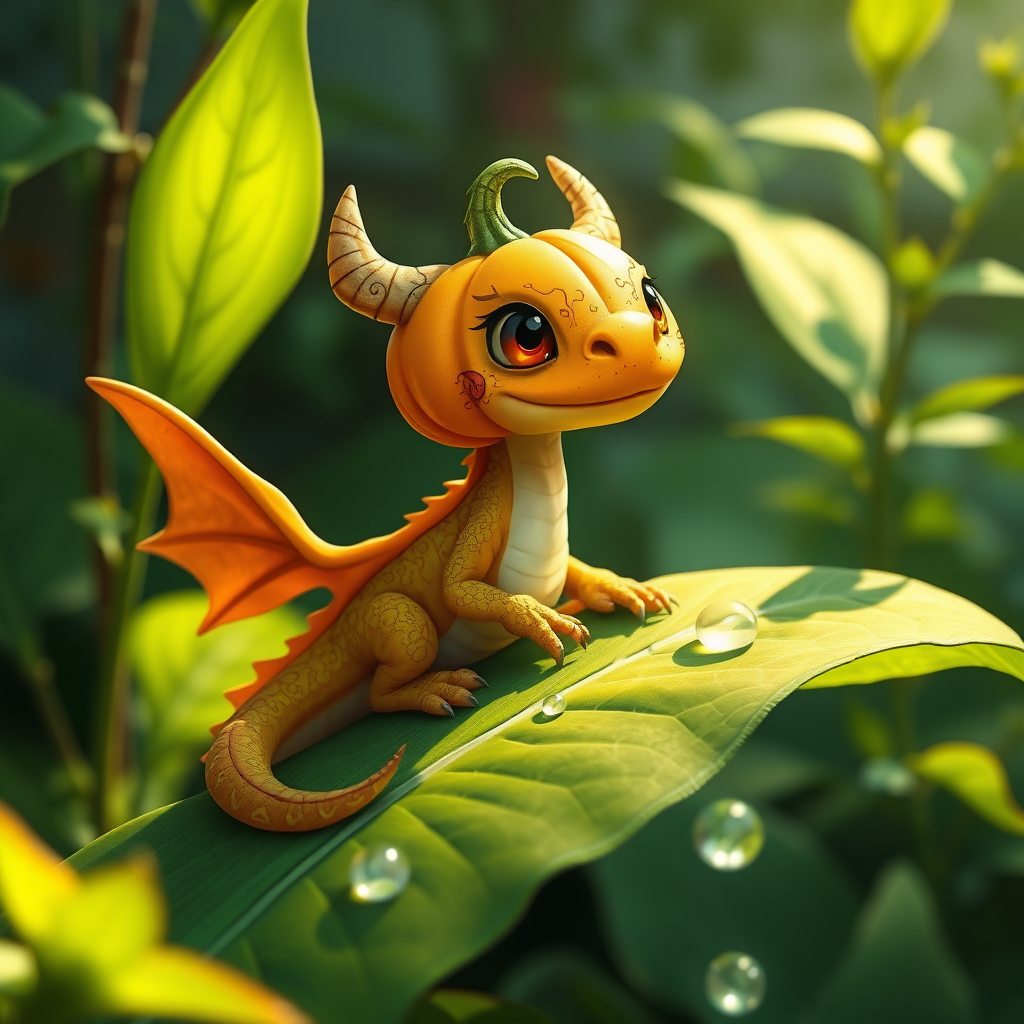} &
        \includegraphics[width=0.19\linewidth]{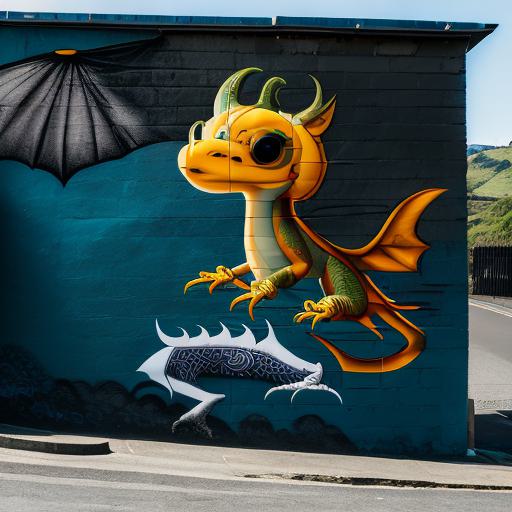} &
        \includegraphics[width=0.19\linewidth]{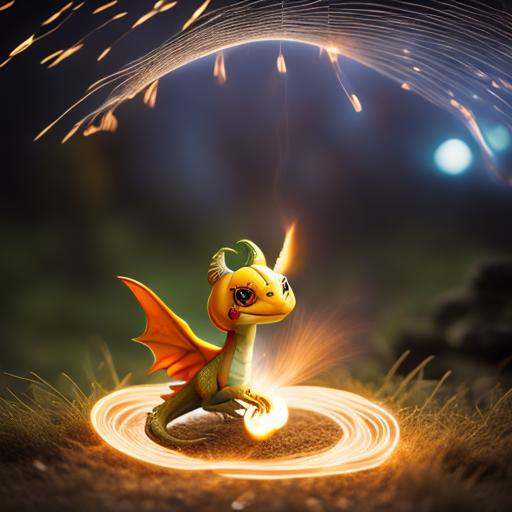} &
        \includegraphics[width=0.19\linewidth]{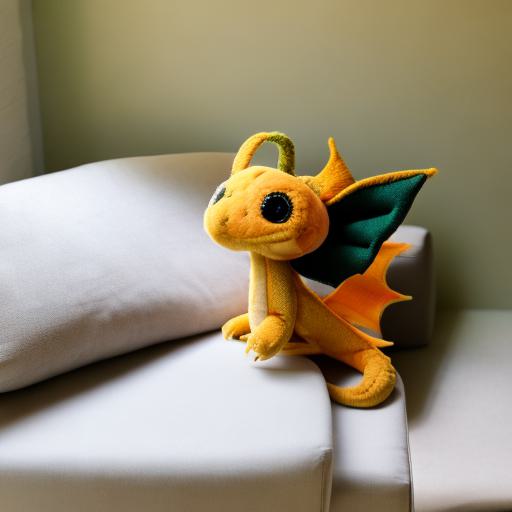} &
        \includegraphics[width=0.19\linewidth]{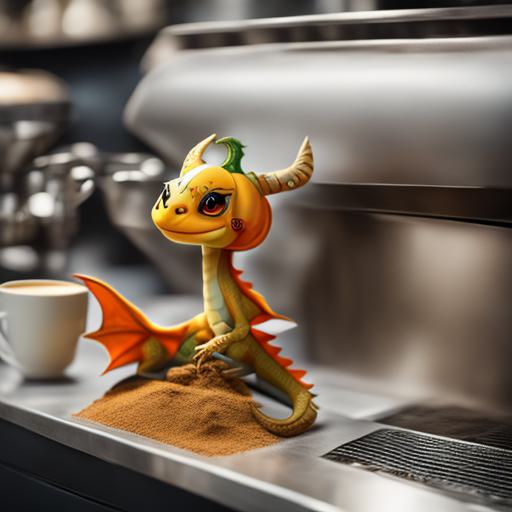} \\

        \textit{``A cat...''} & \textit{``in Ukiyo-e style''} & \textit{``with a rainbow scarf''} & \textit{``Van Gogh painting''} & \textit{``wearing a diploma hat''} \\
        
        % Second sample images
        \includegraphics[width=0.19\linewidth]{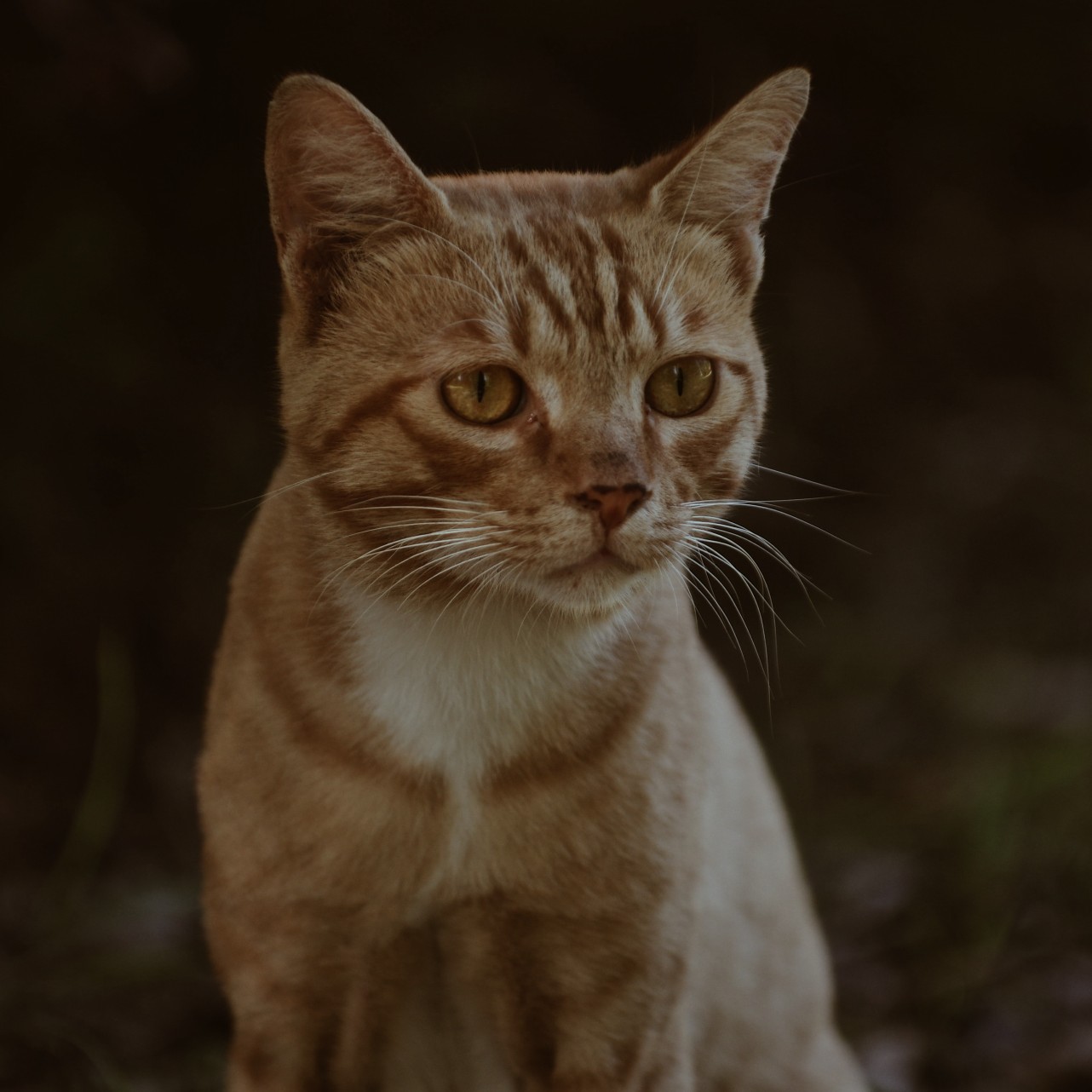} &
        \includegraphics[width=0.19\linewidth]{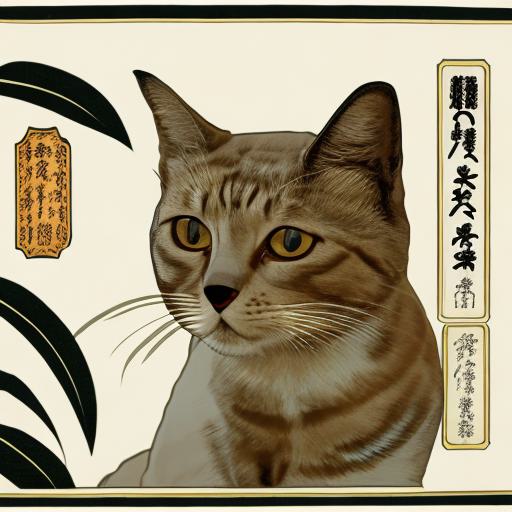} &
        \includegraphics[width=0.19\linewidth]{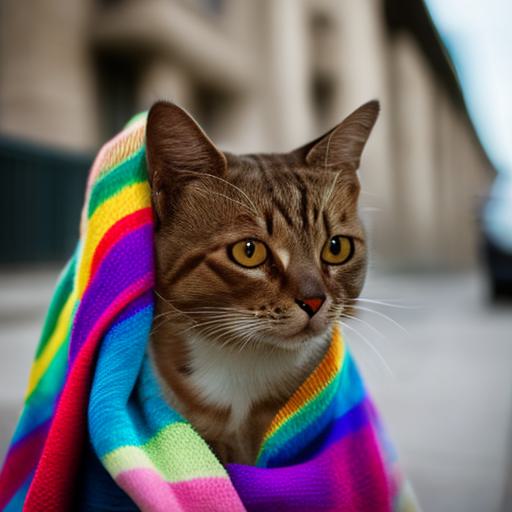} &
        \includegraphics[width=0.19\linewidth]{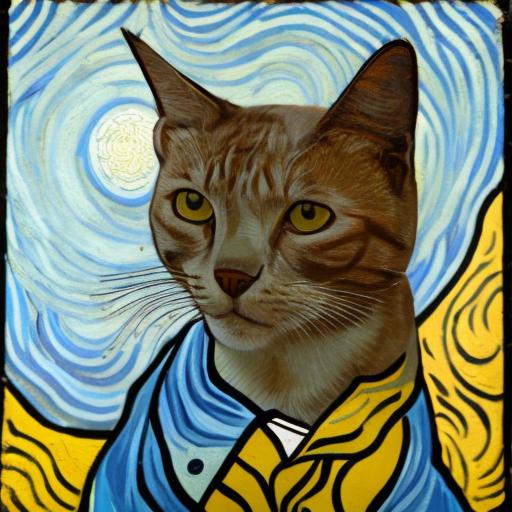} &
        \includegraphics[width=0.19\linewidth]{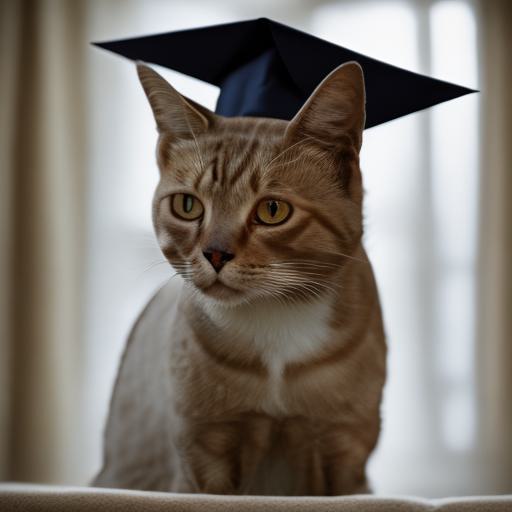} \\
        \vspace{3pt}

    \end{tabular}
    \caption{\textbf{Personalized generations by \ourmethod.} The first column contains reference images. The generated images correspond to the text prompts above each column.}
    \label{fig:comparison_modified}
\end{figure*}

\vspace{-7pt}
\paragraph{Finetuning-based Personalization} DreamBooth \cite{ruiz2023dreambooth} finetunes the entire U-Net with reference images while introducing a regularization loss to mitigate overfitting. On the other hand, Custom Diffusion \cite{kumari2023multi} only finetunes the $K$ and $V$ projections for the cross-attention blocks of the U-Net. 
Text-based personalization methods optimize single (like Textual Inversion \cite{gal2022image}) or multiple (like in P+ \cite{voynov2023p+}) input token embeddings. Later methods \cite{arar2023domain, hao2023vico, yang2023controllable, tewel2023key} build on these, with innovations like Perfusion \cite{tewel2023key} using dynamic rank-1 updates to prevent overfitting while keeping encodings lightweight. However, all finetuning-based methods are computationally intensive, often requiring minutes of finetuning per reference subject at test time.
%Other methods \cite{gal2022image, yang2023controllable, voynov2023p+} personalize the model by optimizing input text token embeddings. Pioneered by Textual Inversion \cite{gal2022image}, which fine-tunes a single token embedding, this line of work has been extended by P+ \cite{voynov2023p+} to support multiple token embeddings. Subsequent works \cite{arar2023domain, hao2023vico, yang2023controllable, tewel2023key} have further improved FT methods. For instance, Perfusion \cite{tewel2023key} employs dynamic rank-1 updates to prevent overfitting, resulting in lightweight encodings for personalized concepts. However, these FT-based methods are computationally expensive as they require at least several minutes of finetuning at test time for each subject to be used for personalization.
\vspace{-5pt}
\paragraph{Finetuning-Free Personalization}
To reduce computational demands, recent research has shifted toward zero-shot personalization methods that eliminate subject-specific finetuning, typically employing image encoders to condition the generation process via the features extracted from the reference images \cite{Wei2023ELITEEV, gal2023encoder, ma2023subject, ye2023ip, patel2024lambda}. Examples include BLIP-Diffusion \cite{li2023blip}, which pretrains a Q-Former to learn image features aligned with text, and IP-Adapter, which uses a frozen CLIP encoder to extract text-aligned visual features that modulate the cross-attention layers of the generative model. Other approaches, like Kosmos-G \cite{pan2023kosmos} and CAFE \cite{zhou2024customization}, connect large language models (LLMs) with diffusion models to condition generation on personalized concepts. SuTI \cite{chen2023subject} takes a different approach by training millions of subject-specific experts and subsequently training a model via apprenticeship learning, enabling effective zero-shot personalized generation at test time.

Alternatively, encoder-free methods such as JeDi \cite{zeng2024jedi} and BootPIG \cite{purushwalkam2024bootpig} use features from the generative model’s backbone to guide the generation. 
%Other works avoid separate encoders (encoder-free) and rather utilize features from the generative model's backbone to guide the generation \cite{zeng2024jedi, purushwalkam2024bootpig}. 
JeDi creates a multi-view synthetic dataset and modifies spatial self-attention to jointly attend to images of the same concept in a batch. BootPIG retains a trainable copy of the original U-Net, adding reference self-attention layers to enable adaptation of the personalized model for reference features.
%and inserts trainable reference self-attention layers to adapt the personalized model to accept reference features. 
While these methods remove the need for an additional encoder, they still require computationally expensive inference, as reference images must be processed in parallel during generation-—a cost that accumulates due to the iterative nature of the diffusion process.
%This overhead becomes particularly costly due to the iterative nature of the diffusion process.

In contrast, our method, \ourmethod, caches a subset of reference features from the U-Net without text conditioning, eliminating the need for parallel inference and reducing memory overhead by avoiding separate model loading. This results in a test-time computational efficiency similar to encoder-based methods while offering the flexibility of encoder-free feature extraction and injection.

Recent works such as BootPIG \cite{purushwalkam2024bootpig} and Toffee-5M \cite{zhou2024toffee} emphasize the importance of synthetic data that explicitly decouples the subject from the background, reporting improved performance. Inspired by these approaches, we adopt a similar generative pipeline to create a synthetic dataset we use to train \ourmethod.

Moreover, since our method is plug-and-play and keeps the U-Net frozen, we reduce the high cost of training faced by other approaches \cite{ma2023subject, pan2023kosmos, purushwalkam2024bootpig, zeng2024jedi, zhou2024toffee, zhou2024customization}. A detailed overview of current methods, the number of trained parameters, and their training cost can be found in Table~\ref{tab:previous_methods_overview}.

%\textbf{DV: PLUG AND PLAY NON E' SPIEGATO. NON SI DICE NULLA SUL COSTO DI TRAINING DI CHI DEVE FINETUNARE TUTTA LA UNET}
%plug and play spiegato nell-introduzione,  costo di training aggiunto

\begin{figure*}[!t]
    \centering
    \includegraphics[width=\linewidth]{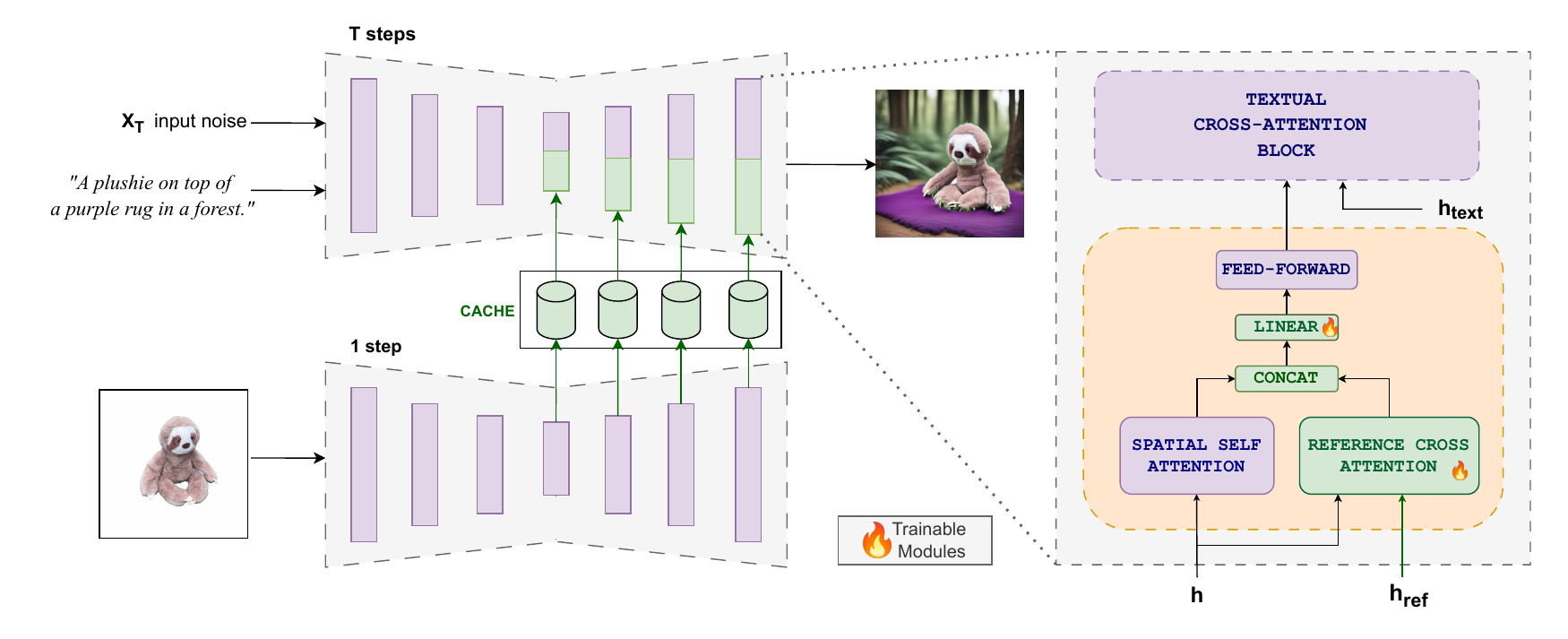}
    \caption{\textbf{Overview of \ourmethod.} Original U-Net layers are shown in violet, while the novel components introduced by \ourmethod\ are highlighted in green. During personalization, features from selected layers of the diffusion denoiser are cached from a single timestep, using a null text prompt. These cached features serve as reference-specific information. During generation, conditioning adapters inject the cached features into the denoiser, modulating the features of the generated image to create a personalized output. 
    }
    \label{fig:method}
\end{figure*}
\vspace{-5pt}
\paragraph{Feature Caching}
Feature caching has been explored to reduce generation time in diffusion models by caching intermediate activations. 
% %Feature caching concepts have been explored as a means to accelerate generation in diffusion models by caching intermediate activations. 
Some studies \cite{wimbauer2024cache, ma2024deepcache} exploit temporal redundancy during the training process to cache activations across timesteps, reducing computational load at later timesteps.
Other works focus on caching layer activations within the diffusion framework, avoiding redundant computations. Learning-to-Cache \cite{ma2024learning} introduces a dynamic caching mechanism that learns to skip computation for selected layers of the diffusion model. 
In contrast to those works, which generally cache intra-model features for some layers to save computations, we utilize feature caching to encode multi-resolution features of a reference image from a few selected layers to condition the generation process of a new personalized image. 
Our approach recalls successful few-shot learners for discriminative problems \cite{snell2017prototypical,triantafillou2018meta,oreshkin2018tadam} and extends them to personalized image generation.

\section{Method}

% epsilon al momento non è usato
Given a pretrained text-to-image generative model $\epsilon_\theta$ and an image containing a reference subject $\mathbf{I}_\text{ref}$, the goal of personalized sampling is to generate novel images containing the reference subject in various contexts while maintaining textual control.
We propose \ourmethod, a novel approach for extracting conditioning signals from $\mathbf{I}_\text{ref}$ and guiding of the image generation process. This method leverages a pretrained diffusion model, conditioning adapters that are pretrained with a synthetic dataset,
%for the purpose of personalization 
and a feature cache from the reference image. Sample outputs generated by \ourmethod\ are shown in Fig.~\ref{fig:comparison_modified}, with a method overview in Fig.~\ref{fig:method}.

At the core of \ourmethod, we utilize the denoiser in the pretrained diffusion model to extract multi-resolution features from $\mathbf{I}_\text{ref}$ by caching the activations of a few selected layers. 
% The core of \ourmethod\ relies on using the denoiser in the pretrained diffusion model to extract multi-resolution features of $\mathbf{I}_\text{ref}$ by caching the activations of a few selected layers.
To improve generalization, we cache features using a forward pass with a null text prompt.
%For the purpose of caching, a void text prompt is used to enhance generalization. 
When personalized sampling is performed, the cached features are processed by adapters to act as conditioning signals, modulating the denoiser features of the image under generation at corresponding layers.
%the same layers from which the caching was performed. 
These adapters, once pretrained on a synthetic dataset, enable zero-shot personalized generation with any new reference image, requiring no further finetuning once its features are cached.
% These subnetworks are extra parameters that are trained once on a synthetic dataset and allow for zero-shot personalized generation with any new reference image, without the need for further fine-tuning, once its features are cached with the aforementioned mechanism.

In the following, we detail the three main aspects of \ourmethod, namely i) how to cache reference features (Sec.~\ref{sec:method:caching}); ii) how to condition the diffusion model on the cached features for personalized sampling (Sec.~\ref{sec:method:conditioning}); iii) how to train the adapters used for model conditioning (Sec.~\ref{sec:method:training}).

\vspace{-4pt}
\subsection{Caching Reference Features}
\label{sec:method:caching}

To extract information from the reference image for personalized sampling, we perform a forward pass through the denoiser of the diffusion model at a single timestep. We select $t = 1$, the least noisy timestep, to obtain clean features that are optimal for conditioning the personalized generation process. Additionally, we remove the text conditioning to decouple visual content of the reference image from the text caption, thus also eliminating the need for user-provided captions for reference images. This contrasts with methods such as JeDi \cite{zeng2024jedi}, which are sensitive to caption content.

During the forward pass, activations are computed for all layers of the denoiser, but only a subset is cached. Based on our experiments with the Stable Diffusion U-Net, we find that caching features from a middle bottleneck layer and every second layer in the decoder offers the best balance between generation quality and caching efficiency (see Sec.~\ref{sec:ablations}).
%The aforementioned forward pass computes activations for all layers of the denoiser. However, we only cache a subset of those. Specifically, with reference to the U-Net in Stable Diffusion, we experimentally found that only caching features of a middle bottleneck layer and every second layer in the decoder is the best tradeoff between generation quality and amount of caching (see Sec. \ref{sec:ablations}).

%More formally, the feature cache $\mathcal{H}_\text{FC}$ is the set of features output by denoiser layers in the selected set $\mathcal{L}$ at timestep $t=1$ for a null text prompt $\varnothing$ and reference image $\mathbf{I}_\text{ref}$, i.e.,
Formally, the feature cache $\mathcal{H}_\text{FC}$ consists of the activations of the denoiser $\epsilon_{\theta}$ from the selected layers $\mathcal{L}$ at timestep $t=1$, using a null text prompt $\varnothing$ and noisy reference image $\mathbf{I}_\text{ref} + \mathbf{n}_t$, with noise realization $\mathbf{n}_t$, expressed as: 
\begin{align}
    \mathcal{H}_\text{FC} &= \left\lbrace \mathbf{h}_{\text{ref},L} : L \in \mathcal{L}  \right\rbrace \\
    \mathbf{h}_{\text{ref},L} &= \epsilon_{\theta}(\mathbf{I}_\text{ref}+\mathbf{n}_t,\varnothing,t;l)\vert_{t=1,l=L}. 
    \label{eq:cache}
\end{align}

% More formally, we can compute the multi-resolution cached features $H_c$ with a forward pass through the denoising backbone $\epsilon_{\theta}$ for selected layers as:
% \begin{align}
%     H_c = \epsilon_{\theta}(\mathbf{x}_t, \varnothing , t)
% \end{align}
% Where $t=1$ and $\varnothing$ is the null-text conditioning.

Notice that the cached features have different spatial resolutions, from the low-resolution bottleneck layer to the higher-resolution decoder layers, allowing a multi-resolution representation of the reference image. This is particularly useful for enabling both global semantics and fine-grained detail guidance.
%Notice that the cached features present different spatial resolutions, from the lower one of the bottleneck layer, to the higher one in the decoder, allowing a multi-resolution representation of the reference image. This is particularly useful to have guidance not only on global semantics but also fine-grained details.
%
As in prior work \cite{purushwalkam2024bootpig, Wei2023ELITEEV, ma2023subject}, we foreground-segment the reference image to isolate the subject from the background before caching its features. 
% We also remark that, consistently with existing methods \cite{}, the reference image is foreground-segmented to isolate the subject of personalization from the background before being processed for feature caching. We used ...

\subsection{Conditioning on Cached Reference Features} \label{sec:method:conditioning}

We propose a novel conditioning adapter mechanism composed of: i) a cross-attention block between the cached features and the features of the image under generation; ii) a concatenation operation between the features from the self-attention block of the original U-Net denoising backbone and those of the cross-attention block; and iii) a projection layer. A block diagram is shown in Fig.~\ref{fig:method} (right).

Omitting the layer subscript for clarity, the conditioning adapter mechanism is expressed mathematically as follows:
\begin{align}
&\mathbf{q} = \mathbf{W}_Q \mathbf{h}, \quad \mathbf{k}_c = \mathbf{W}_K \mathbf{h}_\text{ref}, \quad \mathbf{v}_c = \mathbf{W}_V \mathbf{h}_\text{ref}, \\
&\mathbf{a}_c = \text{softmax}\left( \frac{\mathbf{q} \mathbf{k}_c^T}{\sqrt{d}} \right)\mathbf{v}_c, \\
&\mathbf{a} = \mathbf{W}_{\text{proj}} \left( [\mathbf{a}; \mathbf{a}_c] \right),
\end{align}
%da aggiungere tutti i mathcal quando le formule sono finalizzate. DV: mathcal solo per gli insiemi. vettori e matrici in mathbf
%
where $\mathbf{h} \in \mathbb{R}^{N \times d}$ are the current $d$-dimensional features of the $N$-pixel image under generation, and $\mathbf{h}_\text{ref}$ are the cached reference features from Eq.~\eqref{eq:cache}. $\mathbf{W}_Q$, $\mathbf{W}_K$, $\mathbf{W}_V$, and $\mathbf{W}_{\text{proj}}$ are learnable projection matrices, whose training is described in Sec.~\ref{sec:method:training}. The concatenation $[\mathbf{a}; \mathbf{a}_c] \in \mathbb{R}^{N \times 2d}$ combines the output of 
%the primary 
self-attention 
%over the image under generation 
$\mathbf{a} \in \mathbb{R}^{N \times d}$ and 
%the 
cross-attention 
%output 
$\mathbf{a}_c \in \mathbb{R}^{N \times d}$.
The concatenation operation allows a flexible information fusion without explicit alignment constraints, compared to other approaches in similar works (see Sec.~\ref{sec:ablations}).
The learnable projection matrix $\mathbf{W}_{\text{proj}}$ reduces the dimensionality of the concatenated features back to $\mathbb{R}^{N \times d}$ to maintain compatibility with the original backbone.  

%Instead of simply summing their contributions, which may misalign scales, we concatenate the outputs of the self-attention and cross-attention along the feature dimension and apply a learnable linear projection to reduce the dimensionality of the concatenated features to match the original U-Net backbone.

Overall, the approach proposed for the adapter enriches feature representations used in the diffusion process of the image under generation by allowing the model to leverage both primary and conditioning-based contextual information from the cache.
%without explicit alignment constraints. 

\subsection{Training the Conditioning Adapters} \label{sec:method:training}

The additional parameters introduced in Sec.~\ref{sec:method:conditioning} to process the cached features must be trained on a large and varied dataset to ensure they generalize for any reference subject. 

Collecting paired data for this training process would be prohibitively expensive, as it requires multiple images of the same subject in different contexts. To address this, we draw inspiration from the recently proposed synthetic data generation pipeline in BootPIG \cite{purushwalkam2024bootpig} to construct our training data. First, we utilize a large language model (Llama 3.2 \cite{dubey2024llama}) to generate captions for potential target images. Each caption is used to generate an image via Stable Diffusion \cite{rombach2022high}. We then use the Segment Anything Model (SAM) \cite{kirillov2023segment} and Grounding DINO \cite{liu2023grounding} to accurately segment the reference subject based on the text caption and generate a foreground mask of the main object in the caption.

We treat the Stable Diffusion-generated image as the target image, the foreground object pasted on a white background as the reference image, and the LLM-generated caption as the textual prompt during our training pipeline. Compared to BootPIG, our pipeline employs open-source models, making it more accessible. 
%, we have also doubled the number of generated training data, demonstrating improved performance. 
We will release our synthetic dataset to facilitate reproducibility and further research, since similar datasets, including BootPIG's \cite{purushwalkam2024bootpig}, have not been publicly released. Additional details on the dataset and its statistics can be found in the \supplmat.

We train the newly introduced adapters' parameters ($\mathbf{W}_Q$, $\mathbf{W}_K$, $\mathbf{W}_V$, and $\mathbf{W}_{\text{proj}}$) with the standard score matching loss \cite{song2019generative} using both the text-conditioned noisy input and the cached reference features:
%for the model having both the text-conditioned noisy input and the cached reference features.
\begin{equation}
\mathcal{L}_{\text{diffusion}} = \mathbb{E}_{\mathbf{x}_0, \epsilon, \mathbf{c}_{T}, \mathbf{I}_\text{ref}, t} \left[ || \epsilon - \epsilon_{\theta}'(\mathbf{x}_t, \mathbf{c}_{T}, \mathcal{H}_\text{FC}, t) ||^2_2\right],
\end{equation}
where $\mathbf{x}_0$ is the target image, $\mathbf{c}_{T}$ is the text prompt generated by the large language model, $\epsilon$ is Gaussian noise, and $t$ is the diffusion timestep sampled uniformly from ${1, \dots, T}$. The noisy image at timestep $t$, $\mathbf{x}_t$, is obtained by gradually adding noise to $\mathbf{x}_0$ during the forward diffusion process. The function $\epsilon_{\theta}'$ represents the modified denoising model that incorporates the conditioning adapters.

\section{Experimental Results}
In this section, we present our experimental results, including quantitative and qualitative comparisons, an ablation study, and an analysis section that visualizes the behavior of the newly introduced cross-attention mechanism in the adapters.

%\subsection{Finetuning on multiple references}

\begin{table*}[t]
    \centering
    \setlength{\tabcolsep}{3.5pt}
    \caption{\textbf{Quantitative results on DreamBooth.} \ourmethod\ obtains a better balance between DINO score and CLIP-T  compared to all baselines,
    %\ourmethod\ sets a new state of the art and 
    while also offering a more efficient computational tradeoff (see Table~\ref{tab:previous_methods_overview}). }
    \begin{tabular}{l l c c c c c}
        \toprule
        & Method  & Backbone & \#Ref & DINO ($\uparrow$) & CLIP-I ($\uparrow$) & CLIP-T ($\uparrow$) \\
        \midrule
        \multirow{6}{*}{\rotatebox{90}{\parbox{2cm}{\centering test-time finetuning}}}
        & DreamBooth~\cite{ruiz2023dreambooth}  & Imagen & 3-5  & 0.696 & 0.812 & 0.306 \\
        & DreamBooth~\cite{ruiz2023dreambooth}  & SD 1.5  & 3-5 & 0.668 & 0.803 & 0.305 \\
        & Textual Inversion~\cite{gal2022image}  & SD 1.5  & 3-5 & 0.569 & 0.780 & 0.255 \\
        & Custom Diffusion~\cite{kumari2023multi}  & SD 1.5 & 3-5  & 0.643 & 0.790 & 0.305 \\
        & BLIP-Diffusion (FT)~\cite{li2023blip} & SD 1.5 &  3-5 & 0.670 & 0.805 & 0.302 \\
        %& \textbf{\ourmethod\ (ours, FT)} & SD 1.5 &  3-5 & \\
        \midrule
        \multirow{13}{*}{\rotatebox{90}{\parbox{2cm}{\centering finetuning free}}}
        & ELITE~\cite{Wei2023ELITEEV} & SD 1.5 & 1 & 0.621 & 0.771 & 0.293 \\
        & BLIP-Diffusion~\cite{li2023blip} & SD 1.5 & 1 & 0.594 & 0.779 & 0.300  \\
        & IP-Adapter~\cite{ye2023ip} & SD 1.5 & 1 &  0.667 & 0.813 & 0.289 \\
        & Kosmos-G~\cite{pan2023kosmos} & SD 1.5 & 1 & 0.694 & 0.847 & 0.287  \\
        & Jedi~\cite{zeng2024jedi} & SD 1.5  &  1 & 0.619 & 0.782 & 0.304    \\
        & \textbf{\ourmethod\ (ours)} & SD 1.5 & 1 & 0.713 & 0.810 & 0.298 \\
        \cdashline{2-7}
        & Re-Imagen~\cite{chen2022re} & Imagen & 1-3 & 0.600 & 0.740 & 0.270 \\
        & SuTI~\cite {chen2023subject}& Imagen & 1-3 &  0.741 & 0.819 & 0.304 \\
        & Subject-Diffusion~\cite{ma2023subject} & SD 2.1 & 1 & 0.771 & 0.779 & 0.293 \\
        & BootPig~\cite{purushwalkam2024bootpig}& SD 2.1 & 3 & 0.674 & 0.797 & 0.311 \\ 
        & ToffeeNet~\cite{zhou2024toffee} & SD 2.1 & 1 & 0.728 & 0.817 & 0.306 \\
        & CAFE~\cite{zhou2024customization} & SD 2.1 & 1 & 0.715 & 0.827 & 0.294 \\
        & \textbf{\ourmethod\ (ours)} & SD 2.1 & 1 & 0.767  &  0.816 & 0.301 \\
        \bottomrule
    \end{tabular}
    \label{tab:dreambench}
\end{table*}

\vspace{-5pt}
\paragraph{Implementation Details}
We evaluate our method on two versions of Stable Diffusion (SD) \cite{rombach2022high}, specifically versions 1.5 and 2.1, to ensure fair comparison with state-of-the-art methods across different backbones. 
As described in the ablation study, our caching and conditioning mechanism is applied to the middle layer and every second layer of the decoder. 
The total number of trainable parameters for \ourmethod\ is 25M. 
We use the original SD codebase and train the model on 4$\times$ 80GB A100 GPUs for 25k steps with a batch size of 128, using the AdamW optimizer with a learning rate of $10^{-5}$.

Input images are resized to $512 \times 512$, with scale, shift, and resize augmentations applied to reference images to enhance model robustness to 
%reference 
perturbations. Ablations are conducted on SD 1.5. We generate images with 50 sampling steps, employing classifier-free guidance for image and text conditioning, using a guidance scale of 7.5. 
%Detailed descriptions of the sampling process and the impact of parameters can be found in the Supplementary Material.

\vspace{-5pt}
\paragraph{Evaluation}
Quantitative evaluations are conducted on the DreamBooth dataset \cite{ruiz2023dreambooth}, following prior approaches.
DreamBooth consists of 30 subjects, each with 25 text prompts. We use a single input image per subject and generate 4 images per subject-prompt combination, resulting in 3,000 generated images. 
We use pretrained DINO ViT-S/16 and CLIP ViT-B/32 models to calculate the average cosine similarity of global image embeddings between generated and reference images, with metrics denoted as DINO and CLIP-I, respectively. To assess text alignment, we calculate the cosine similarity between embeddings from generated images and text prompts using CLIP's image and text encoders \cite{hessel2021clipscore}, with the corresponding score denoted as CLIP-T.

\begin{figure*}[htbp]
    \centering
    \setlength{\tabcolsep}{1.5pt} % Adjust padding between columns if needed
    \begin{tabular}{c c ccc ccc}
        % First sample prompts
        \multicolumn{1}{c}{\textit{``A dog''}} & &\multicolumn{3}{c}{\textit{``wearing a santa hat''}} &  \multicolumn{3}{c}{\textit{``with the Eiffel Tower in the background''}} \\
        
        % First sample images
        \includegraphics[width=0.135\linewidth]{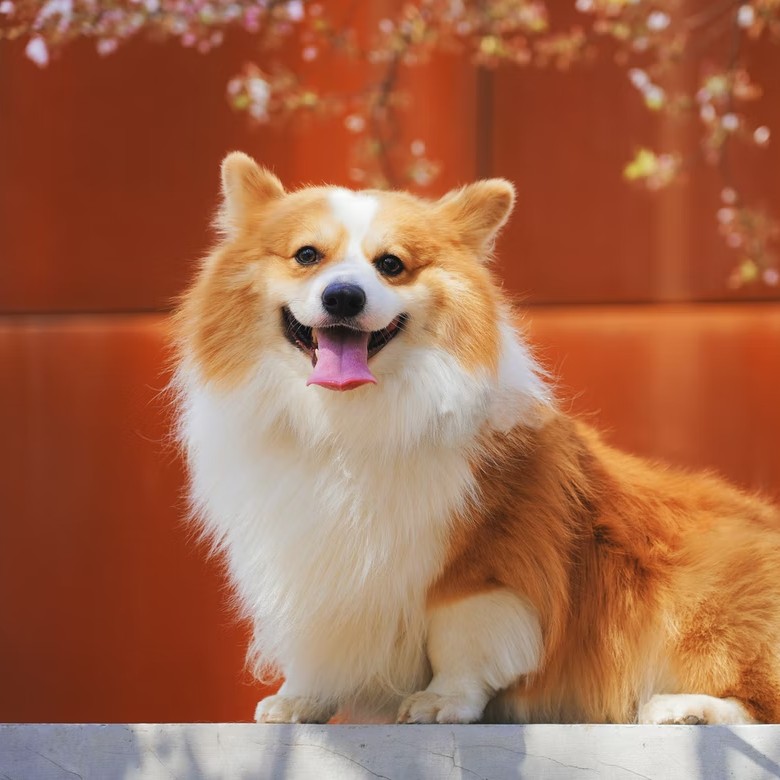} &
        &
        \includegraphics[width=0.135\linewidth]{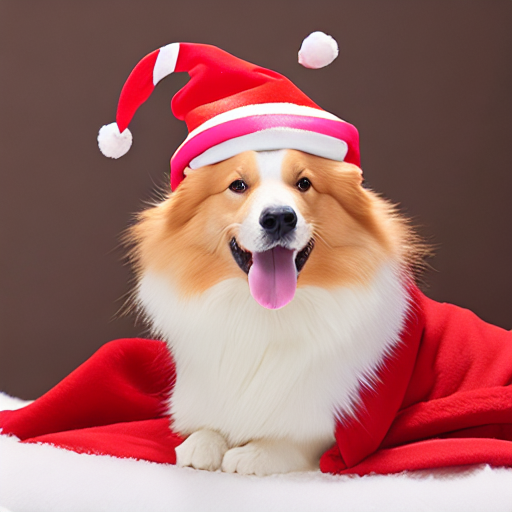} &
        \includegraphics[width=0.135\linewidth]{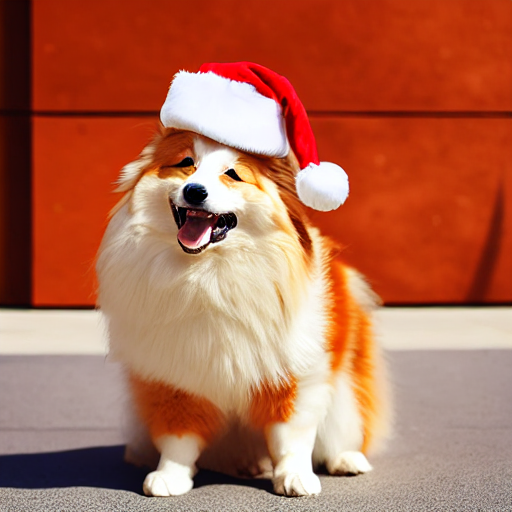} &
        \includegraphics[width=0.135\linewidth]{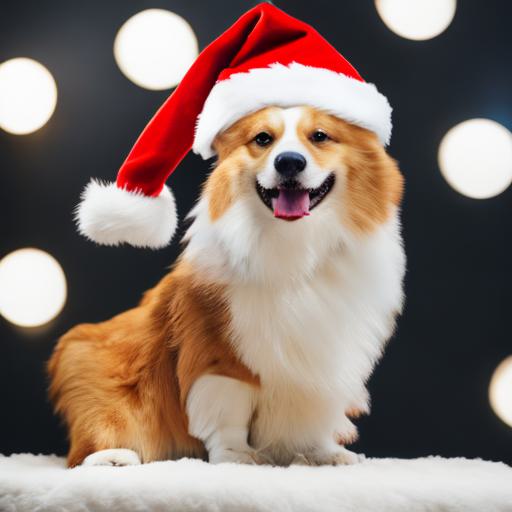} &
        \includegraphics[width=0.135\linewidth]{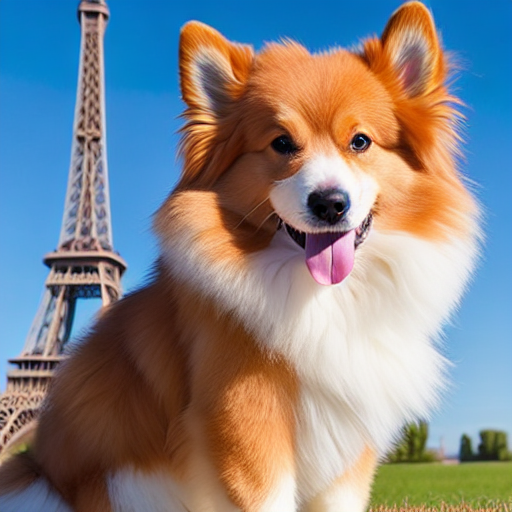} &
        \includegraphics[width=0.135\linewidth]{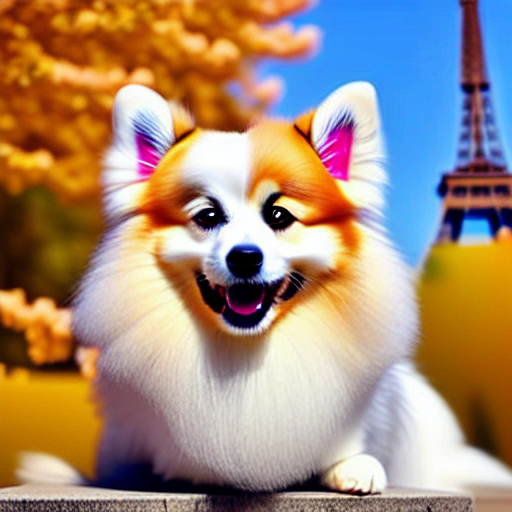} &
        \includegraphics[width=0.135\linewidth]{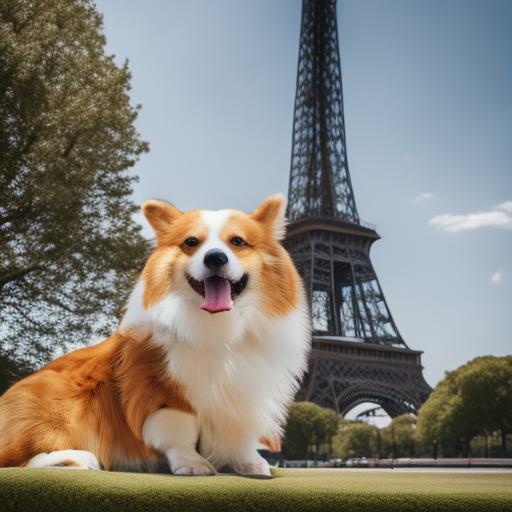} \\

        % Second sample prompts
        \multicolumn{1}{c}{\textit{``A can''}} & & \multicolumn{3}{c}{\textit{``floating on top of water''}} &  \multicolumn{3}{c}{\textit{``with a mountain in the background''}} \\
        % Second sample images
        \includegraphics[width=0.135\linewidth]{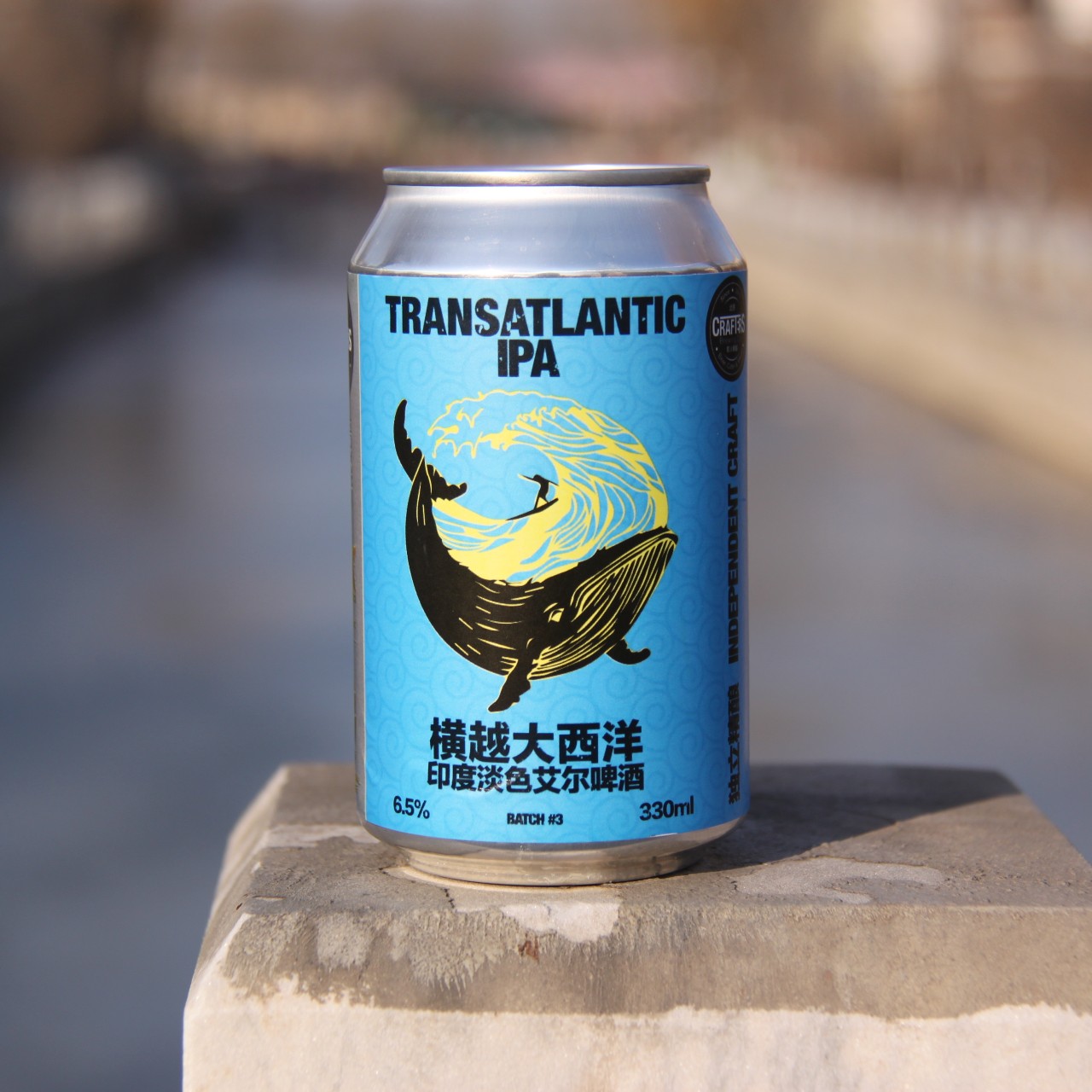} &
        &
        \includegraphics[width=0.135\linewidth]{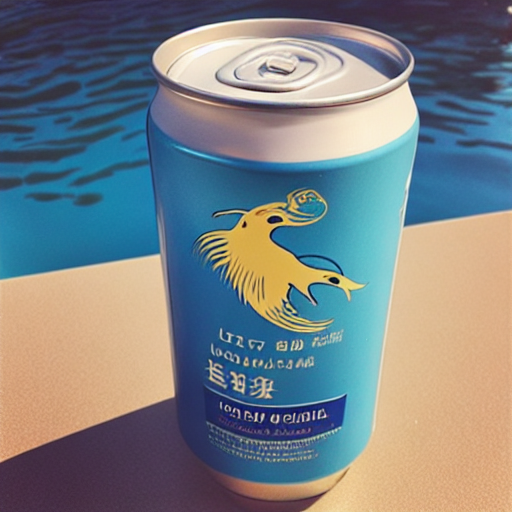} &
        \includegraphics[width=0.135\linewidth]{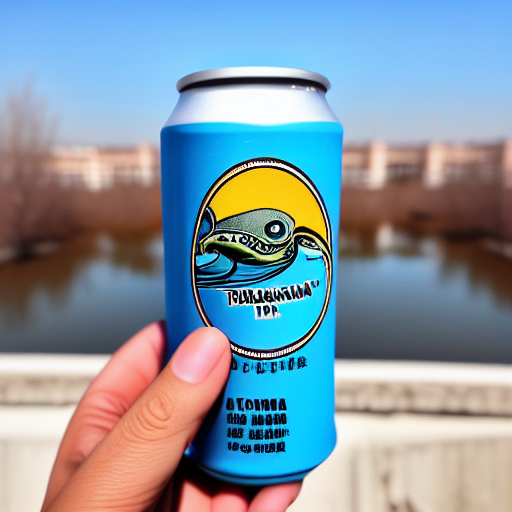} &
        \includegraphics[width=0.135\linewidth]{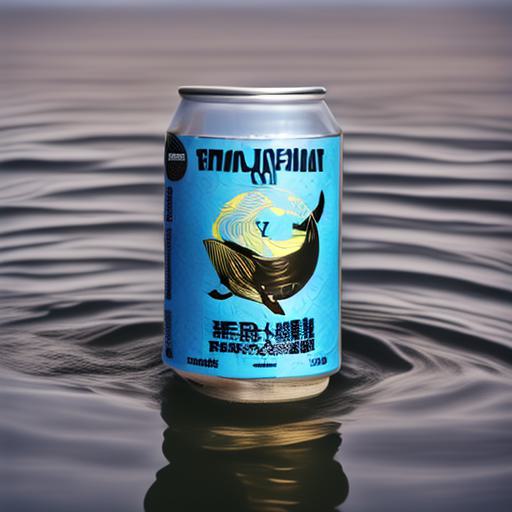} &
        \includegraphics[width=0.135\linewidth]{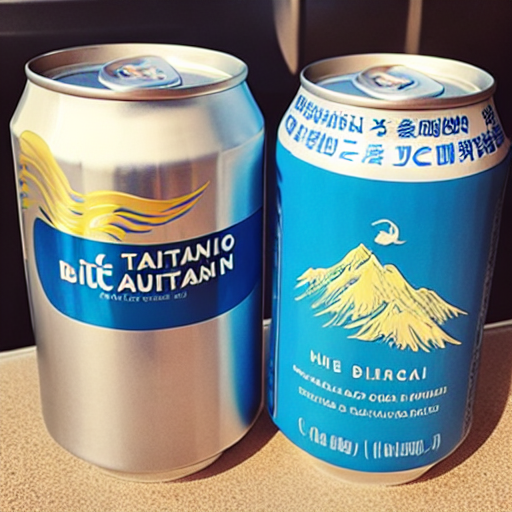} &
        \includegraphics[width=0.135\linewidth]{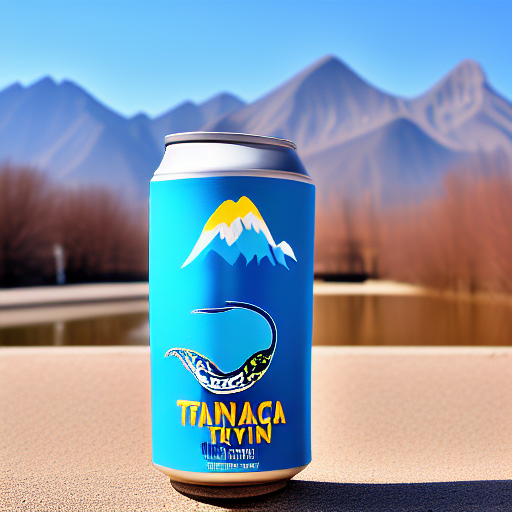} &
        \includegraphics[width=0.135\linewidth]{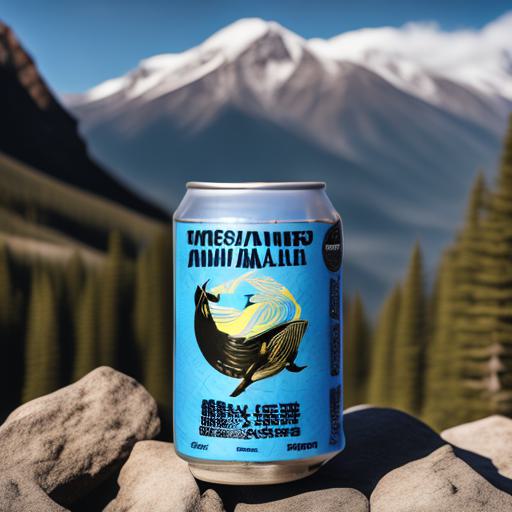} \\
        
        % Third sample prompts
        \multicolumn{1}{c}{\textit{``A toy''}} & &\multicolumn{3}{c}{\textit{``on the beach''}} &  \multicolumn{3}{c}{\textit{``on top of a white rug''}} \\
        % Third sample images
        \includegraphics[width=0.135\linewidth]{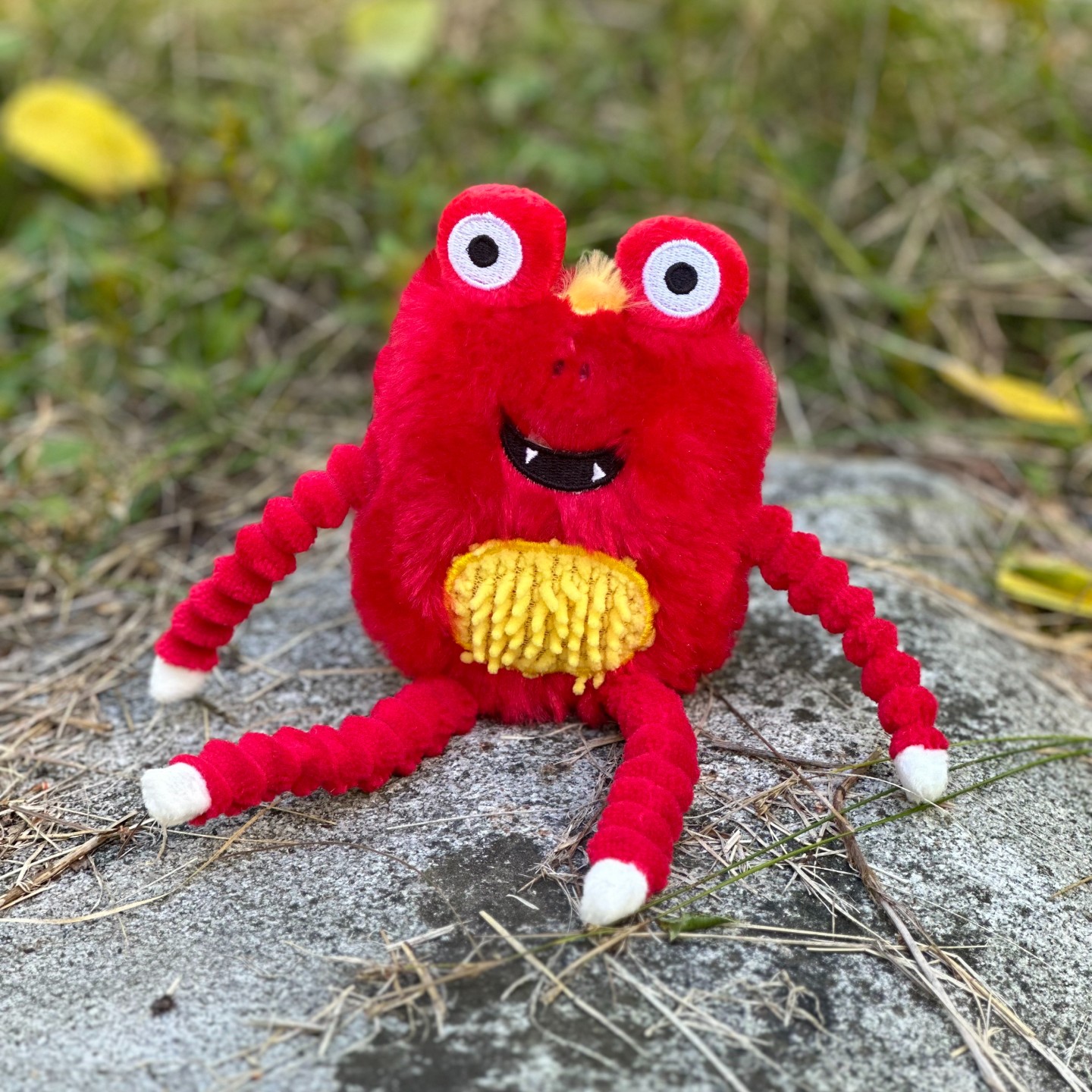} &
        &
        \includegraphics[width=0.135\linewidth]{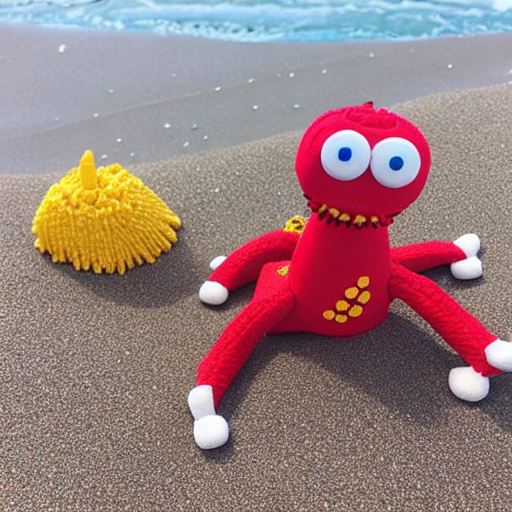} &
        \includegraphics[width=0.135\linewidth]{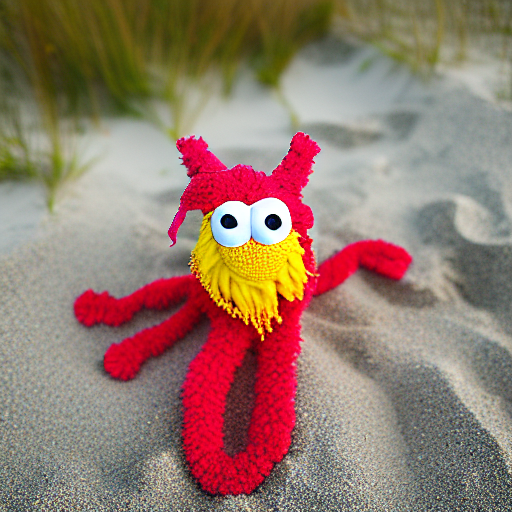} &
        \includegraphics[width=0.135\linewidth]{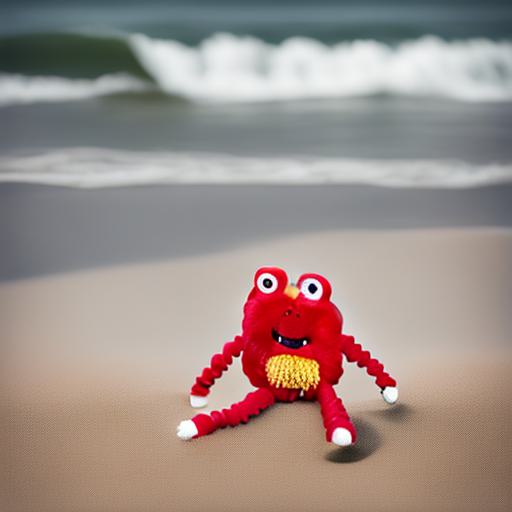} &
        \includegraphics[width=0.135\linewidth]{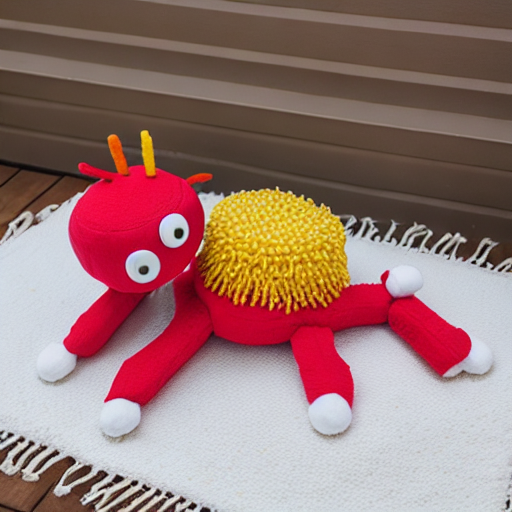} &
        \includegraphics[width=0.135\linewidth]{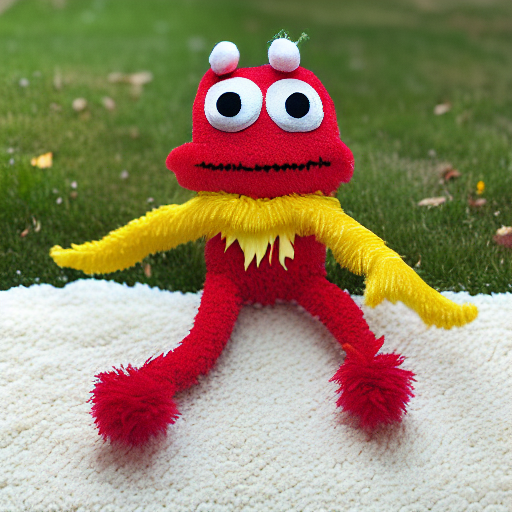} &
        \includegraphics[width=0.135\linewidth]{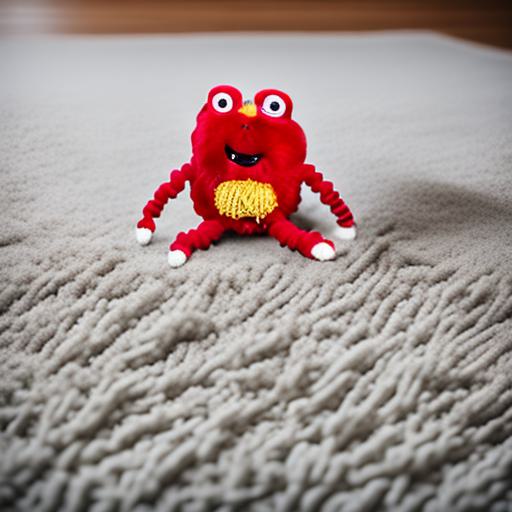} \\
        
        % Method names
        \textbf{Reference} & & \textbf{BLIP-D} & \textbf{Kosmos-G} & \textbf{\ourmethod\ } &  \textbf{BLIP-D} & \textbf{Kosmos-G} & \textbf{\ourmethod\ } \\
    \end{tabular}
    \caption{\textbf{Visual comparison.}  Personalized generations on sample concepts. \ourmethod\ preserves reference concept appearance and does not suffer from background interference. BLIP-D \cite{li2023blip} and Kosmos-G \cite{pan2023kosmos} cannot faithfully preserve visual details from the reference. }
    \label{fig:comparison}
\end{figure*}

\subsection{Zero-Shot Personalization}
We compare \ourmethod\ against state-of-the-art methods for finetuning-based and zero-shot personalization. Table~\ref{tab:dreambench} presents quantitative results, indicating the diffusion backbone and the number of reference images for each method. Our approach achieves competitive or superior performance compared to other computationally-intensive state-of-the-art methods, which are trained on larger datasets and with significantly more parameters. We refer the reader to Table~\ref{tab:previous_methods_overview} for the data requirements, training time, and parameter count of the various methods. We remark that generally DINO is a preferred metric for image similarity with respect to CLIP-I, as it is more sensitive to the appearance and fine-grained details of the subjects.

We also present qualitative comparisons with Kosmos-G~\cite{pan2023kosmos} and BLIP-Diffusion \cite{li2023blip}. We remark that several other methods are not reproducible due to the lack of code, datasets, or trained checkpoints.
As seen in Fig.~\ref{fig:comparison}, our method excels in subject preservation and textual alignment, producing visually superior results. We also notice that Kosmos-G reports a high CLIP-I score, but after inspecting the generated images, it is clear that the score does not entirely reflect the preservation of the reference subject in generated images. In fact, Kosmos-G presents a high degree of background interference, where the partial replication of the reference background boosts the alignment score. For this reason, we also report foreground-masked metrics on the subjects, like MCLIP-I and MDINO, in the \supplmat. 

%This is due to the phenomenon of background interference, which can increase the scores by generating similar backgrounds with respect to the reference images. 

\vspace{-3pt}

\subsection{Inference Time Evaluation}
The computational efficiency of our method is compared to the reference-based method BootPIG~\cite{purushwalkam2024bootpig} and encoder-based approaches such as Kosmos-G~\cite{pan2023kosmos} and Subject-Diffusion~\cite{ma2023subject}. 
%We assess the computational efficiency of our method by comparing it to reference-based method BootPIG~\cite{purushwalkam2024bootpig} and encoder-based approaches like Kosmos-G~\cite{pan2023kosmos} and Subject Diffusion~\cite{ma2023subject}. 
Table~\ref{tab:time_comparison} provides a detailed comparison of inference time, accounting for both personalization time (\eg, the time to generate the cache for \ourmethod) and the time to sample the personalized image. We also report the increase in model size, \ie, the storage (in FP16 precision) required for the extra parameters to allow for personalization, showing that \ourmethod\ is one order of magnitude more compact than the state of the art. Overall, \ourmethod\ offers a lightweight solution that achieves state-of-the-art performance with faster inference and reduced computational overhead.

%the sum of the time needed for personalization (e.g., the time to generate the cache for \ourmethod) and the time to sample the personalized image.

%aggiungere commento ai risultati

\begin{table}[t]
    \centering
    \setlength{\tabcolsep}{1.5pt}
    \caption{\textbf{Computational comparison.} *: time to generate an image with 100 timesteps, evaluated on a single NVIDIA A100 GPU.}
    \begin{tabular}{lccc}
            \toprule
        %&  MCLIP-I ($\uparrow$) & MDINO ($\uparrow$)    
        Method  &  Inference Time*  & Extra Params Size \\
          
            \midrule
            ELITE \cite{Wei2023ELITEEV} & 6.24 s &  914 MB\\

            BLIP-Diffusion \cite{li2023blip} & 3.92 s  & 760 MB
 \\

           % SubjectDiffusion\cite{ma2023subject} &   & 1.4GB  \\

            BootPig \cite{purushwalkam2024bootpig} & 7.55 s  & 1900 MB\\

            \textbf{\ourmethod\ (ours)} & 3.88 s  & 42 MB \\

            %ours without caching 6.18

            \bottomrule
        \end{tabular}
    
    \label{tab:time_comparison}
\end{table}

\subsection{Ablation Studies} \label{sec:ablations}
We validate our design choices through a series of studies, examining different conditioning mechanisms, evaluating our feature caching approach, and analyzing the impact of synthetic dataset scaling.
%We conduct several studies to validate our design choices. We first examine different conditioning mechanisms, followed by ablation of our feature caching approach and the impact of scaling the synthetic training dataset.

\paragraph{Reference Feature Integration}
We compare various conditioning strategies to integrate reference features in Table~\ref{tab:ref_integration}. 
Our spatial cross-attention block with concatenation between the output of self- and cross-attention (``Spatial Concat'') was evaluated against different alternatives, including IP-Adapter's conditioning mechanism \cite{ye2023ip} (``Textual Sum''), which sums the decoupled cross-attention output with that from textual cross-attention. 
We also tested a variant (``Spatial Sum'') where self- and cross-attention conditioning outputs are summed. 
% We also explored a restricted variant, ``Spatial Sum'', where the outputs of self-attention and cross-attention conditioning are summed. 
Additionally, we also assessed an alternative conditioning procedure inspired by ViCo~\cite{hao2023vico} (``Decoupled Blocks''), involving independent and interleaved cross-attention blocks.
Results in Table \ref{tab:ref_integration} indicate that the proposed ``Spatial Concat'' offers the best balance of text alignment and parameter efficiency.

\begin{table}[t]
    \centering
    \setlength{\tabcolsep}{3pt}
    \caption{\textbf{Reference feature integration.} \ourmethod\ uses the best tradeoff between accuracy and complexity.}
        \begin{tabular}{lccc}
            \toprule
        %&  MCLIP-I ($\uparrow$) & MDINO ($\uparrow$)    
        Method  &  CLIP-I ($\uparrow$) & CLIP-T ($\uparrow$)  & Params \\
          
            \midrule

            Textual Sum \cite{ye2023ip} &  0.788 & 0,282 & 19M \\

            Spatial Sum  &   0.812 & 0.293 & 16M \\

            Decoupled Blocks \cite{hao2023vico}  & 0.808 & 0.300 & 61M \\

            \textbf{Spatial Concat (ours)}    & 0.810 & 0.298 & 25M \\

            \bottomrule
        \end{tabular}
    
    \label{tab:ref_integration}
\end{table}

We further explored optimal conditioning insertion within the U-Net backbone in Table~\ref{tab:ref_position}, determining that applying conditioning (and therefore feature caching) at the middle layer and every second layer of the decoder achieved the best tradeoff between performance and parameter count. 
%Additionally, we explored where to insert the conditioning within the U-Net backbone, and, consequently, which features to cache. We tested different configurations to determine the optimal trade-off between performance and parameter count, and found that inserting the conditioning mechanism in the middle layer and every second layer of the decoder achieved the best results. 
% Results for each configuration are presented in Table~\ref{tab:ref_position}. 
%Additional details on th are in the \supplmat.

\begin{table}[t]
    \centering
    \setlength{\tabcolsep}{2pt}
    \caption{\textbf{Cache positioning} in the U-Net backbone offers a further tradeoff between accuracy and complexity.}
    \label{tab:ref_position}
        \begin{tabular}{cccccc}
            \toprule
            Encoder & Middle & Decoder & CLIP-I ($\uparrow$) & CLIP-T ($\uparrow$) & Params \\
            \midrule
            \cmark & \xmark & \xmark &   0.721  &  0.303  & 11M \\
            \cmark & \cmark & \xmark &   0.749  &  0.306  & 19M \\
            \xmark & \cmark & \xmark &   0.716  &  0.302  & 8M \\
            \xmark & \xmark & \cmark &   0.799  &  0.296  & 17M \\
            \xmark & \cmark & \cmark &   0.810  &  0.298  & 25M \\
            \cmark & \cmark & \cmark &   0.813  &  0.297  & 36M \\
            \bottomrule
        \end{tabular}
\end{table}

\begin{table}[t]
    \centering
    \setlength{\tabcolsep}{3pt}
    \caption{\textbf{Caching with text} is not influential and adds complexity.}
    %Ablation - Encoding with text caption}
        \begin{tabular}{ccc}
            \toprule
        %&  MCLIP-I ($\uparrow$) & MDINO ($\uparrow$)    
        Text-Free   &  CLIP-I ($\uparrow$) & CLIP-T ($\uparrow$)   \\
          
            \midrule

            \xmark &  0.811  & 0.295  \\

            \cmark  & 0.810 & 0.298 \\

            \bottomrule
        \end{tabular}
    
    \label{tab:abl_text}
\end{table}

\vspace{-14pt}
\paragraph{Text Input for Cached Features}
Our feature caching procedure is designed to be text-free, leveraging the classifier-free guidance used during pretraining where captions were occasionally omitted. We compare this approach with a version including textual inputs during caching (\eg, `\textit{A photo of ...}''). Table~\ref{tab:abl_text} shows that adding text conditioning slightly reduces text alignment while increasing complexity and potentially introducing noise in cases of inaccurate
%reference 
captions.
%We designed our feature caching procedure to be text-free, leveraging the classifier-free guidance used during the pretraining of diffusion models, where text captions were occasionally omitted. 
%We compared this text-free approach to a version that included textual conditioning during encoding in the form of a prompt like ``\textit{A photo of ...}''. 
%As shown in Table \ref{tab:abl_text}, the inclusion of text conditioning slightly reduced text alignment, while adding complexity and, potentially, introducing noise in cases of inaccurate reference captions.

\begin{table}[t]
    \centering
    \setlength{\tabcolsep}{3pt}
    \caption{\textbf{Dataset impact} for both synthetic and real data.}
        \begin{tabular}{lcc}
            \toprule
        %&  MCLIP-I ($\uparrow$) & MDINO ($\uparrow$)    
        Dataset   &  CLIP-I ($\uparrow$) & CLIP-T ($\uparrow$)   \\
          
            \midrule

            Synthetic-50K &  0.781 &  0.304  \\

            Synthetic-200K & 0.797 & 0,301 \\

            Synthetic-400K & 0.810 & 0.298 \\

            LAION-5M & 0.814  & 0.242 \\ 

%            LAION-5M + Synth400K ? & \\

            \bottomrule
        \end{tabular}
    
    \label{tab:abl_dataset}
    \vspace{-0.3cm}
\end{table}
\vspace{-13pt}
\paragraph{Dataset Impact}
We demonstrate the importance of our synthetic dataset to train the conditioning adapters and the effect of scaling its size. For this purpose, we created synthetic datasets according to the procedure in Sec.~\ref{sec:method:training} of sizes 50K, 200K, and 400K samples. 
We also tested the real-world 5M samples from the LAION~\cite{schuhmann2022laion} dataset, which, lacking target-caption-reference triplets, required reuse of target images as reference images too.
%While this could potentially provide more varied images, it does not provide target-caption-reference triplets as the synthetic datasets, and training requires resorting to the reuse of the target also as reference image. 
Table~\ref{tab:abl_dataset} shows that increasing dataset size improves image alignment, though slightly reduces textual alignment. Notably, LAION improves image alignment but struggles with textual alignment. This highlights the importance of triplet data (target image, reference image, and caption) for effective zero-shot personalization, ensuring both subject preservation and textual editability.
%shows that improvements in image alignment can be obtained for increasing dataset size, with marginal worsening of textual alignment. Notably, LAION can be provide good image alignment but fails at ensuring good textual alignment. 
% This reveals that using triplet data consisting of a target image, reference image, and caption is crucial for effective zero-shot image personalization. This ensures that the model does not merely learn to replicate the conditioning input, but instead maintains both subject-specific preservation and textual editability.

\subsection{Visualizing Reference Impact}

\begin{figure}[t]
\centering
\setlength{\tabcolsep}{0pt} 
\begin{tabular}{cc@{\hspace{0.2cm}}cc}
\includegraphics[width=0.115\textwidth]{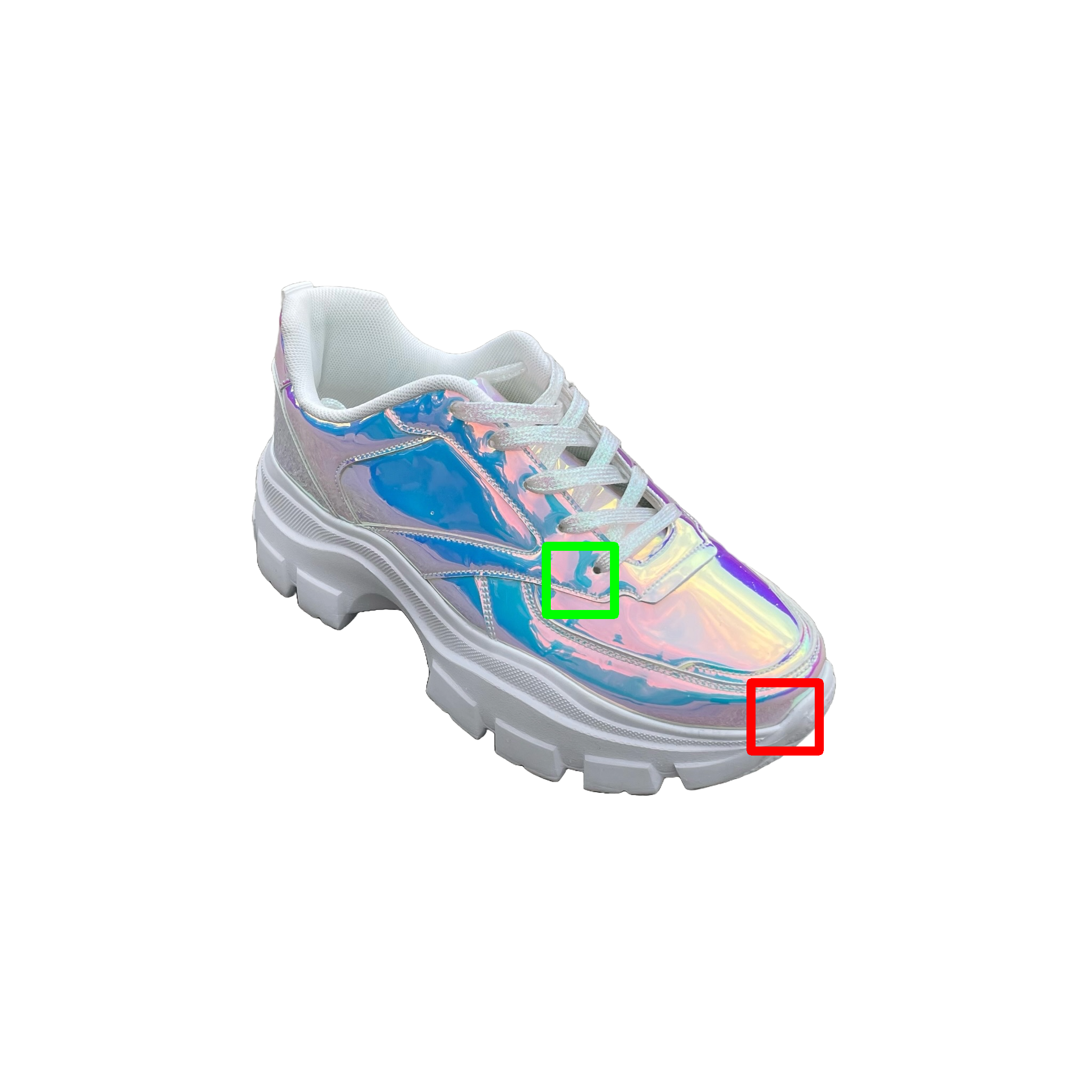} &
\includegraphics[width=0.115\textwidth]{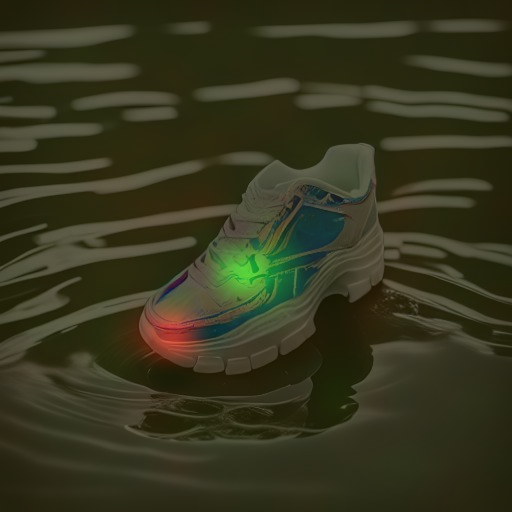} &
\includegraphics[width=0.115\textwidth]{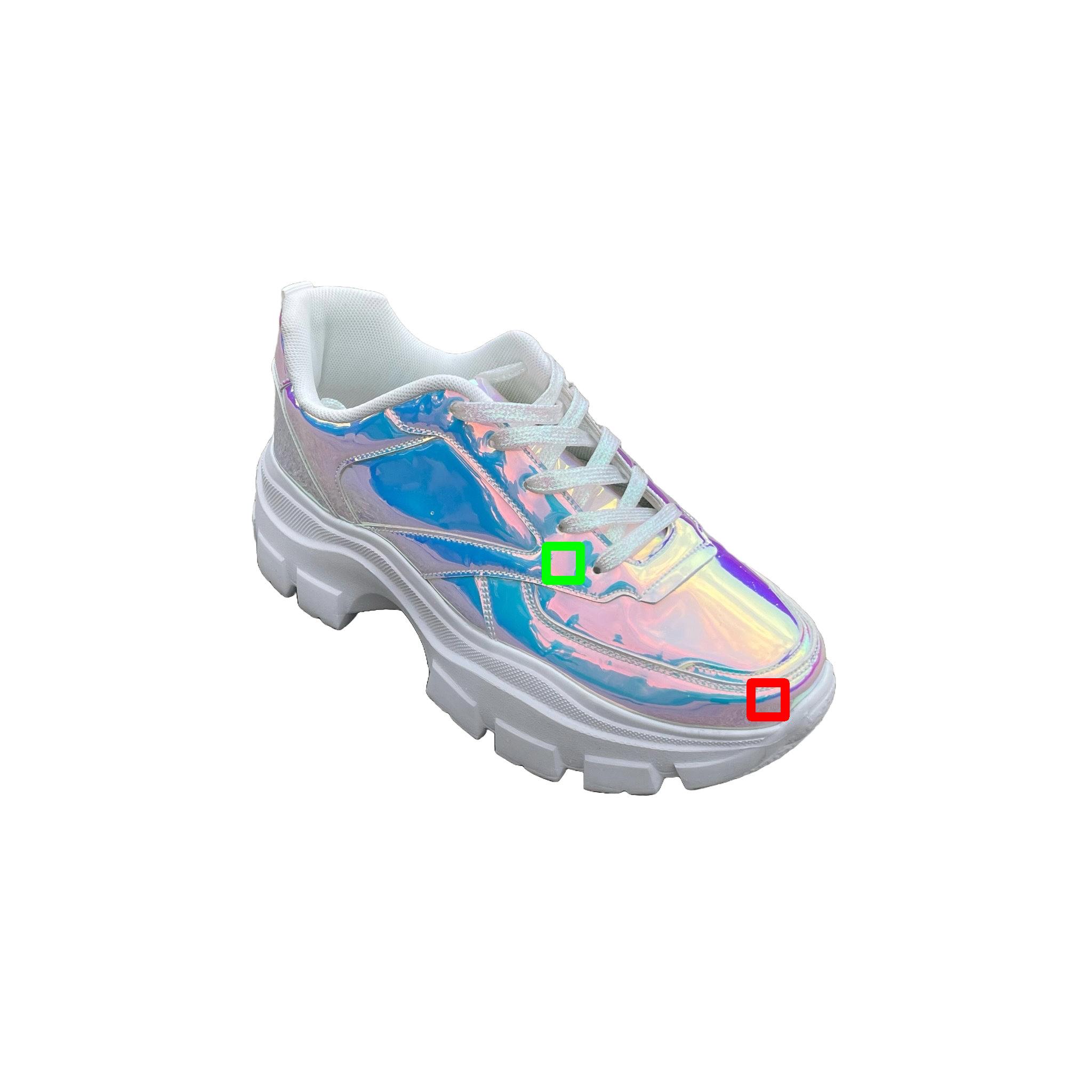} &
\includegraphics[width=0.115\textwidth]{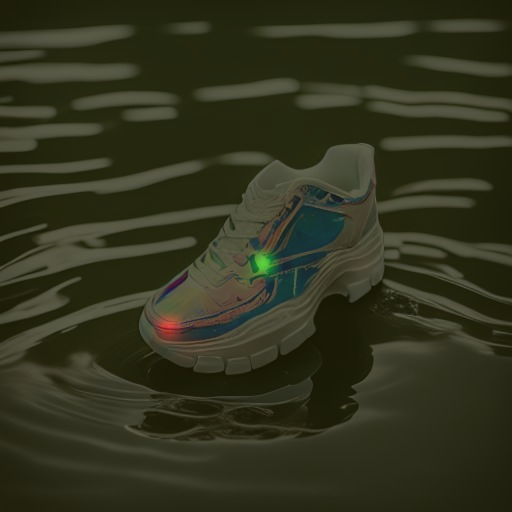} \\

\includegraphics[width=0.115\textwidth]{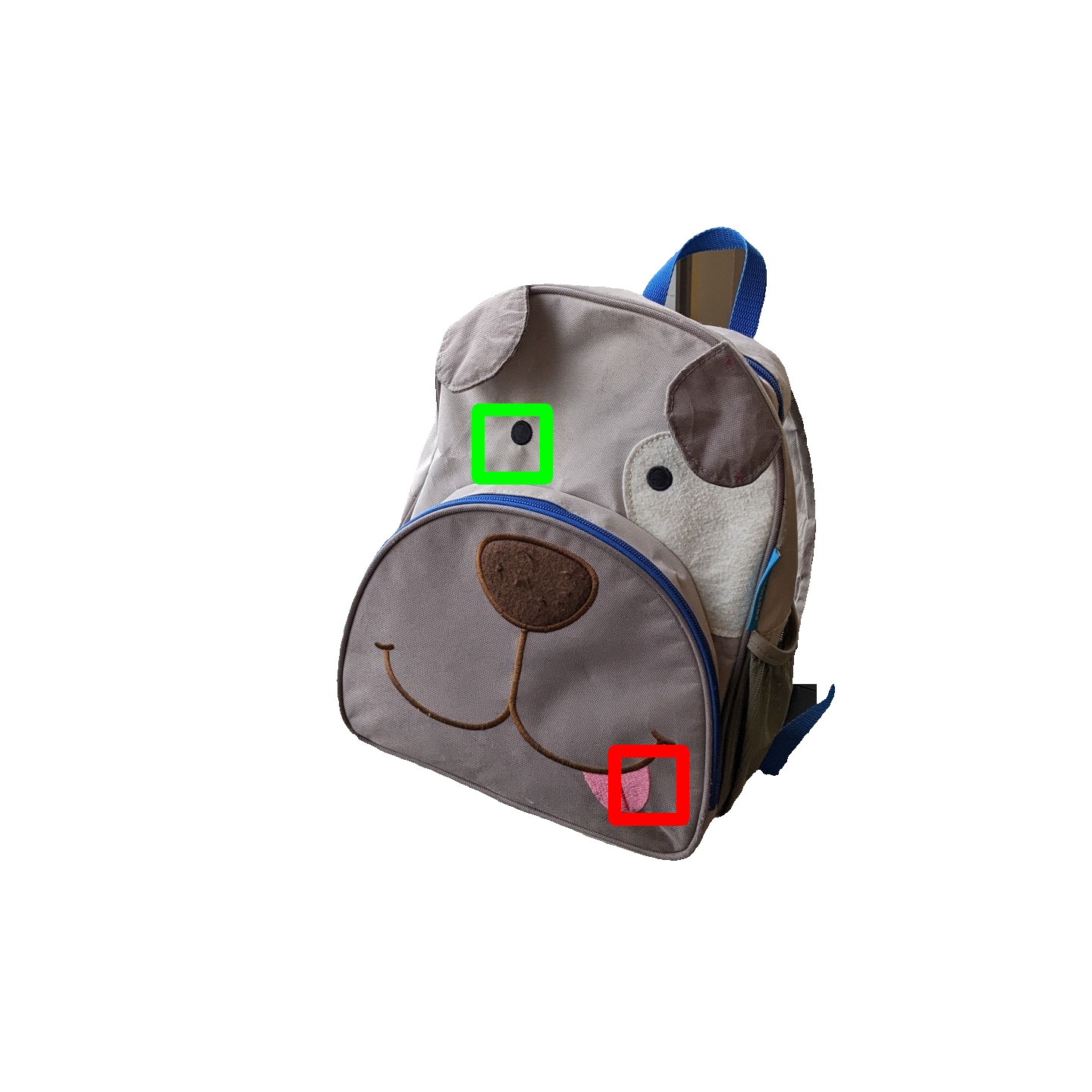} &
\includegraphics[width=0.115\textwidth]{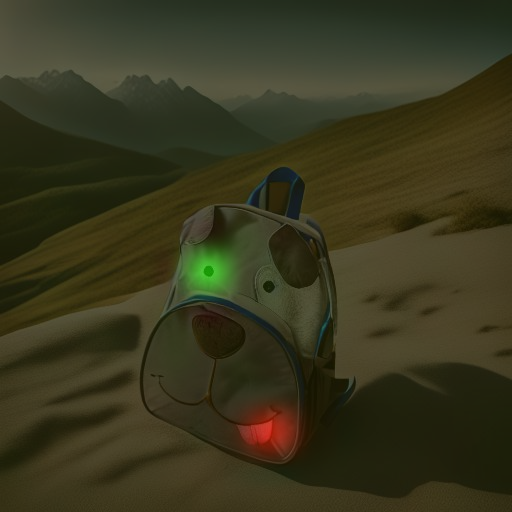} &
\includegraphics[width=0.115\textwidth]{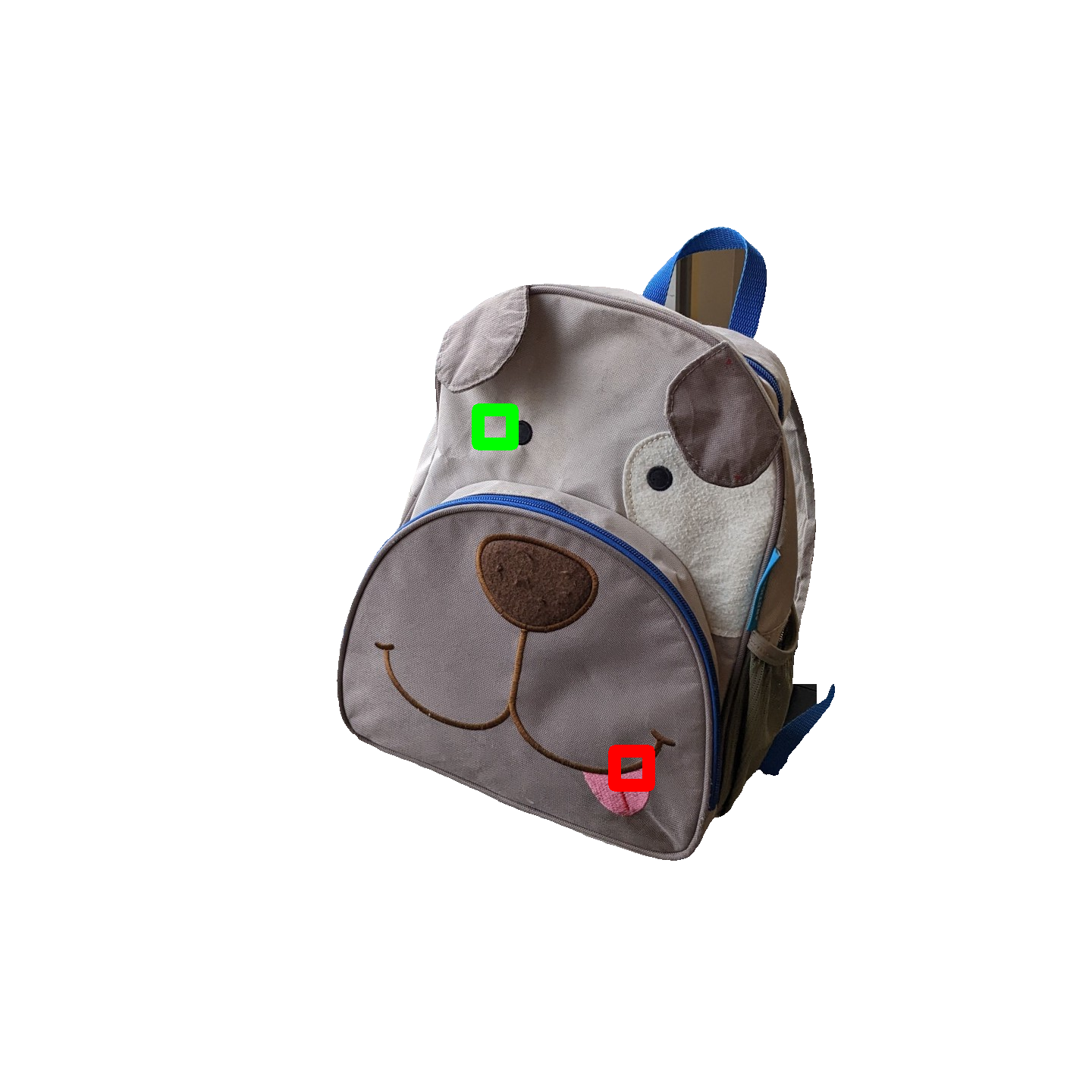} &
\includegraphics[width=0.115\textwidth]{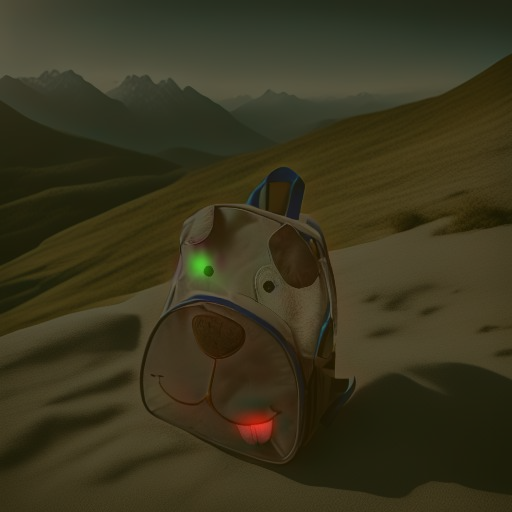} \\[2ex]
\end{tabular}
\setlength{\tabcolsep}{6pt} % Reset to default spacing
\caption{\textbf{Visualization of reference image impact.} Cross-attention maps between cached reference features and features of the image under generation. \textit{Left:} attention map at layers at $16\times 16$ resolution (left reference, right generated). \textit{Right:} $32 \times 32$. Attention values are highly localized in the region of interest.}
\label{fig:attn_maps}
\vspace{-0.3cm}
\end{figure}

Finally, we analyze how the cross-attention mechanism in \ourmethod\ impacts image generation by visualizing cached reference feature influence. 
Attention map visualizations at different resolutions are provided in Fig.~\ref{fig:attn_maps}.
Specifically, attention maps between the query from the current generation and the key derived from reference features reveal a highly localized focus on the subject, without interference from background elements. This mechanism models correspondences effectively, integrating reference information into the generated image. 

\vspace{-6pt}
\section{Discussion and Conclusions}
\vspace{-3pt} 
In this paper, we proposed \ourmethod, a novel approach to personalized text-to-image generation that uses feature caching to overcome the limitations of existing 
%encoder- and reference-based 
methods. By caching reference features from a small subset of layers of the U-Net only once, our method significantly reduces both computational and memory demands, enabling efficient, real-time personalized image generation. 
Unlike previous approaches, \ourmethod\ avoids the need for costly finetuning, external image encoders, or parallel reference processing, making it lightweight and suitable for plug-and-play deployment. 
Our experiments demonstrate that \ourmethod\ achieves state-of-the-art zero-shot personalization with only 25M additional parameters and a fast training process.
%, making it suited for mobile and resource-limited applications. 
% We plan to release our code, model, and synthetic dataset to promote reproducibility and support further research in this domain.
While \ourmethod\ is a promising direction towards more efficient personalized generation, it has some limitations. 
Although effective for single-subject personalization, our approach may require adaptation for complex multi-subject generation where feature interference can occur.
Additionally, certain edge cases, such as highly abstract or stylistic images, may challenge the caching mechanism's capacity to accurately preserve subject details. To address these challenges, future work may explore adaptive caching techniques or multi-reference feature integration. 
% for increased robustness
% across diverse scenarios.
% and further improve robustness across diverse scenarios. 

\newpage
{
    \small
    \bibliographystyle{ieeenat_fullname}
    \bibliography{main}
}

\clearpage

\renewcommand{\thefigure}{S\arabic{figure}}
\renewcommand{\theHfigure}{S\arabic{figure}}
\renewcommand{\thesection}{S\arabic{section}}
\renewcommand{\theHsection}{S\arabic{section}}
\renewcommand{\theequation}{S\arabic{equation}}
\renewcommand{\theHequation}{S\arabic{equation}}
\renewcommand{\thetable}{S\arabic{table}}
\renewcommand{\theHtable}{S\arabic{table}}

\setcounter{equation}{0}
\setcounter{figure}{0}
\setcounter{table}{0}
\setcounter{section}{0}
\setcounter{page}{1}

\maketitlesupplementary

\begin{figure*}[t]
    \centering
    \includegraphics[width=0.99\linewidth]{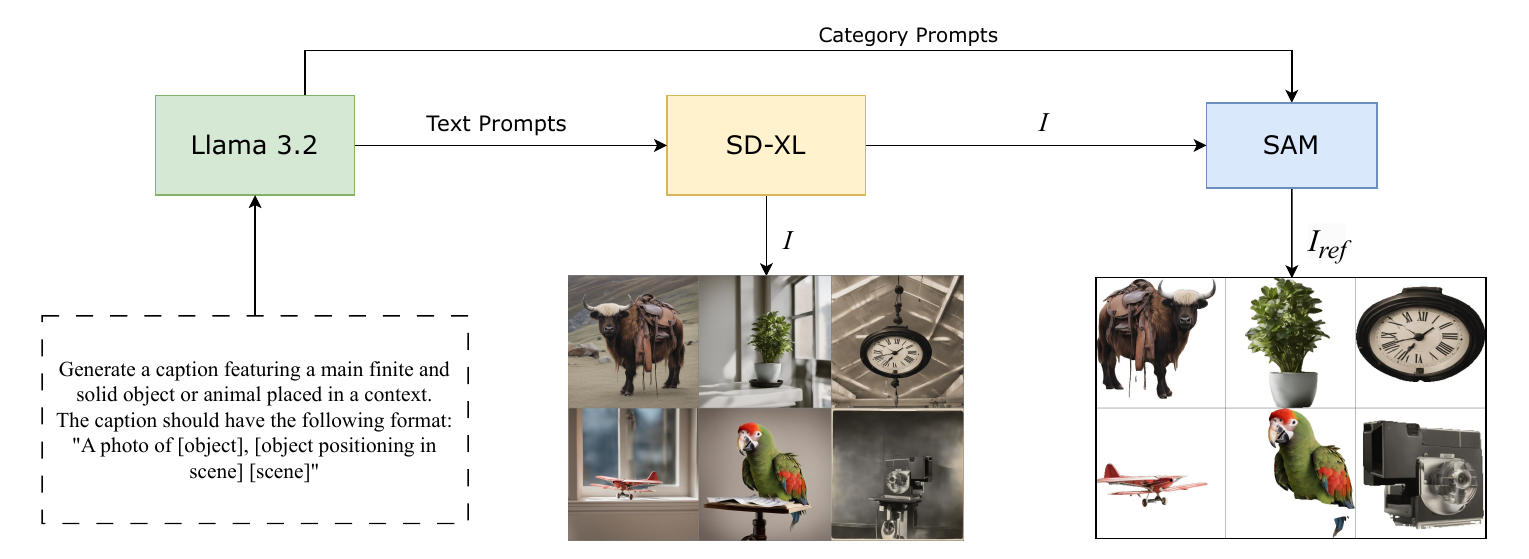}
    \caption{\textbf{Data synthesis pipeline} inspired by BootPIG \cite{purushwalkam2024bootpig}.}
    \label{fig:data_synthesis}
\end{figure*}

\begin{figure}[h!]
    \centering
    \includegraphics[width=0.99\linewidth]{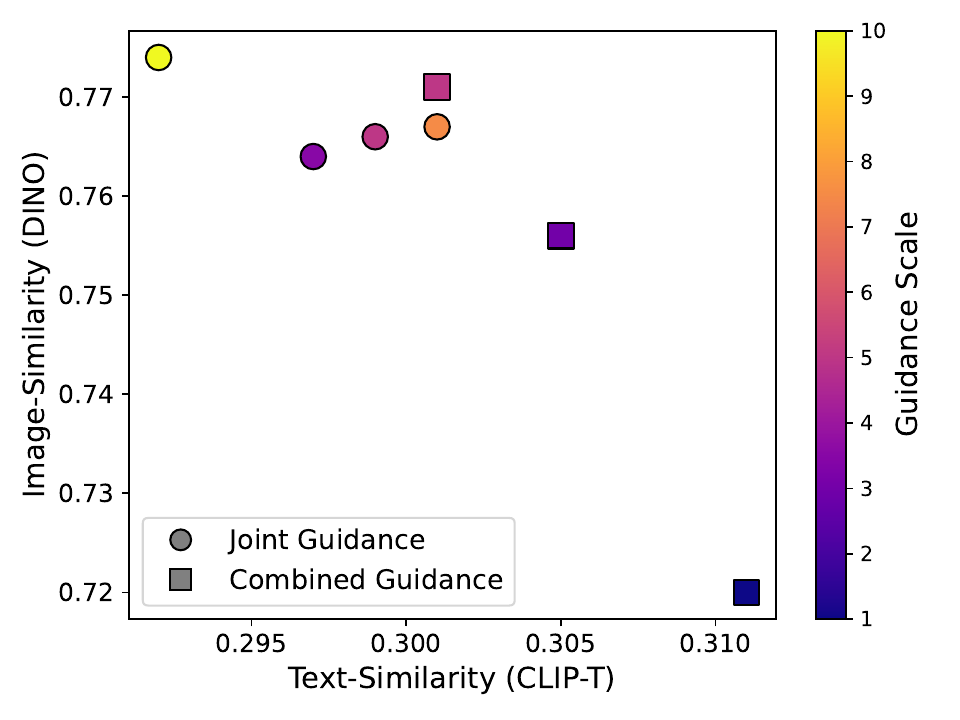}
    \caption{Sampling Space Exploration. For Combined Guidance, we leave the text scale $c_T=7.5$ and we vary the image scale $c_I$. }
    \label{fig:sampling_space}
\end{figure}

\section{Synthetic Dataset}
\label{sec:suppl:synthetic}

\begin{figure*}[h!]
    \centering
    \includegraphics[width=0.79\linewidth]{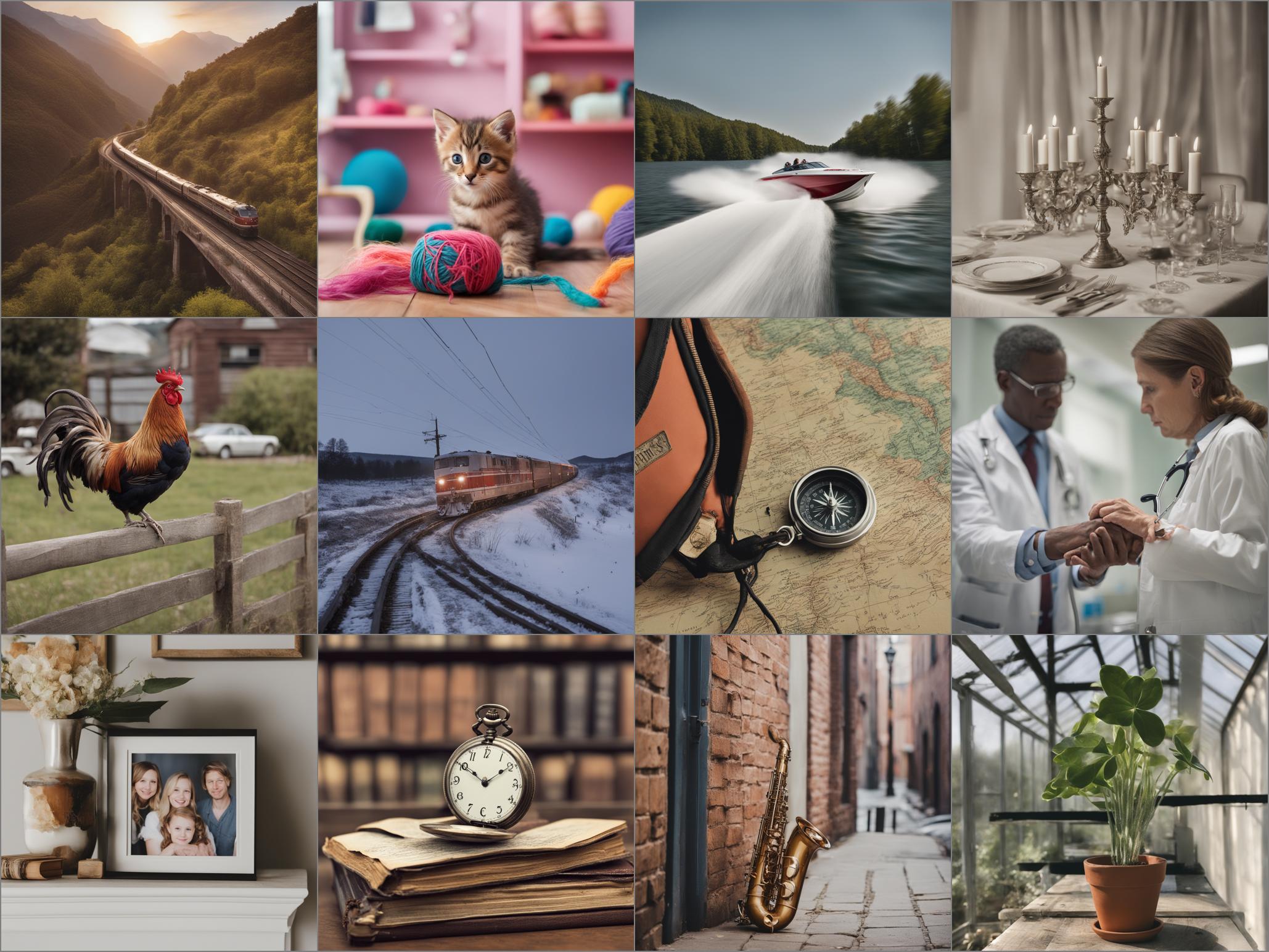}
    \includegraphics[width=0.79\linewidth]{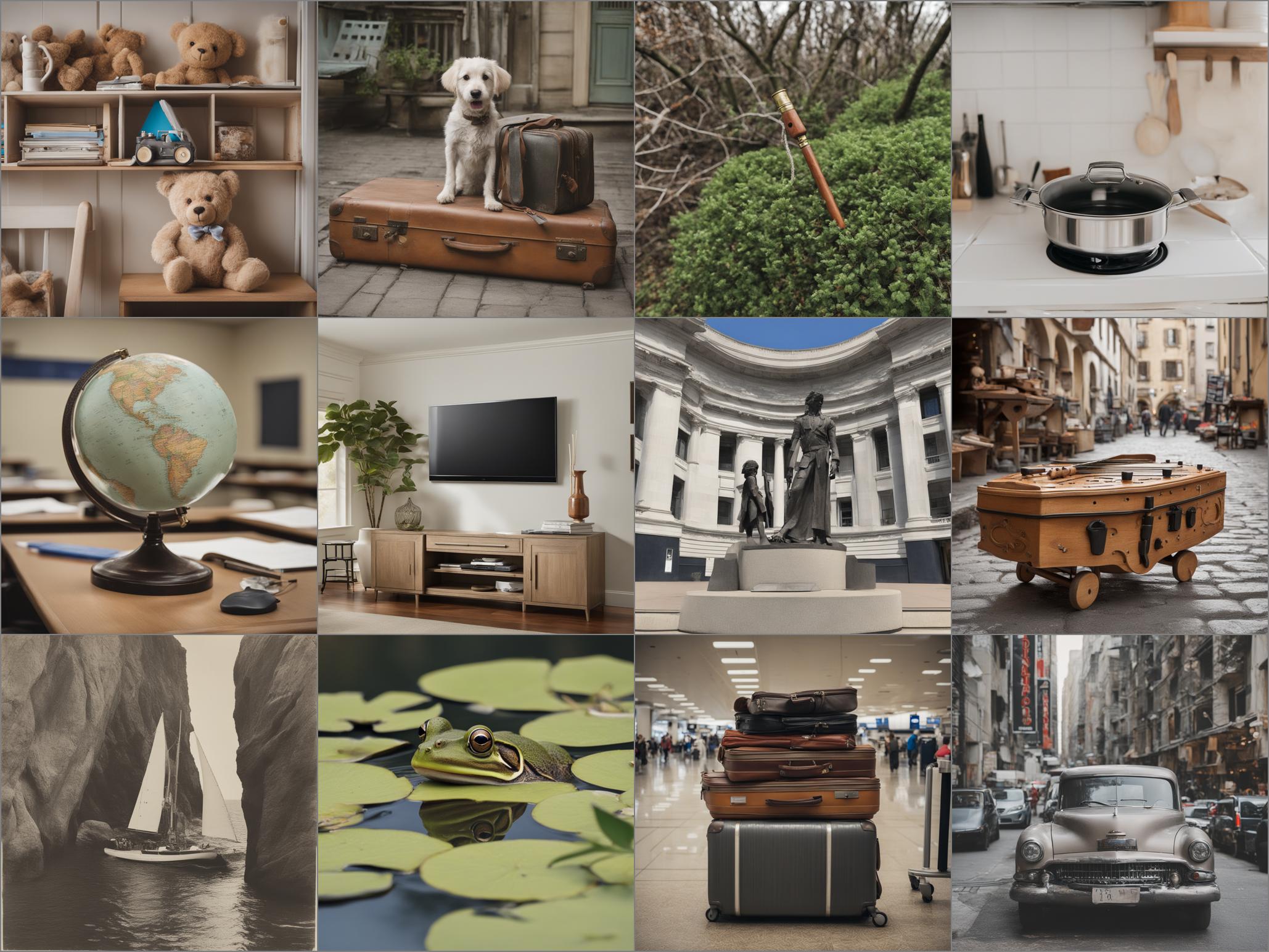}
    \caption{\textbf{Synthetic dataset samples} generated via the process outlined in Fig.~\ref{fig:data_synthesis}.}
    \label{fig:data_examples}
\end{figure*}

In this section, we describe our dataset generation pipeline, inspired by the success of BootPIG, with some modifications to ensure the pipeline adopts open-source models and is fully reproducible. Figure~\ref{fig:data_synthesis} provides an overview of the data creation process. We also show some examples of generated synthetic data in Figure~\ref{fig:data_examples}.

We use the \textit{lang-sam} pipeline\footnote{https://github.com/luca-medeiros/lang-segment-anything} to segment both generated and reference images based on textual conditioning, using a combination of Grounding-DINO and SAM. For caption generation, we leverage the LLama 3.2 8B \cite{dubey2024llama}, with a carefully crafted prompt that aims to generate diverse and descriptive captions of concrete objects, placing them in various meaningful contexts. We filter the generated captions to ensure the dataset's diversity and remove duplicates or highly similar captions. We write a simple filtering script that counts the number of occurrences for each object/category and filter out redundant captions.

The filtered captions are then used to prompt SD-XL \cite{podell2023sdxl} with a Classifier-Free Guidance (CFG) scale of $3.5$, employing $25$ denoising steps to generate the images. Our entire data generation pipeline is reproducible, and we plan to release it alongside the code for \ourmethod. Additionally, we will provide access to our generated dataset to encourage further research in this area.

\section{Additional Evaluations}
\label{sec:suppl:evals}

\begin{table}[t]
    \centering
    \setlength{\tabcolsep}{3pt}
    \caption{\textbf{Masked metrics} quantitative evaluation.}
        \begin{tabular}{lcccc}
            \toprule
            
        Method  &   MCLIP-I ($\uparrow$) & MDINO ($\uparrow$) \\

            \midrule

            DreamBooth & 0.868 & 0.712\\

            Custom Diffusion & 0.864 & 0.711 \\
            
            \midrule

            JeDI & 0.876 & 0.751 \\

            BLIP-D & 0.862 & 0.669\\

            ELITE & 0.861 & 0.681\\

            Toffee-5M & 0.874 & 0.803\\

            Ours & \textbf{0.906}  & \textbf{0.837} \\

            \bottomrule
        \end{tabular}
    
    \label{tab:eval_masked}
\end{table}

\subsection{Masked Metrics} Recent studies \cite{zeng2024jedi, zhou2024toffee} emphasize the value of evaluating masked versions of image similarity metrics to eliminate potential interference from background elements, thus ensuring the evaluation focuses on the fidelity of the personalized object. We use Grounded-SAM \cite{ren2024grounded} to segment both generated and reference images, subsequently computing the CLIP-I and DINO scores for these segments. The results for these masked metrics are reported in Table \ref{tab:eval_masked}. \ourmethod\ achieves a higher score on both metrics, demonstrating its superiority in subject preservation.

\subsection{Qualitative Results} 
\begin{figure*}[htbp]
    \centering
    \setlength{\tabcolsep}{1.5pt} % Adjust padding between columns if needed
    \begin{tabular}{c c c c c}
        % Column prompts (First row)
        \textit{``A dragon...''} & \textit{``flying in the sky''} & \textit{``in a flower garden''} & \textit{``frozen''} & \textit{``chinese painting''} \\
        
        % First sample images
        \includegraphics[width=0.19\linewidth]{figures/flux_dragon.png} &
        \includegraphics[width=0.19\linewidth]{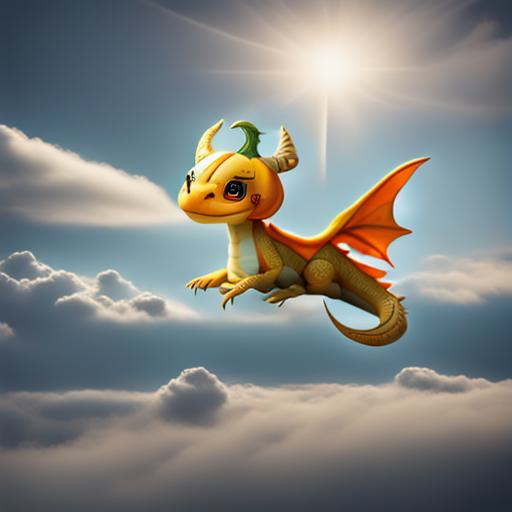} &
        \includegraphics[width=0.19\linewidth]{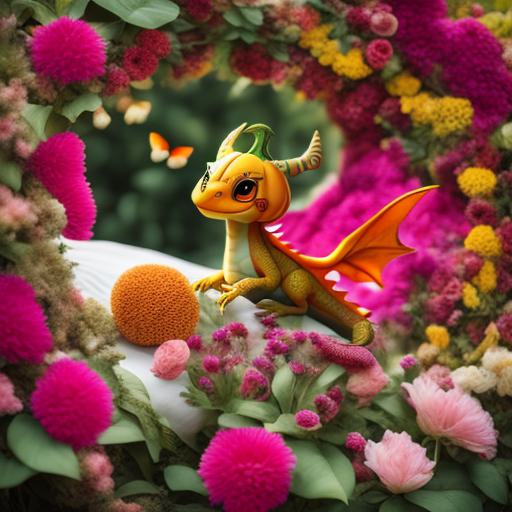} &
        \includegraphics[width=0.19\linewidth]{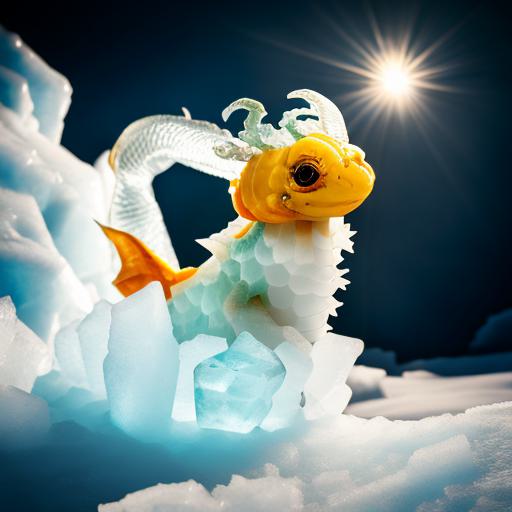} &
        \includegraphics[width=0.19\linewidth]{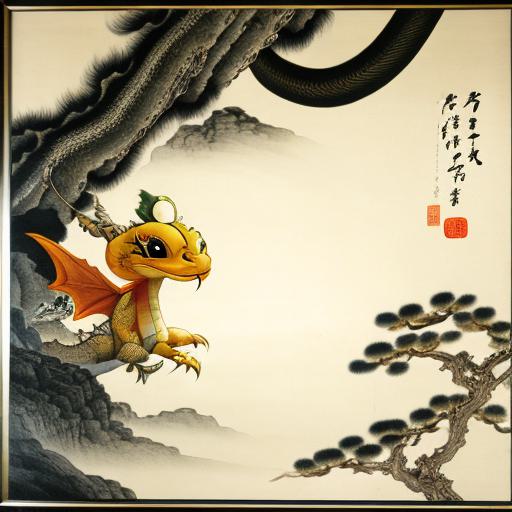} \\

        \vspace{3pt}
        
        % Column prompts (Second row)
        \textit{``An elephant...''} & \textit{``in minecraft''} & \textit{``as street graffiti''} & \textit{``dressed as a wizard''} & \textit{``as a plushie''} \\
        
        % Second sample images
        \includegraphics[width=0.19\linewidth]{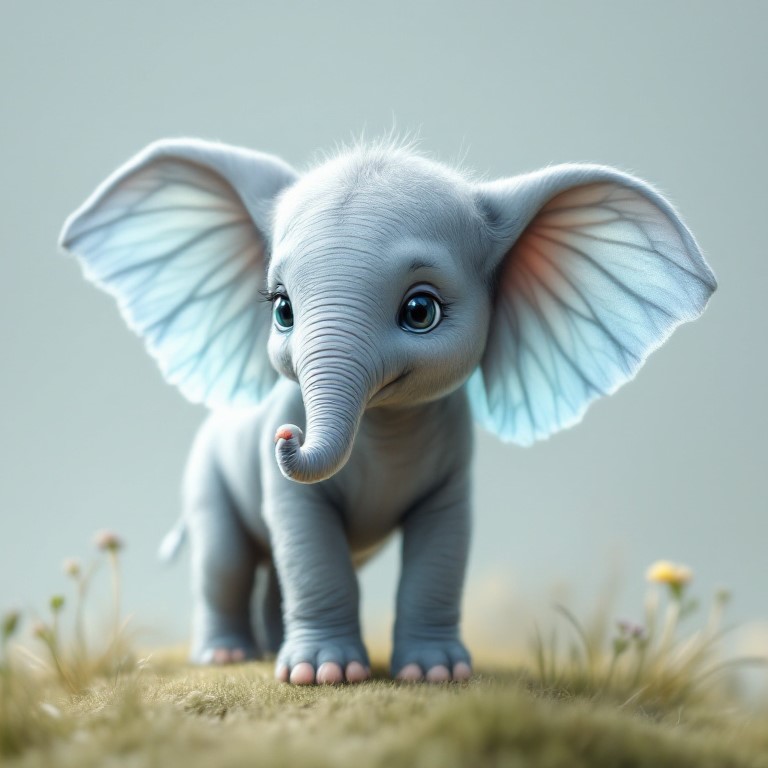} &
        \includegraphics[width=0.19\linewidth]{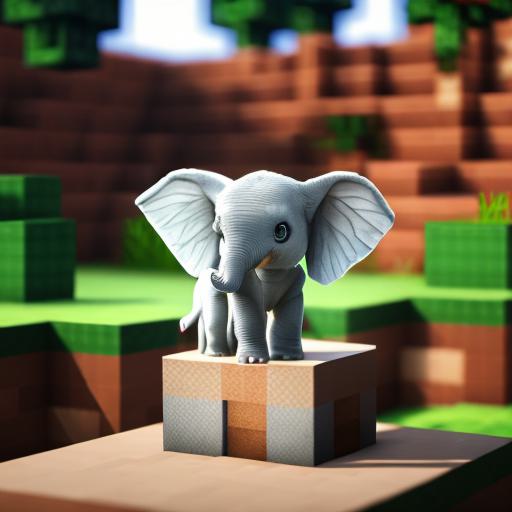} &
        \includegraphics[width=0.19\linewidth]{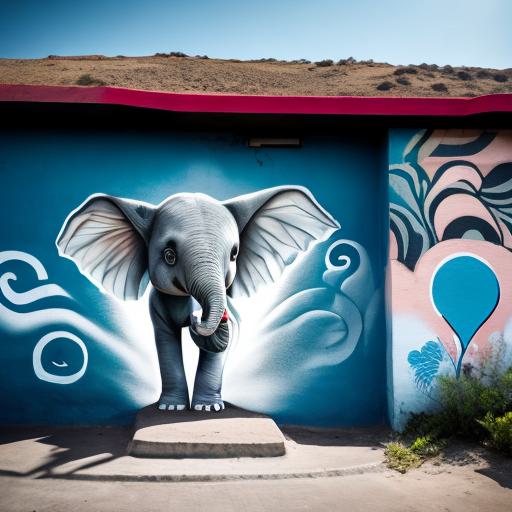} &
        \includegraphics[width=0.19\linewidth]{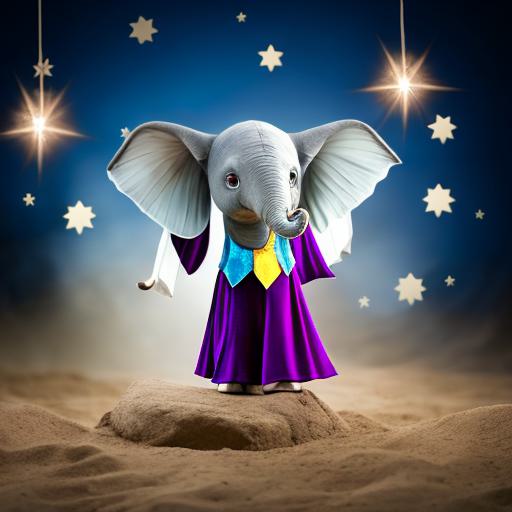} &
        \includegraphics[width=0.19\linewidth]{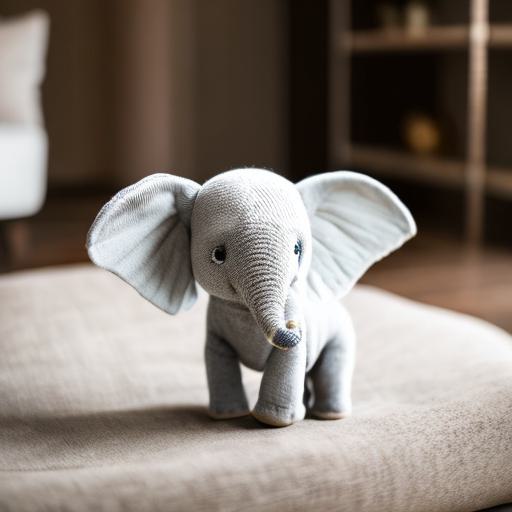} \\
        \vspace{3pt}
        % Column prompts (Second row)
% Column prompts (Second row)
        \textit{``An astrounaut...''} & \textit{``in a snow ball''} & \textit{``in boiling water''} & \textit{``on the mountains''} & \textit{``having breakfast''} \\
        
        % Second sample images
        \includegraphics[width=0.19\linewidth]{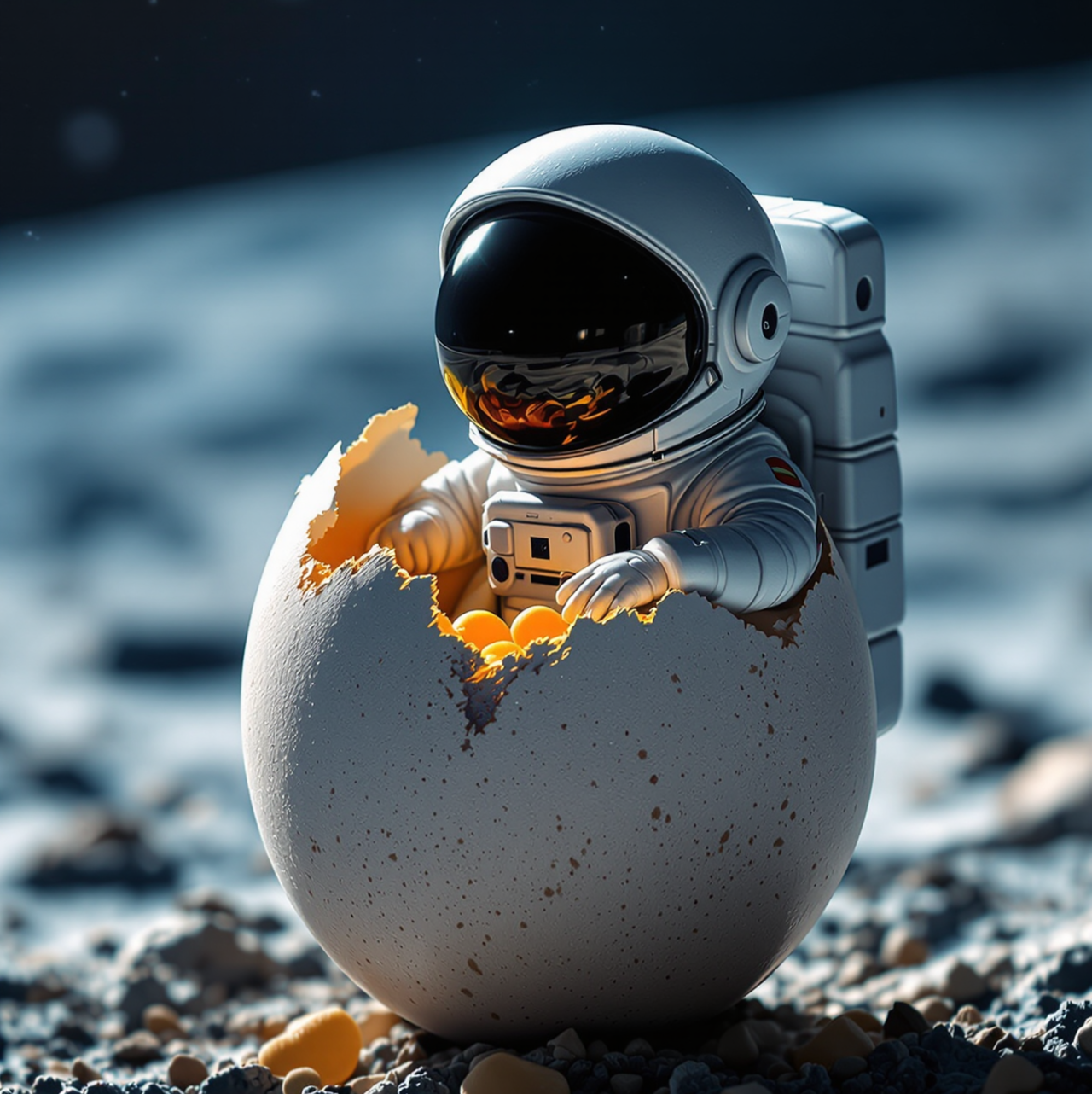} &
        \includegraphics[width=0.19\linewidth]{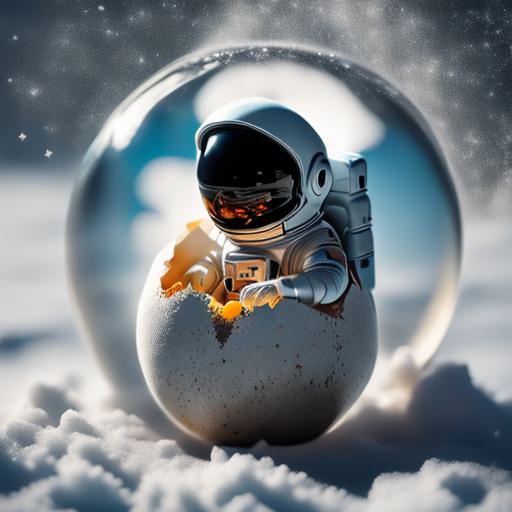} &
        \includegraphics[width=0.19\linewidth]{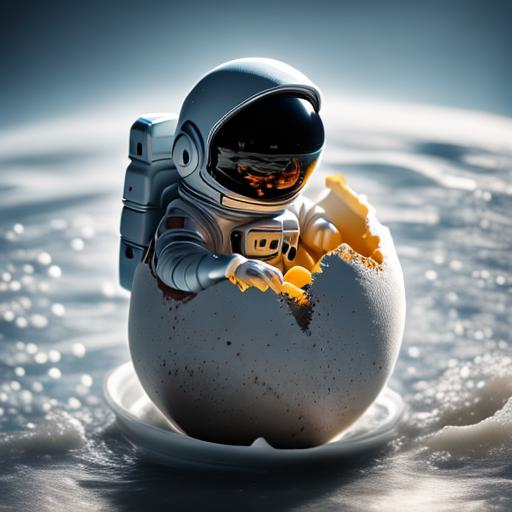} &
        \includegraphics[width=0.19\linewidth]{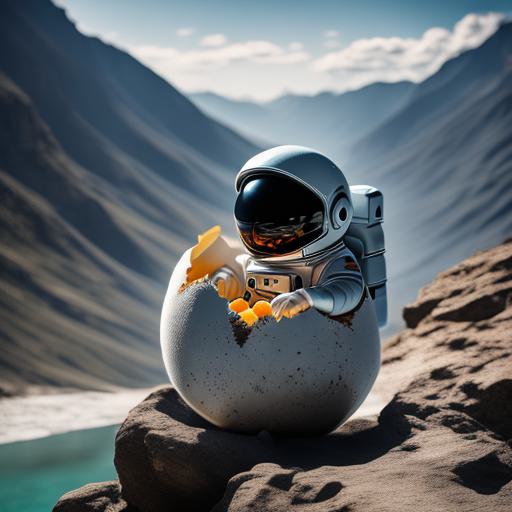} &
        \includegraphics[width=0.19\linewidth]{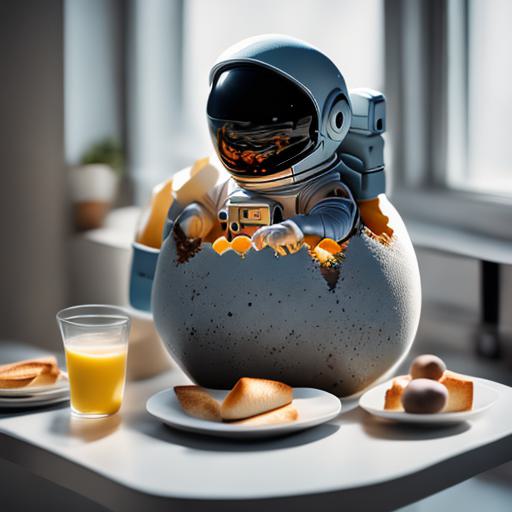} \\

         % Column prompts (Second row)
        \textit{``A chair...''} & \textit{``in a rustic cabinet''} & \textit{``royal throne''} & \textit{``futuristic setting''} & \textit{``Van Gogh painting''} \\

        % Second sample images
        \includegraphics[width=0.19\linewidth]{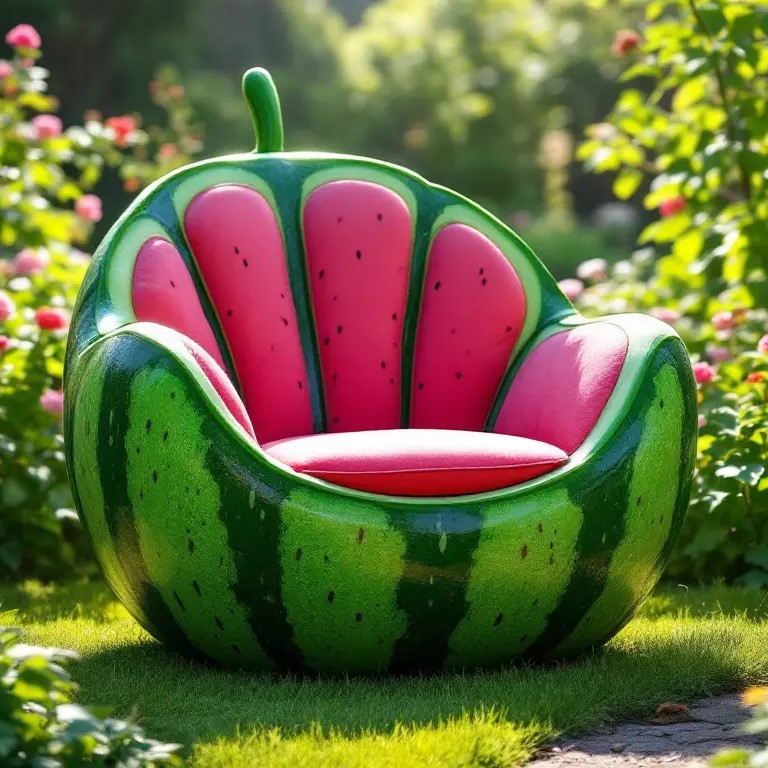}&
        \includegraphics[width=0.19\linewidth]{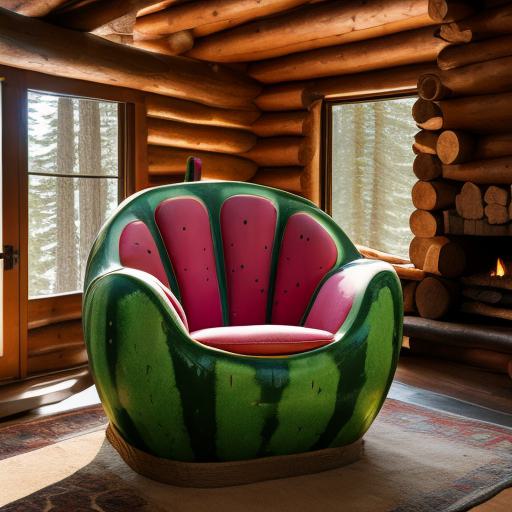} &
        \includegraphics[width=0.19\linewidth]{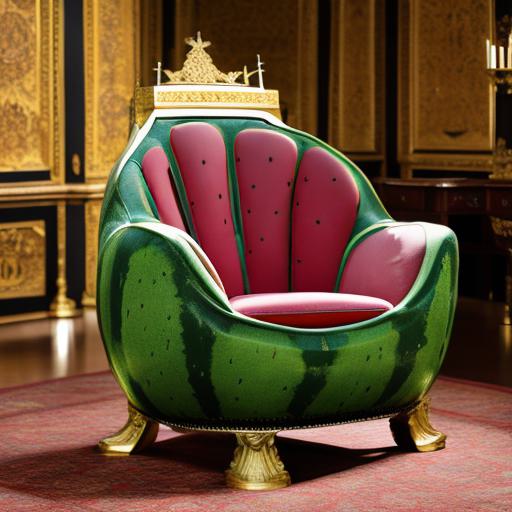} &
        \includegraphics[width=0.19\linewidth]{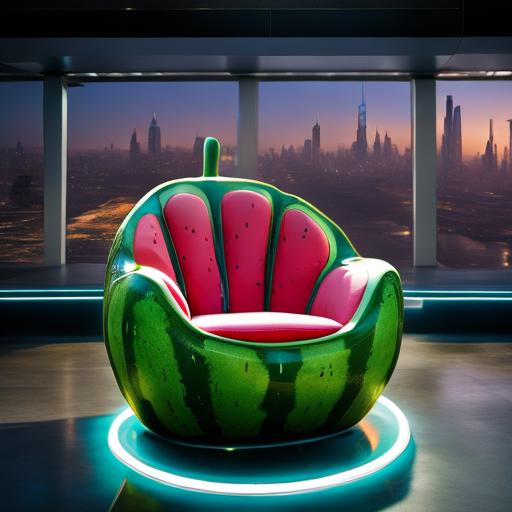} &
        \includegraphics[width=0.19\linewidth]{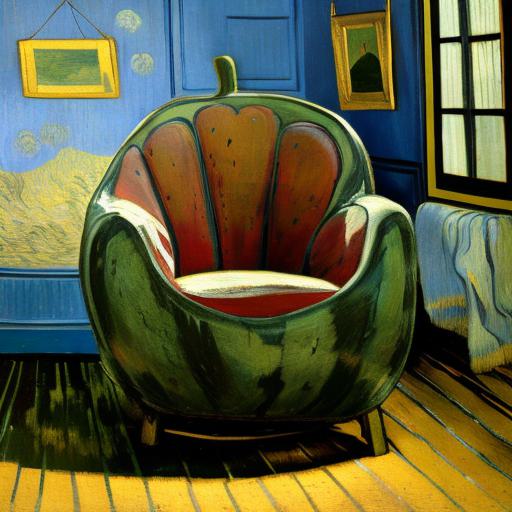} 
         \\
        \vspace{3pt}
        % Column prompts (Second row)
        \textit{``A guitar...''} & \textit{``in a snow globe''} & \textit{``made of ice''} & \textit{``Monet painting''} & \textit{``underwater''} \\
        
        % Second sample images
        \includegraphics[width=0.19\linewidth]{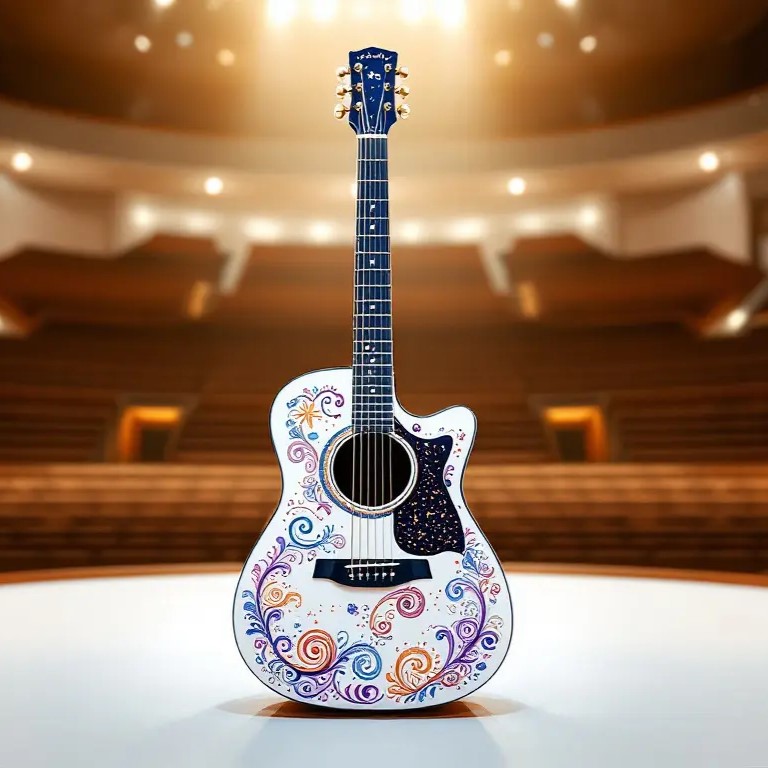} &
        \includegraphics[width=0.19\linewidth]{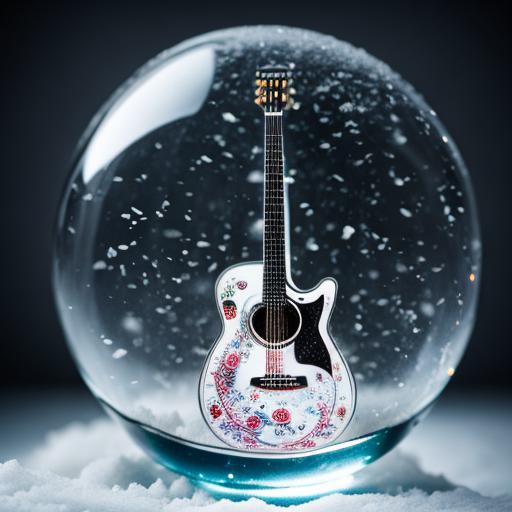} &
        \includegraphics[width=0.19\linewidth]{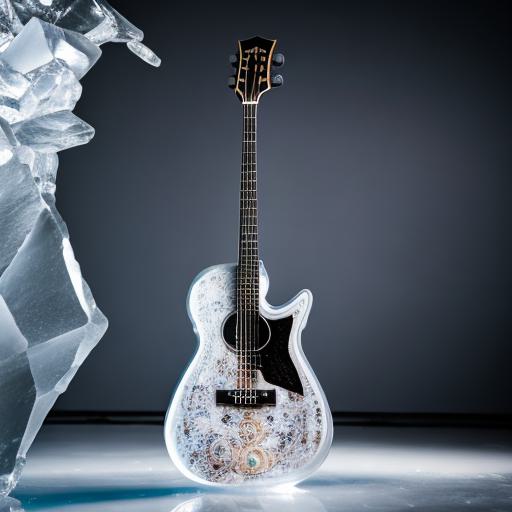} &
        \includegraphics[width=0.19\linewidth]{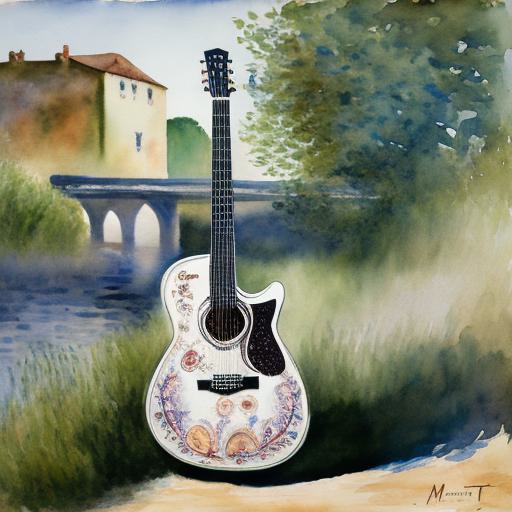} &
        \includegraphics[width=0.19\linewidth]{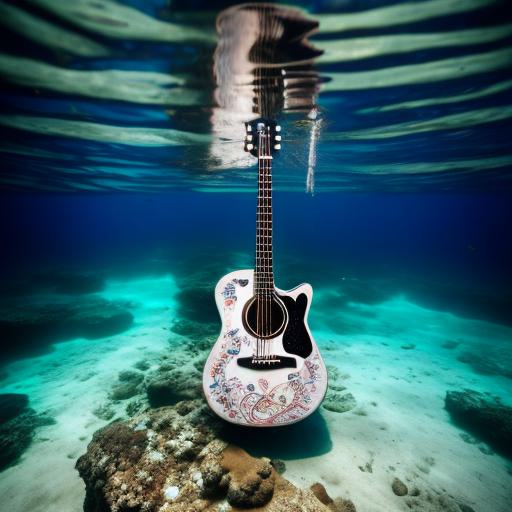} \\
        \vspace{3pt}
    \end{tabular}
    \caption{\textbf{Personalized generations by \ourmethod.} The proposed method is able to adapt to different text prompts and leverage diffusion prior to perform appearance and style editing of the personalized content. We also notice how the background interference is completely absent in generated images due to our design choice of caching masked reference features.}
    \label{fig:qualitative_supp}
\end{figure*}

In this section, we present additional qualitative generations produced by \ourmethod\ (Figure \ref{fig:qualitative_supp}). We conduct experiments using both synthetically generated subjects and real subjects from the Dreambooth dataset. Our results demonstrate that our method effectively follows complex text prompts. Interestingly, despite the absence of explicit training for subject modification (as seen in editing datasets), our approach successfully adapts and transforms the input subject in various contexts, rather than simply replicating the reference.

\begin{figure*}[htbp]
    \centering
    \setlength{\tabcolsep}{1.5pt} % Adjust padding between columns if needed
    \begin{tabular}{c c ccc ccc}
        % First sample prompts
        \multicolumn{1}{c}{\textit{``A backpack''}} & &\multicolumn{3}{c}{\textit{``in an ocean of milk''}} &  \multicolumn{3}{c}{\textit{``in the jungle''}} \\
        
        % First sample images
        \includegraphics[width=0.135\linewidth]{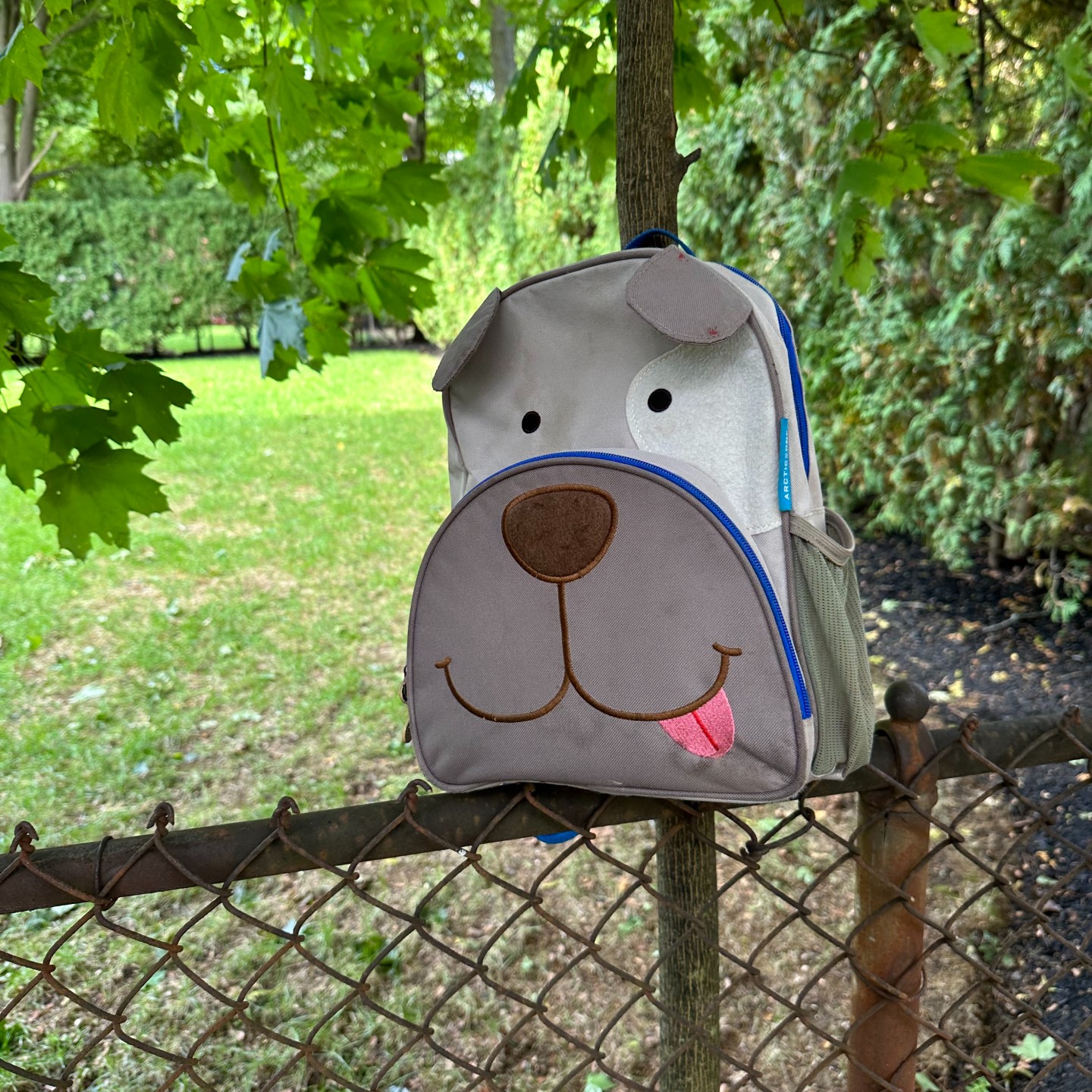} &
        &
        \includegraphics[width=0.135\linewidth]{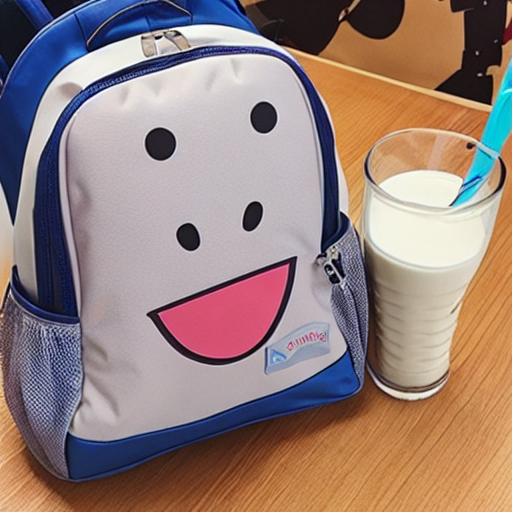} &
        \includegraphics[width=0.135\linewidth]{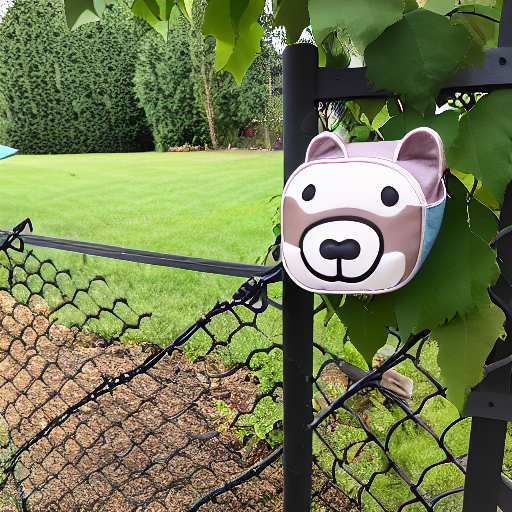} &
        \includegraphics[width=0.135\linewidth]{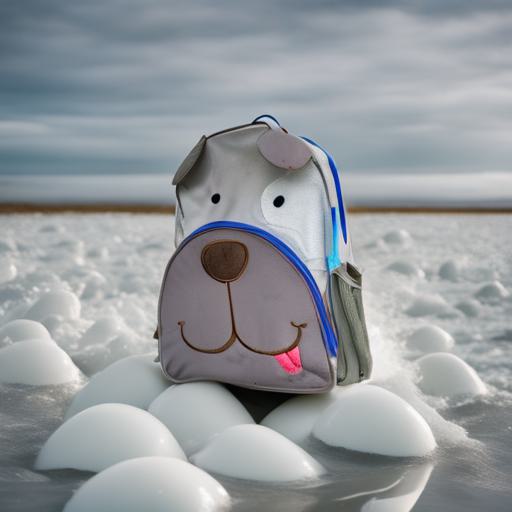} &
        \includegraphics[width=0.135\linewidth]{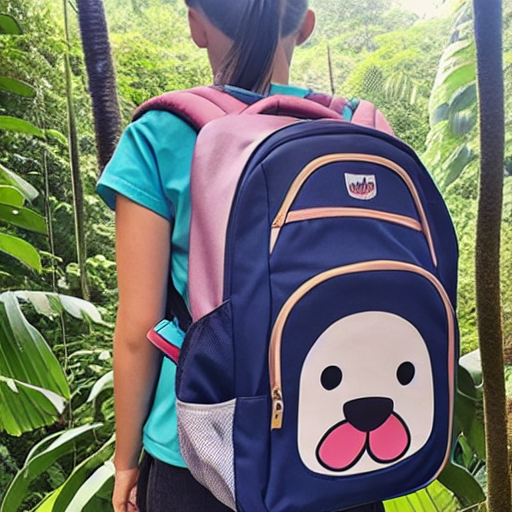} &
        \includegraphics[width=0.135\linewidth]{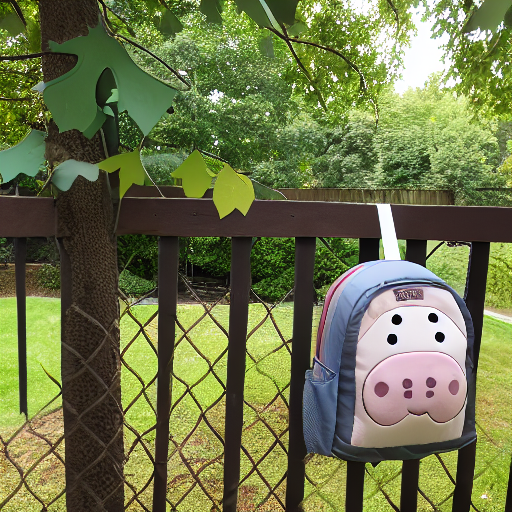} &
        \includegraphics[width=0.135\linewidth]{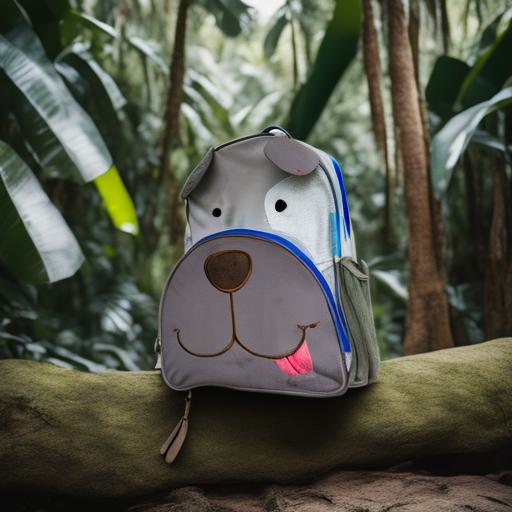} \\

        % Second sample prompts
        \multicolumn{1}{c}{\textit{``A toy''}} & & \multicolumn{3}{c}{\textit{``with a tree and autumn leaves''}} &  \multicolumn{3}{c}{\textit{``floating on top of water''}} \\
        % Second sample images
        \includegraphics[width=0.135\linewidth]{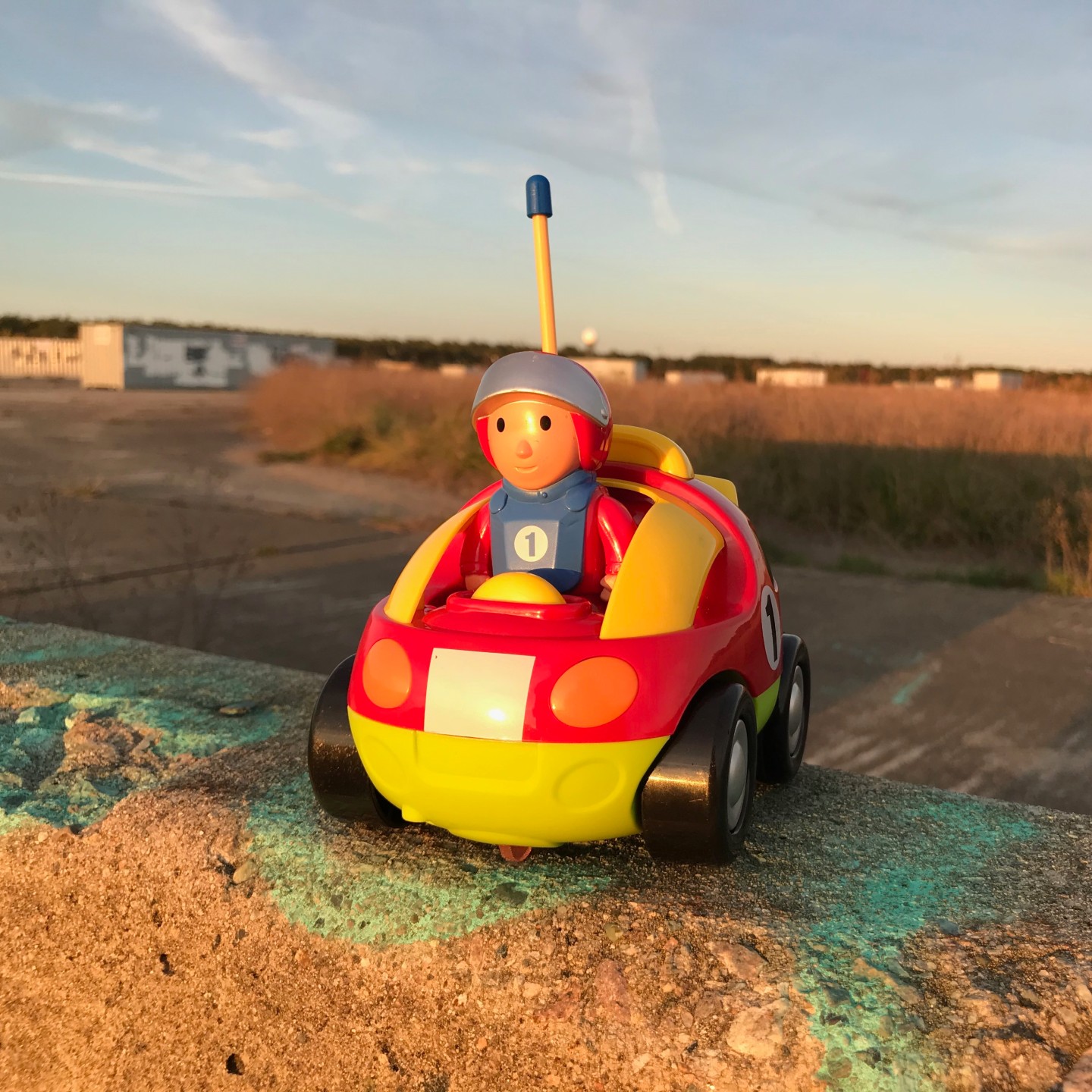} &
        &
        \includegraphics[width=0.135\linewidth]{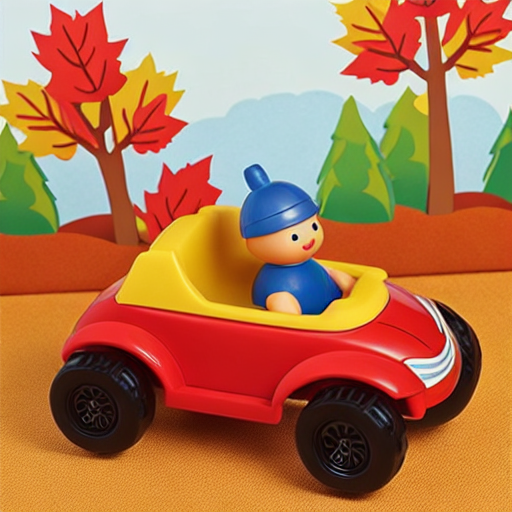} &
        \includegraphics[width=0.135\linewidth]{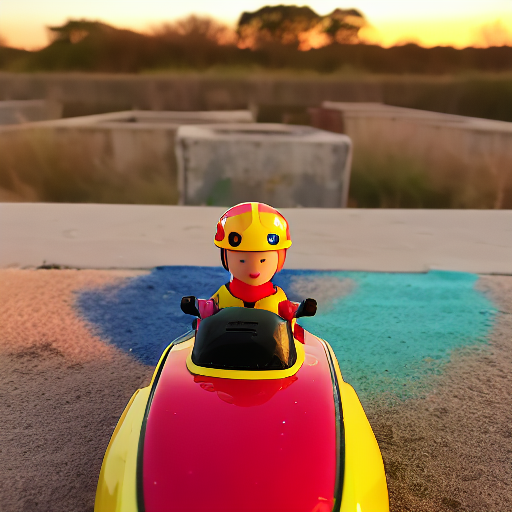} &
        \includegraphics[width=0.135\linewidth]{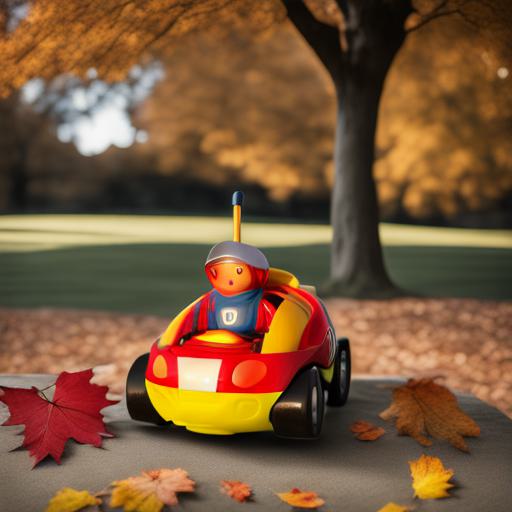} &
        \includegraphics[width=0.135\linewidth]{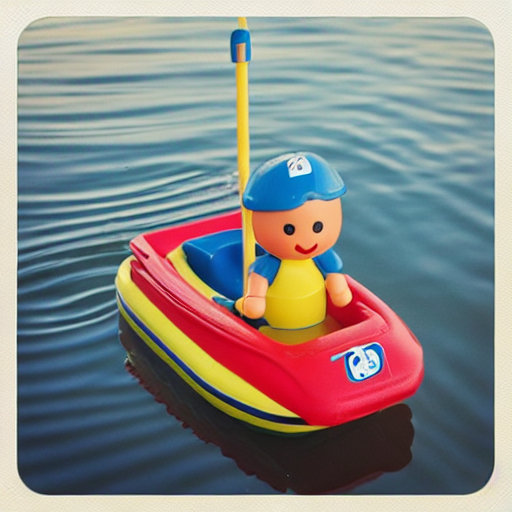} &
        \includegraphics[width=0.135\linewidth]{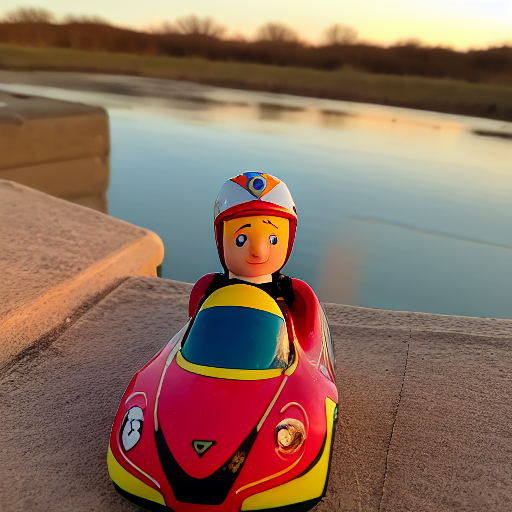} &
        \includegraphics[width=0.135\linewidth]{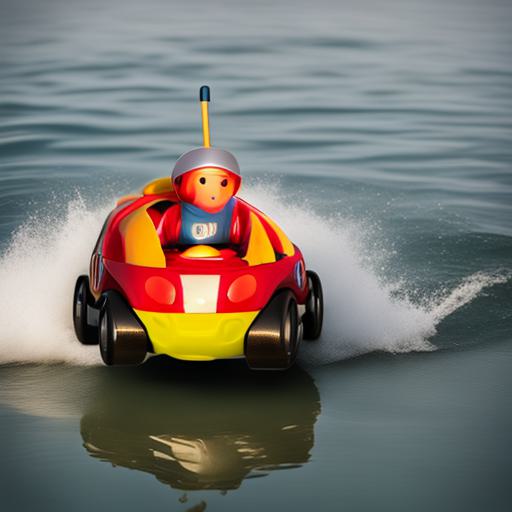} \\
        
        % Third sample prompts
        \multicolumn{1}{c}{\textit{``A sneaker''}} & &\multicolumn{3}{c}{\textit{``red''}} &  \multicolumn{3}{c}{\textit{``on top of a mirror''}} \\
        % Third sample images
        \includegraphics[width=0.135\linewidth]{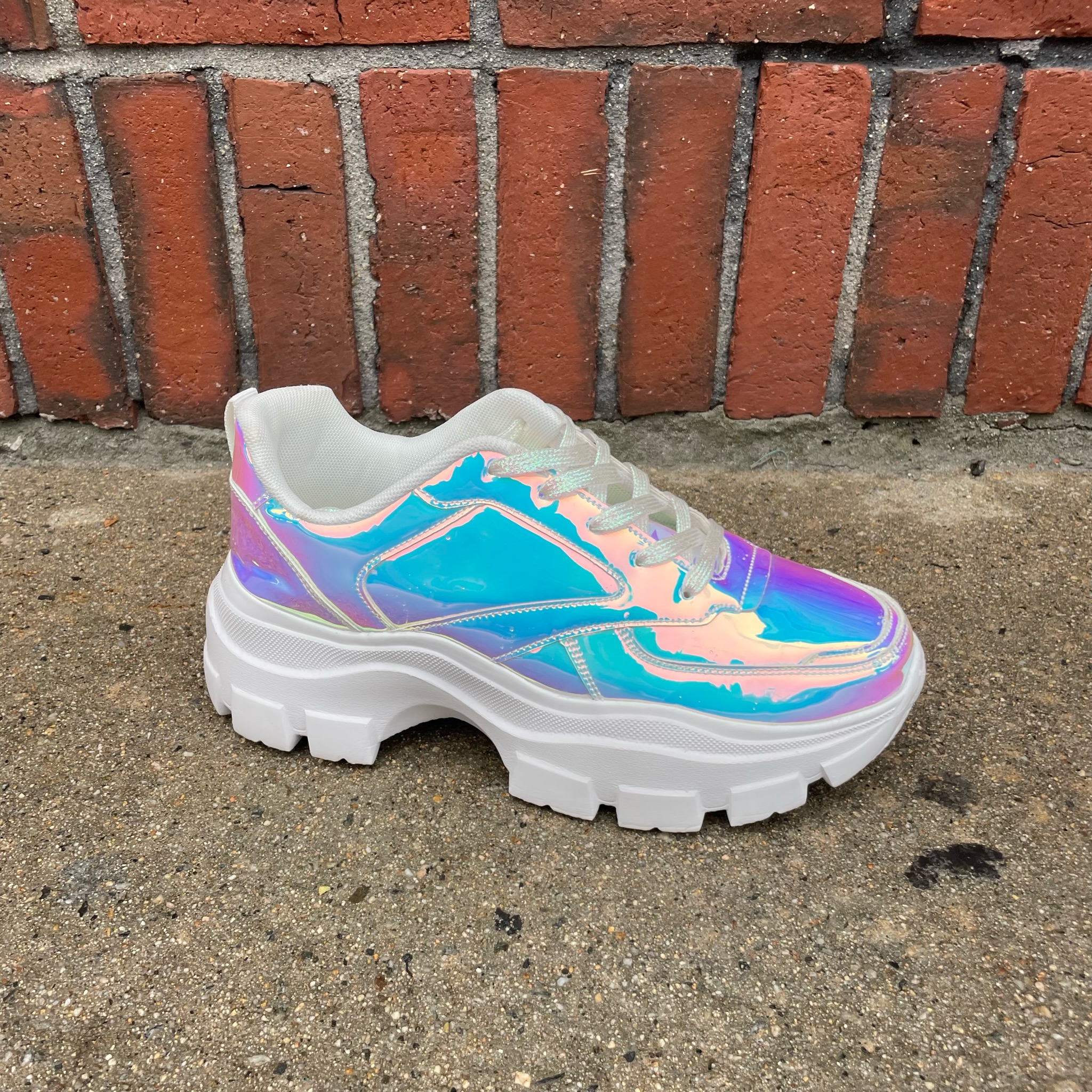} &
        &
        \includegraphics[width=0.135\linewidth]{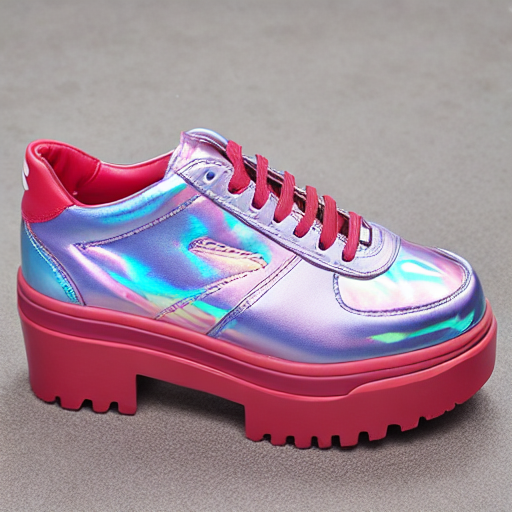} &
        \includegraphics[width=0.135\linewidth]{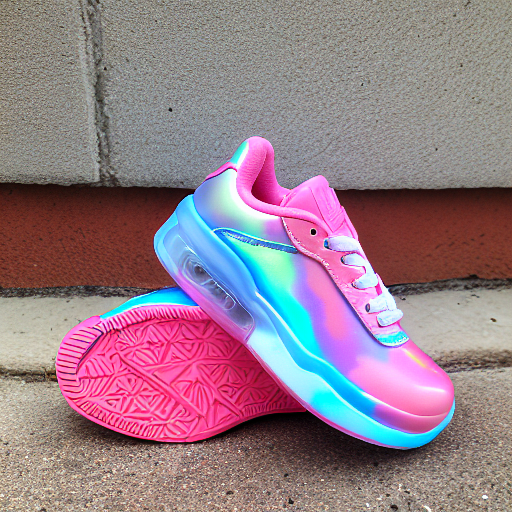} &
        \includegraphics[width=0.135\linewidth]{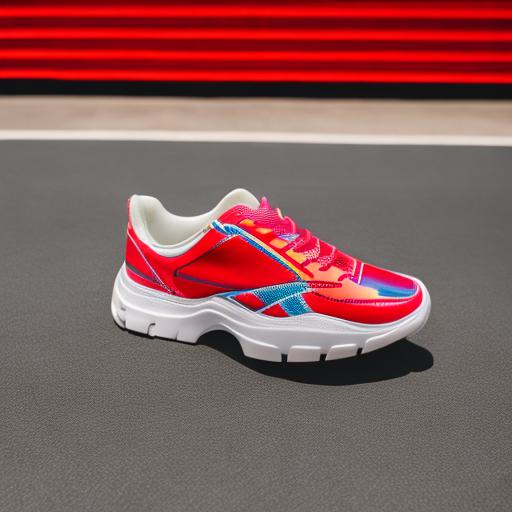} &
        \includegraphics[width=0.135\linewidth]{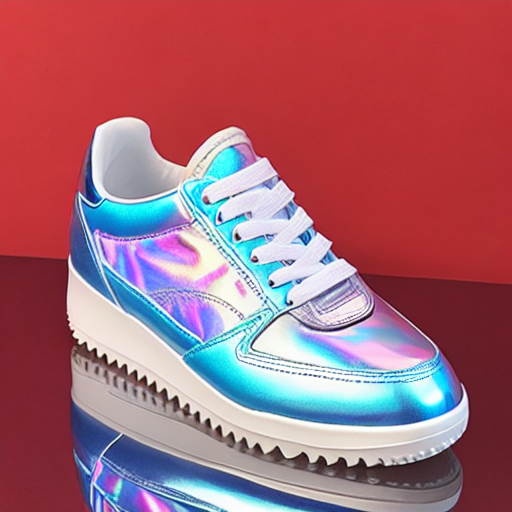} &
        \includegraphics[width=0.135\linewidth]{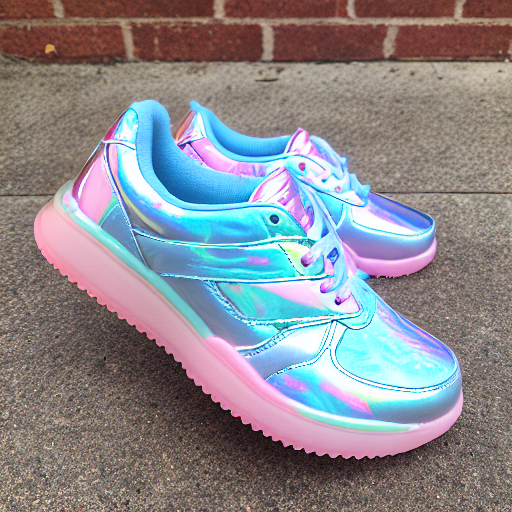} &
        \includegraphics[width=0.135\linewidth]{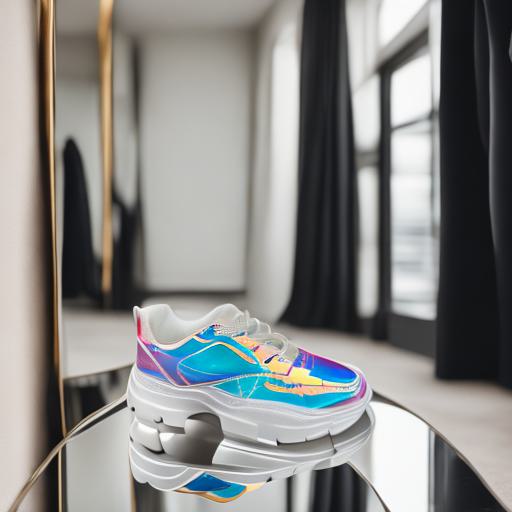} \\

         % First sample prompts
        \multicolumn{1}{c}{\textit{``A dog''}} & &\multicolumn{3}{c}{\textit{``in a firefighter outfit''}} &  \multicolumn{3}{c}{\textit{``on top of a purple rug in a forest''}} \\
        
        % First sample images
        \includegraphics[width=0.135\linewidth]{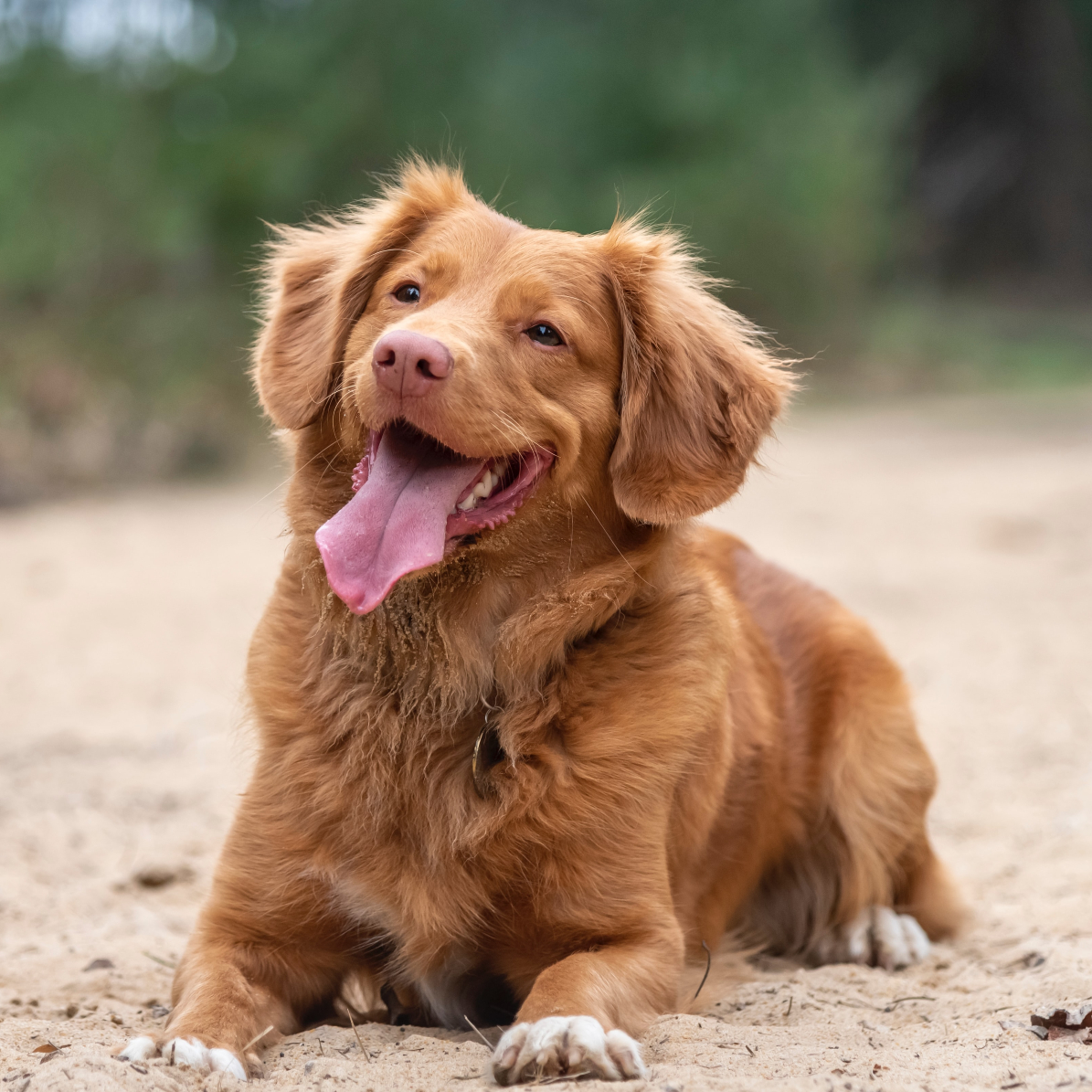} &
        &
        \includegraphics[width=0.135\linewidth]{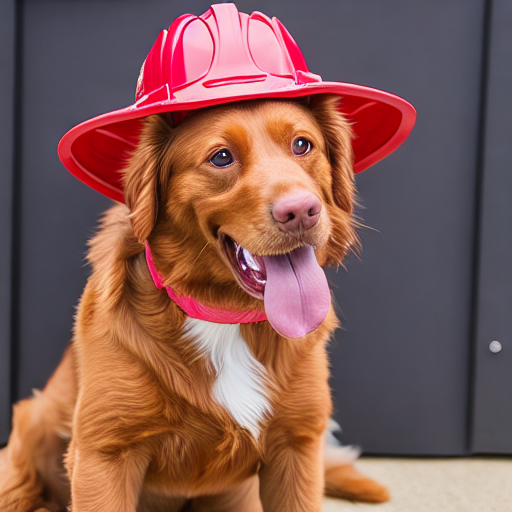} &
        \includegraphics[width=0.135\linewidth]{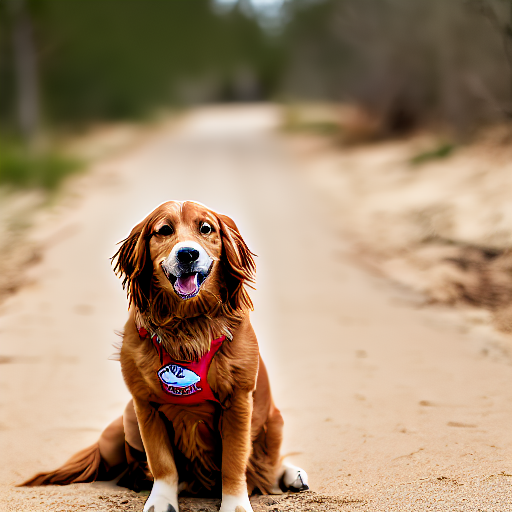} &
        \includegraphics[width=0.135\linewidth]{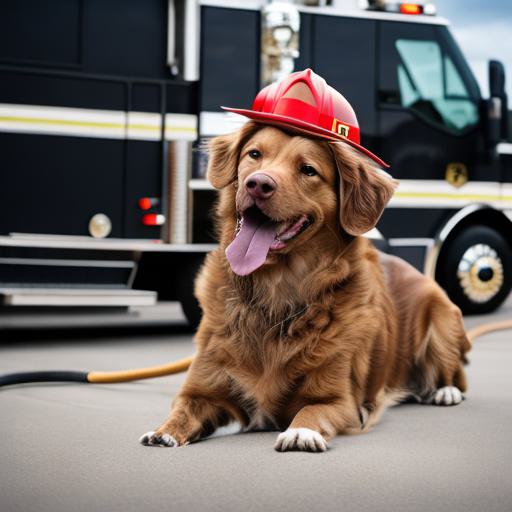} &
        \includegraphics[width=0.135\linewidth]{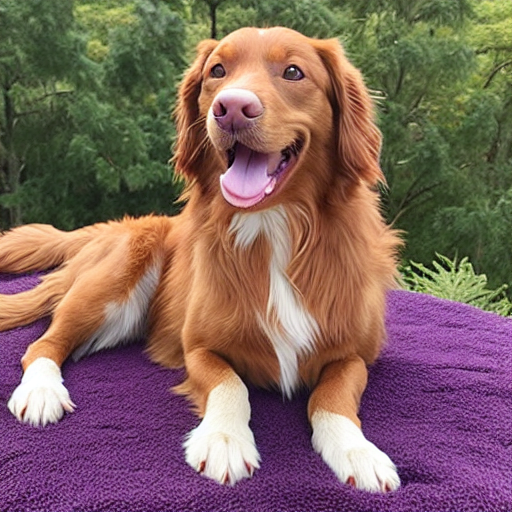} &
        \includegraphics[width=0.135\linewidth]{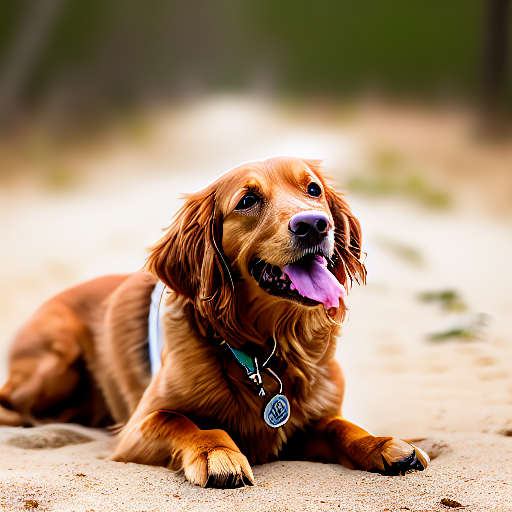} &
        \includegraphics[width=0.135\linewidth]{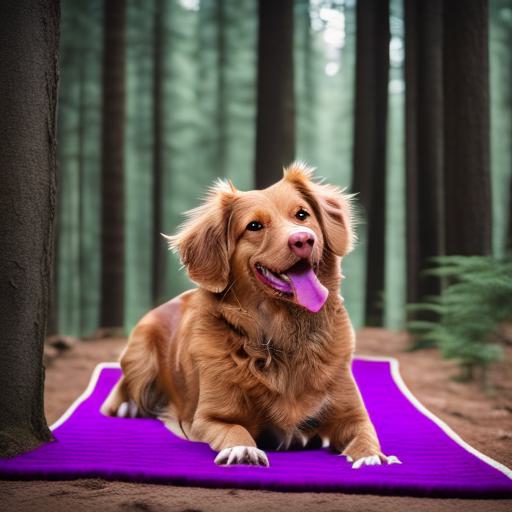} \\

         % First sample prompts
        \multicolumn{1}{c}{\textit{``A plushie''}} & &\multicolumn{3}{c}{\textit{``wet''}} &  \multicolumn{3}{c}{\textit{``in the snow''}} \\
        
        % First sample images
        \includegraphics[width=0.135\linewidth]{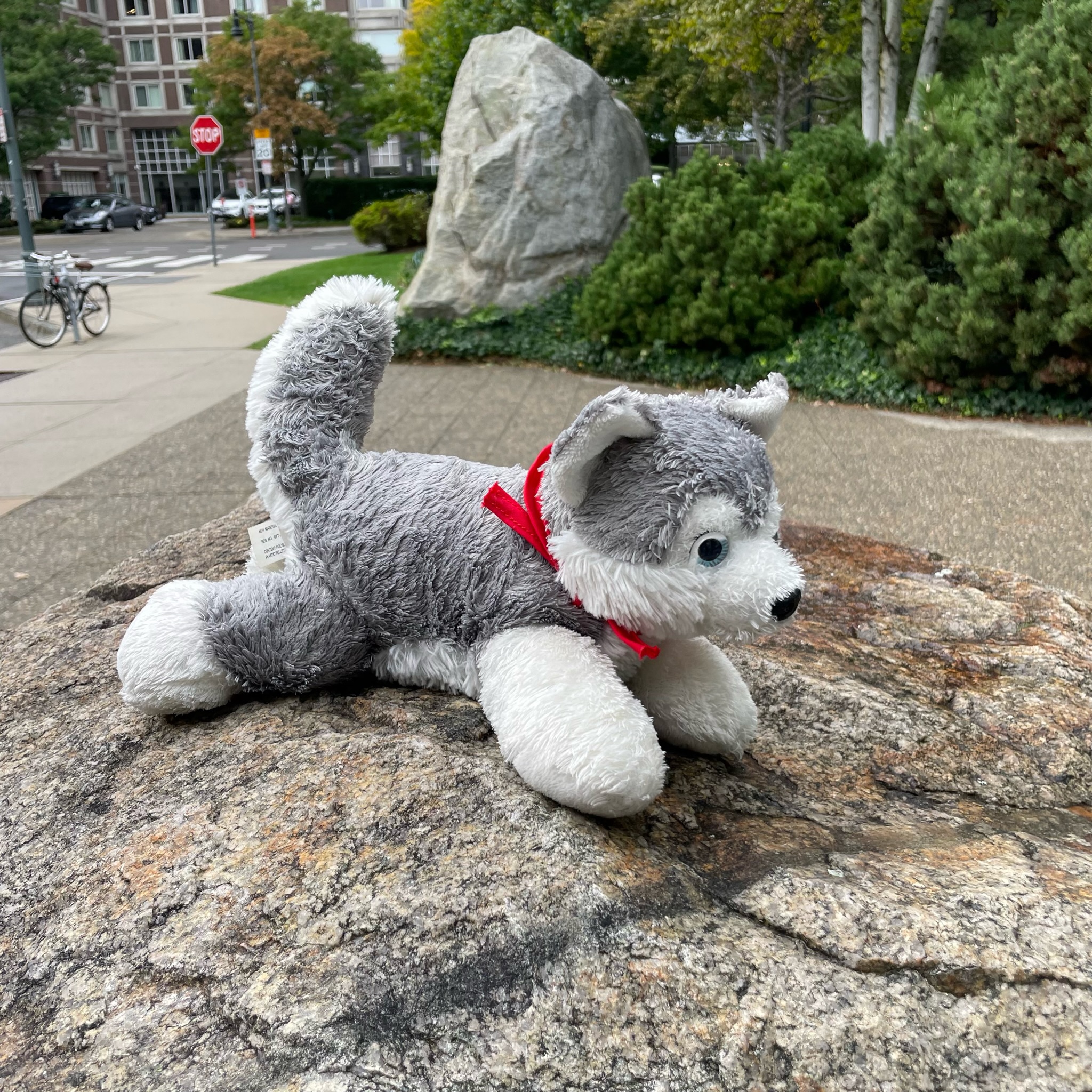} &
        &
        \includegraphics[width=0.135\linewidth]{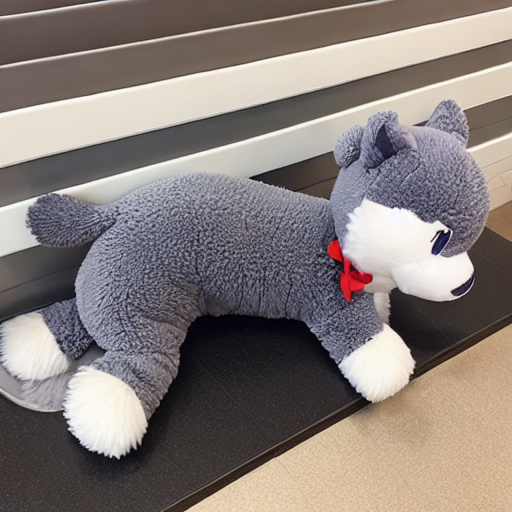} &
        \includegraphics[width=0.135\linewidth]{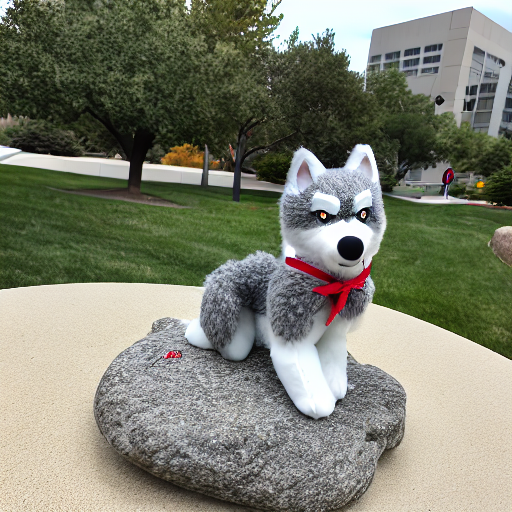} &
        \includegraphics[width=0.135\linewidth]{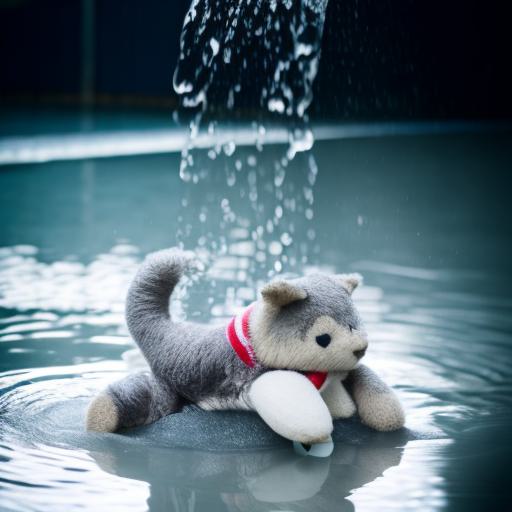} &
        \includegraphics[width=0.135\linewidth]{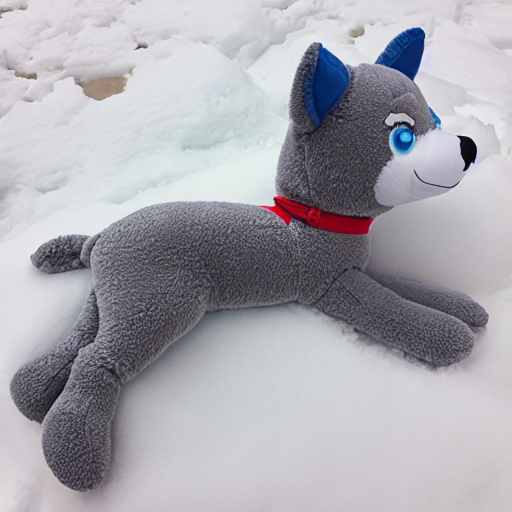} &
        \includegraphics[width=0.135\linewidth]{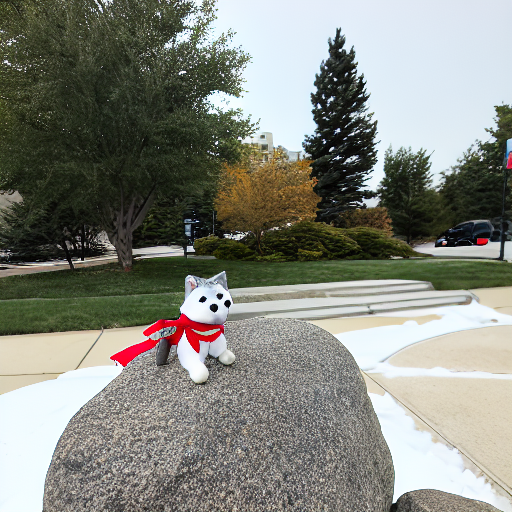} &
        \includegraphics[width=0.135\linewidth]{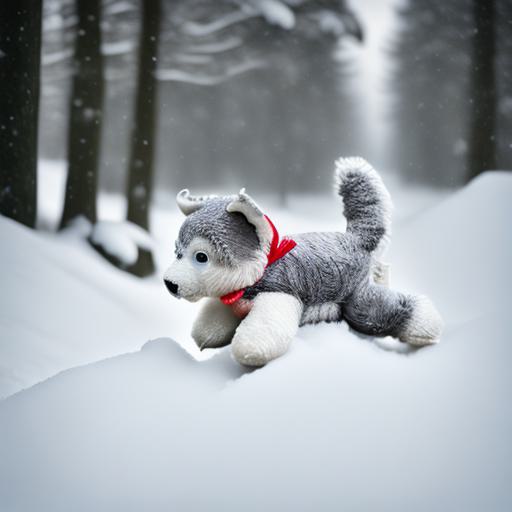} \\

          % First sample prompts
        \multicolumn{1}{c}{\textit{``A robot''}} & &\multicolumn{3}{c}{\textit{``on top of a dirty road''}} &  \multicolumn{3}{c}{\textit{``with a blue house in the background''}} \\
        
        % First sample images
        \includegraphics[width=0.135\linewidth]{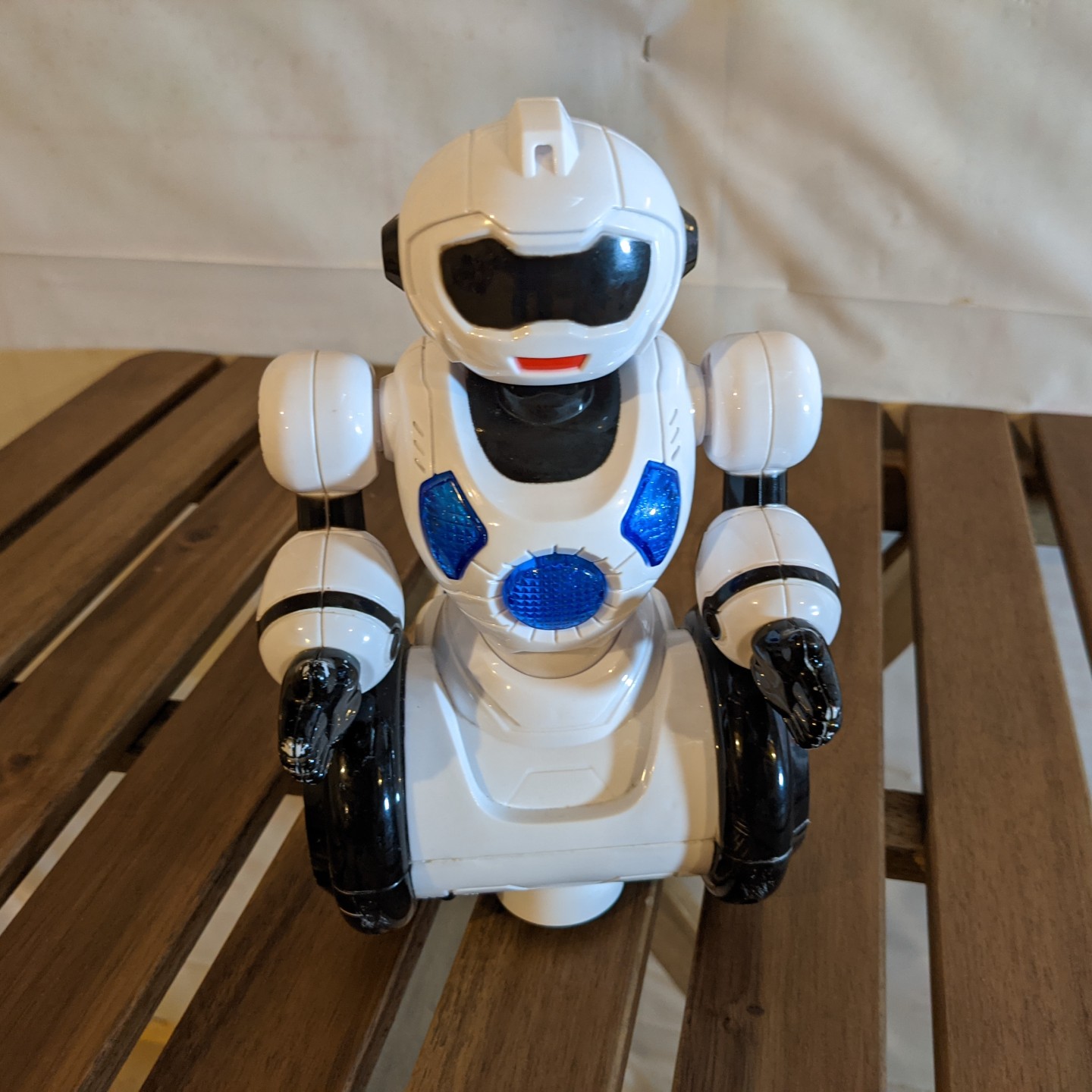} &
        &
        \includegraphics[width=0.135\linewidth]{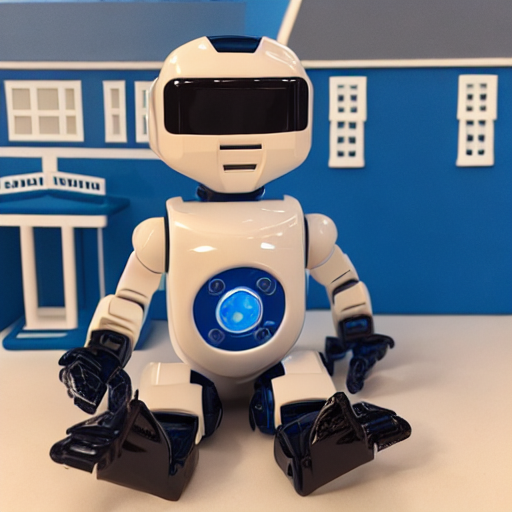} &
        \includegraphics[width=0.135\linewidth]{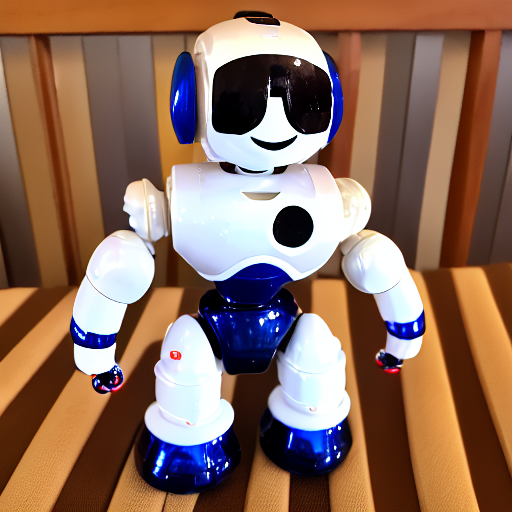} &
        \includegraphics[width=0.135\linewidth]{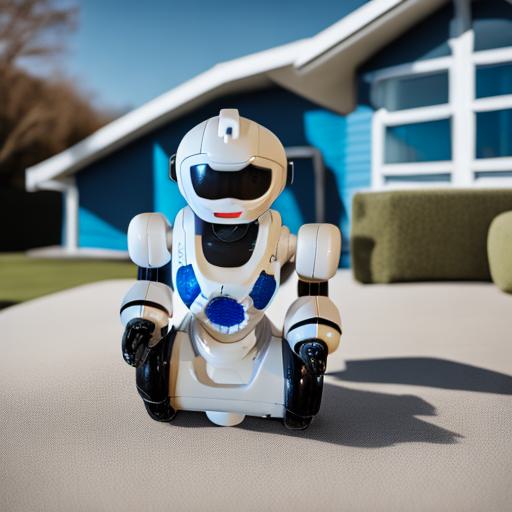} &
        \includegraphics[width=0.135\linewidth]{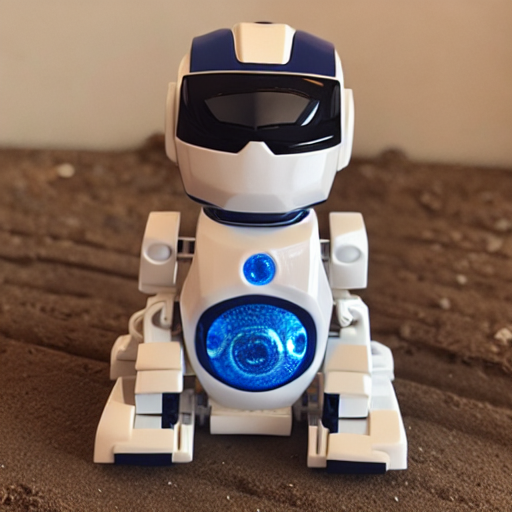} &
        \includegraphics[width=0.135\linewidth]{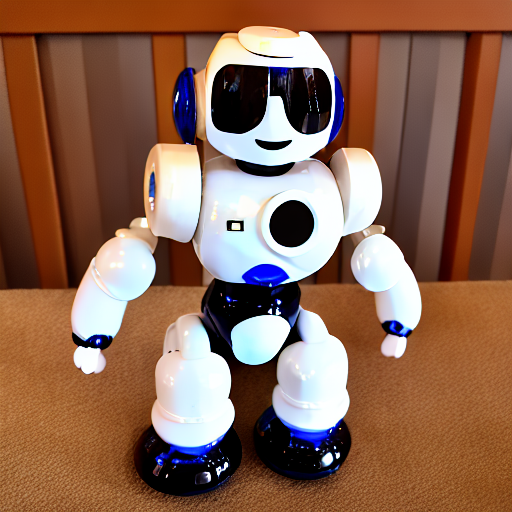} &
        \includegraphics[width=0.135\linewidth]{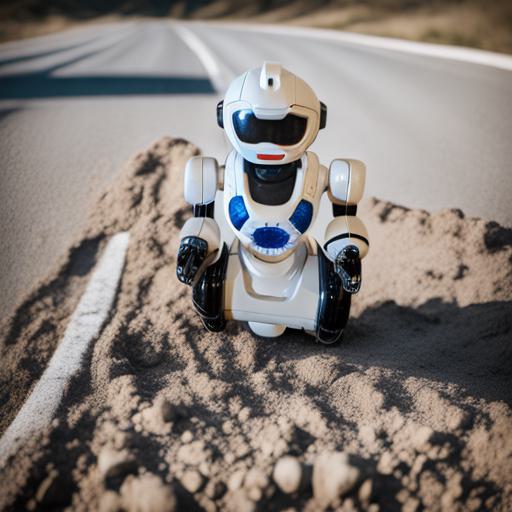} \\

          % First sample prompts
        \multicolumn{1}{c}{\textit{``A monster''}} & &\multicolumn{3}{c}{\textit{``floating on top of water''}} &  \multicolumn{3}{c}{\textit{``with mountains in the background''}} \\
        
        % First sample images
        \includegraphics[width=0.135\linewidth]{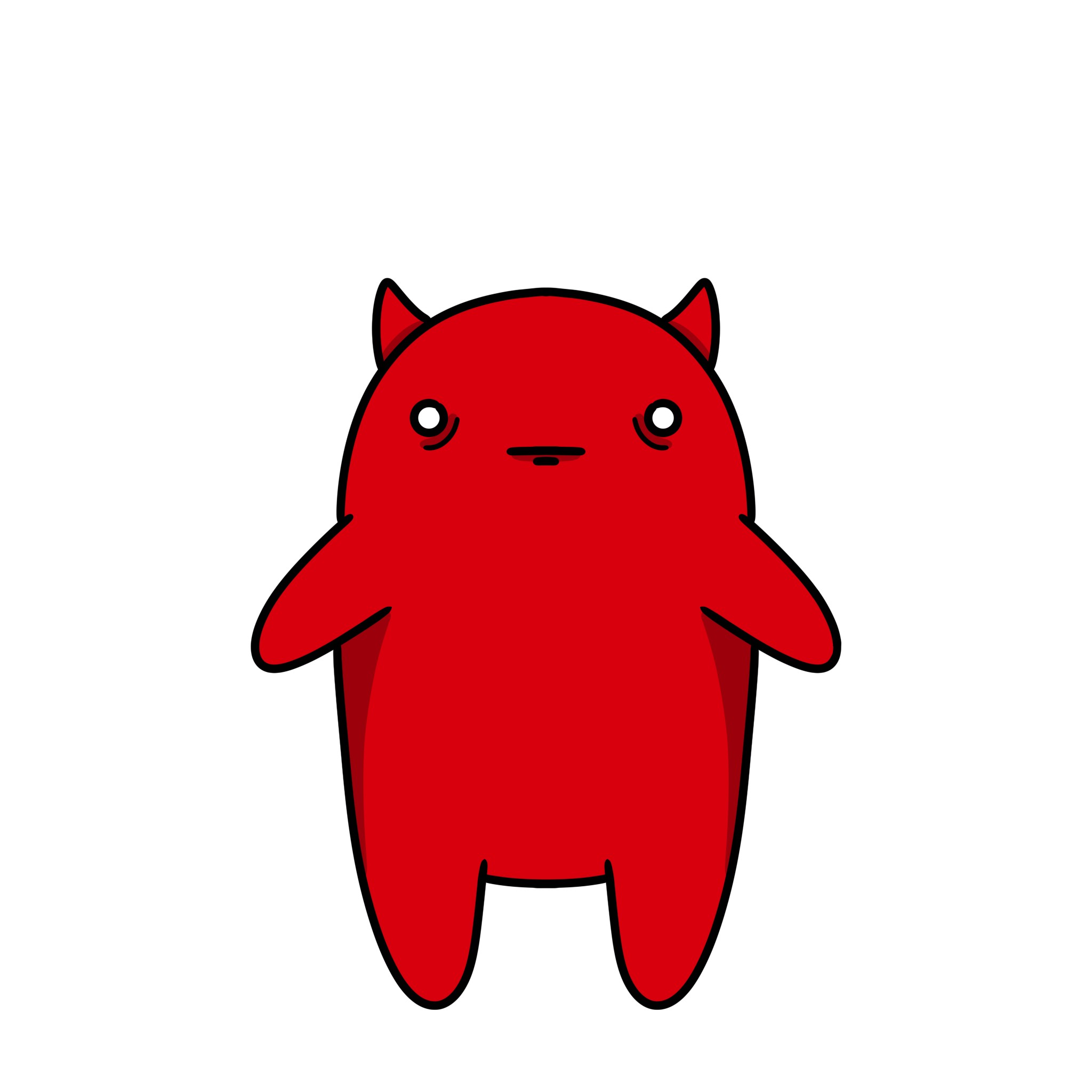} &
        &
        \includegraphics[width=0.135\linewidth]{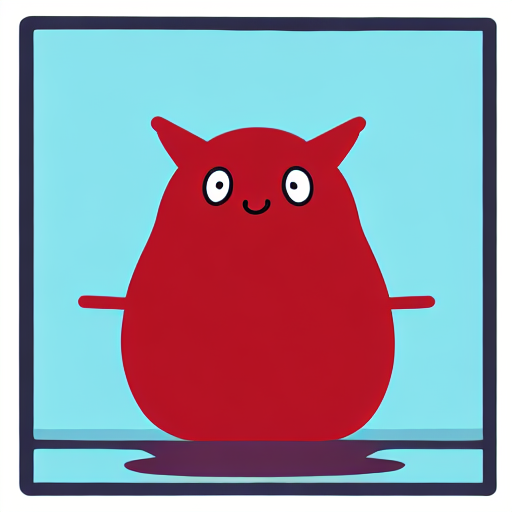} &
        \includegraphics[width=0.135\linewidth]{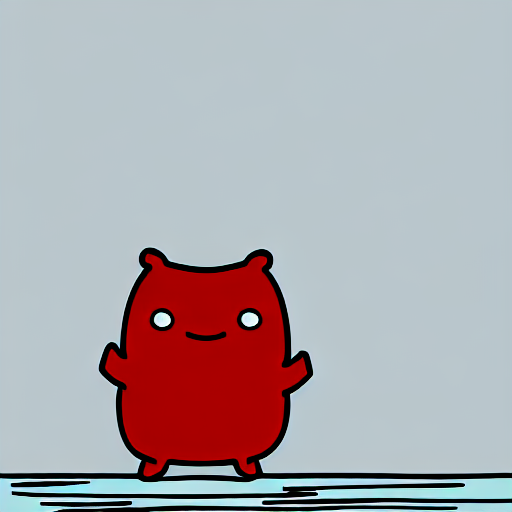} &
        \includegraphics[width=0.135\linewidth]{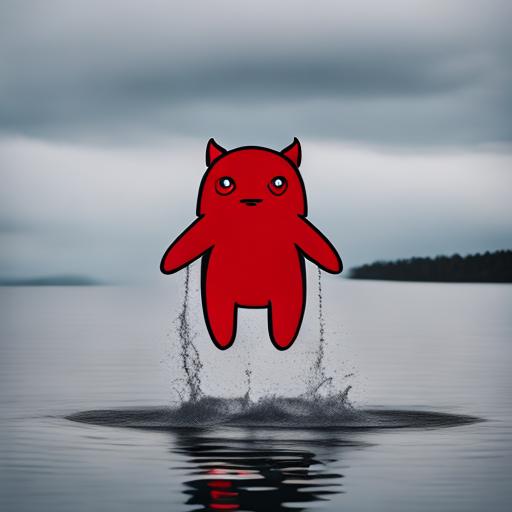} &
        \includegraphics[width=0.135\linewidth]{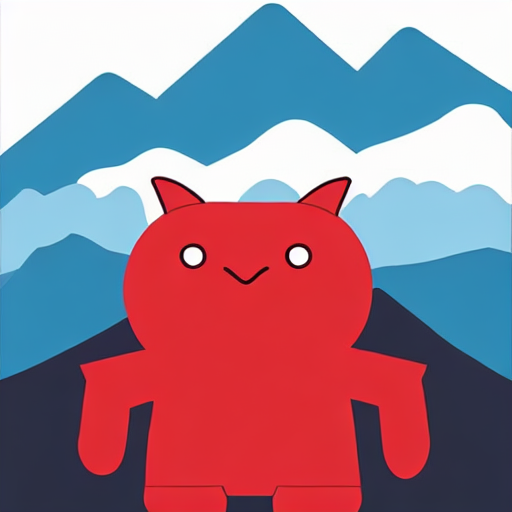} &
        \includegraphics[width=0.135\linewidth]{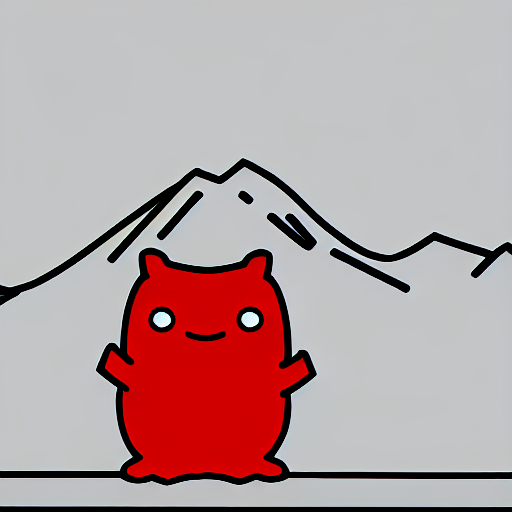} &
        \includegraphics[width=0.135\linewidth]{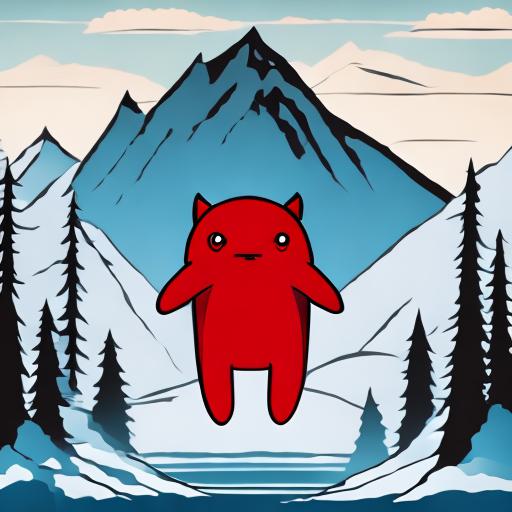} \\

          % Method names
        \textbf{Reference} & & \textbf{BLIP-D} & \textbf{Kosmos-G} & \textbf{\ourmethod\ } &  \textbf{BLIP-D} & \textbf{Kosmos-G} & \textbf{\ourmethod\ } \\

    \end{tabular}
    \caption{\textbf{Visual comparison.}  Personalized generations on sample concepts. \ourmethod\ preserves reference concept appearance and does not suffer from background interference. BLIP-D \cite{li2023blip} and Kosmos-G \cite{pan2023kosmos} cannot faithfully preserve visual details from the reference. }
    \label{fig:comparison_supp}
\end{figure*}

\paragraph{Additional Qualitative Comparisons} We also provide additional qualitative comparisons in Figure \ref{fig:comparison_supp}, including two reproducible open-source baselines: BLIP-D \cite{li2023blip} and Kosmos-G \cite{pan2023kosmos}.

\section{Additional Ablation Study}

\begin{table}
    \centering
    \setlength{\tabcolsep}{3pt}
    \caption{\textbf{Reference features} ablation study.}
        \begin{tabular}{lcc}
            \toprule 
        \# Reference Features  &  CLIP-I ($\uparrow$) & CLIP-T ($\uparrow$) \\
          
            \midrule

            Middle Features &  0.778 & 0.312 \\

            Respective Features &  0.810 & 0.298 \\

            \bottomrule
        \end{tabular}
    
    \label{tab:abl_feat}
\end{table}

\begin{table}
    \centering
    \setlength{\tabcolsep}{3pt}
    \caption{\textbf{Encoding timestep} ablation study.}
        \begin{tabular}{lcc}
            \toprule 
        Timestep  &  CLIP-I ($\uparrow$) & CLIP-T ($\uparrow$) \\
          
            \midrule

            1 &  0.810 &  0.298 \\

            150 & 0.800  & 0.299 \\

            300 &  0.789 &  0.301 \\

            \bottomrule
        \end{tabular}
    
    \label{tab:abl_time}
\end{table}

\paragraph{Impact of Encoding Timestep $t$} The proposed reference encoding mechanism relies on selecting $t = 1$ as a fixed timestep during the encoding process. We validate this design choice in Table \ref{tab:abl_time}, showing that $t = 1$ yields the best performance. This finding aligns with the intuition that less noisy features provide a more informative conditioning signal.
Furthermore, this experiment highlights a significant limitation of reference U-Net-based methods that inject noisy features corresponding to different timesteps. These noisy features are less informative and contain fewer details compared to the low-noise, fixed-timestep references we use to condition the generation independently of the current timestep.

\paragraph{Impact of Multi-Resolution Features} We also investigate the necessity of multi-resolution features for \ourmethod's performance. In a variant of our method, we fixed the cached features to a single resolution (i.e., the bottleneck resolution of the U-Net,($8 \times 8$), after the encoding stage). Our experiments demonstrate that leveraging multiple resolutions significantly enhances performance compared to using a single fixed-resolution cached feature map, as shown in Table \ref{tab:abl_feat}.

\section{Sampling Space and Image Guidance}

In our experiments we follow prior works \cite{brooks2023instructpix2pix, zeng2024jedi} and experiment with different types of guidance for image and text conditioning signal. The first and simpler \textit{joint guidance} approach jointly drops text and image conditioning for the unconditional prediction:
\begin{align*}
    \tilde{e_{\theta}}(z_t, c_I, c_T) = &\ e_{\theta}(z_t, \varnothing, \varnothing) \\ &+ s \cdot (e_{\theta}(z_t, c_I, c_T) - e_{\theta}(z_t, \varnothing, \varnothing))
\end{align*}
Where $\tilde{e_{\theta}}(z_t, c_I, c_T)$ represents the adjusted prediction at denoising step $t$ conditioned on textual conditioning $c_T$ and the image conditioning $c_I$. $e_{\theta}(z_t, \varnothing, \varnothing)$ denotes the unconditional prediction, and $s$ is the guidance scale.
The second approach, that we call \textit{combined guidance} decouples text and image allowing for a more flexible balance between the two conditioning modalities:
\begin{align*}
        \tilde{e_{\theta}}(z_t, c_I, c_T) = &\: e_{\theta}(z_t, \varnothing, \varnothing) \\ &+ s_I \cdot (e_{\theta}(z_t, c_I, \varnothing) - e_{\theta}(z_t, \varnothing, \varnothing))\nonumber \\ &+ s_T \cdot (e_{\theta}(z_t, c_I, c_T) - e_{\theta}(z_t, c_I, \varnothing)) \nonumber
\end{align*}
Our experimental findings suggest that using a higher image guidance scale better preserves the content of the reference image, but reduces editability of the subject. Conversely, decreasing image guidance results in more flexible editing of the reference subject at the expense of reduced subject fidelity. Figure \ref{fig:sampling_space} illustrates these findings on the DreamBooth dataset, comparing the joint and combined guidance strategies.

\section{Broader Impact}
\ourmethod\ allows users to customize the subject of their images, focusing on individual elements such as animals or objects. However, it is crucial to recognize that, like other generative models and image editing tools, this technology has the potential to be misused for creating misleading content. Addressing these ethical risks is an essential and ongoing focus in the field of generative modeling, particularly in relation to deepfake creation. Techniques such as watermarking or content detection are particularly necessary to prevent misuse of this technology.

% Having the supplementary compiled together with the main paper means that:
% % 
% \begin{itemize}
% \item The supplementary can back-reference sections of the main paper, for example, we can refer to \cref{sec:intro};
% \item The main paper can forward reference sub-sections within the supplementary explicitly (e.g. referring to a particular experiment); 
% \item When submitted to arXiv, the supplementary will already included at the end of the paper.
% \end{itemize}
%

\end{document}